\documentclass[twoside]{article}

\usepackage[accepted]{aistats2021}
\usepackage[utf8]{inputenc} 
\usepackage[T1]{fontenc}    
\usepackage{hyperref}       
\usepackage{url}            
\usepackage{booktabs}       
\usepackage{amsfonts}       
\usepackage{nicefrac}       
\usepackage{microtype}      
\usepackage{amsthm}
\usepackage{xr}
\usepackage{enumitem}

\usepackage{graphicx}
\usepackage{subcaption}
\usepackage{amsmath}
\usepackage{amssymb}
\usepackage{xspace}
\usepackage{wrapfig}

\usepackage{textcomp} 
\usepackage{algorithm, algorithmic}

\newcommand{\xadv}[1]{{ x}_{adv}^{(#1)}}
\newcommand{\sourceimage}{{source-image}\xspace}
\newcommand{\Sourceimage}{{Source-image}\xspace}
\newcommand{\targetimage}{{target-image}\xspace}
\newcommand{\Targetimage}{{Target-image}\xspace}
\newcommand{\advimage}{{adv-image}\xspace}

\newcommand{\boundaryimage}{{boundary-image}\xspace}

\newcommand{\sysname}{NonLinear-BA\xspace}

\usepackage{hyperref}
\usepackage{breakurl}

\usepackage{amsmath,mathtools,amssymb,amsthm,dsfont,cleveref}
\usepackage{dsfont}

\def\ddefloop#1{\ifx\ddefloop#1\else\ddef{#1}\expandafter\ddefloop\fi}
\def\ddef#1{\expandafter\def\csname bb#1\endcsname{\ensuremath{\mathbb{#1}}}}
\ddefloop ABCDEFGHIJKLMNOPQRSTUVWXYZ\ddefloop
\def\ddef#1{\expandafter\def\csname b#1\endcsname{\ensuremath{\mathbf{#1}}}}
\ddefloop ABCDEFGHIJKLMNOPQRSTUVWXYZ\ddefloop
\def\ddef#1{\expandafter\def\csname c#1\endcsname{\ensuremath{\mathcal{#1}}}}
\ddefloop ABCDEFGHIJKLMNOPQRSTUVWXYZ\ddefloop
\def\ddef#1{\expandafter\def\csname g#1\endcsname{\ensuremath{\mathcal{#1}}}}
\ddefloop ABCDEFGHIJKLMNOPQRSTUVWXYZ\ddefloop

\DeclareMathOperator*{\argmax}{argmax}

\def\1{\mathds{1}}

\def\diff{\mathrm{d}}

\def\T{{\scriptscriptstyle\mathsf{T}}}

\newtheorem{lemma}{Lemma}

\newtheorem{theorem}{Theorem}
\newtheorem{fact}{Fact}[theorem]
\newtheorem{corollary}{Corollary}

\newtheorem{innercustomthm}{Theorem}
\newenvironment{customthm}[1]
  {\renewcommand\theinnercustomthm{#1}\innercustomthm}
  {\endinnercustomthm}

\newtheorem{innercustomcor}{Corollary}
\newenvironment{customcor}[1]
  {\renewcommand\theinnercustomcor{#1}\innercustomcor}
  {\endinnercustomcor}

\theoremstyle{definition}
\newtheorem{definition}{Definition}

\theoremstyle{remark}
\newtheorem*{remark}{Remark}

\usepackage[round]{natbib}
\usepackage{xcolor}

\newcommand{\f}{\mathbf{f}}
\newcommand{\sgn}{\mathrm{sgn}}
\newcommand{\nablaS}{\nabla S}
\newcommand{\tnablaS}{\widetilde{\nabla S}}
\newcommand{\nablafS}{\nabla \f^\T \nabla S}
\newcommand{\tnablafS}{\widetilde{\nabla \f^\T \nabla S}}

\newcount\Comments  
\usepackage{color}
\definecolor{darkgreen}{rgb}{0,0.5,0}
\definecolor{darkblue}{rgb}{0,0,0.5}
\definecolor{purple}{rgb}{1,0,1}
\newcommand{\kibitz}[2]{\ifnum\Comments=0\textcolor{#1}{#2}\fi}
\newcommand{\Huichen}[1]{\kibitz{blue}      {[HL: #1]}}

\newcommand{\Linyi}[1]{\kibitz{darkgreen}{LL: #1}}

\begin{document}

%

%
\runningauthor{Huichen Li$^{*}$, Linyi Li$^{*}$, Xiaojun Xu, Xiaolu Zhang, Shuang Yang, Bo Li}

\twocolumn[

\aistatstitle{Nonlinear Projection Based Gradient Estimation for Query Efficient Blackbox Attacks}

\aistatsauthor{Huichen Li$^{1*}$\quad Linyi Li$^{1*}$ \quad Xiaojun Xu$^{1}$ \quad Xiaolu Zhang$^{2}$\quad Shuang Yang$^{3}$\quad Bo Li$^{1}$
\\
$^{1}$ University of Illinois at Urbana-Champaign\quad $^{2}$ Ant Financial\quad
$^{3}$ Alibaba Group US\\
\small\texttt{\{\href{mailto:huichen3@illinois.edu}{huichen3},
\href{mailto:linyi2@illinois.edu}{linyi2}, 
\href{mailto:xiaojun3@illinois.edu}{xiaojun3}, 
\href{mailto:lbo@illinois.edu}{lbo}\}@illinois.edu, \href{mailto:yueyin.zxl@antfin.com}{yueyin.zxl@antfin.com}, \href{mailto:shuang.yang@alibaba-inc.com}{shuang.yang@alibaba-inc.com}}
}
\aistatsaddress{\small\textit{* The first two authors contributed equally.}}

]



\begin{abstract}
  \emph{Gradient estimation} and \emph{vector space projection} have been studied as two distinct topics. We aim to bridge the gap between the two by investigating how to efficiently estimate gradient based on a projected low-dimensional space. We first provide lower and upper bounds for gradient estimation under both linear and nonlinear gradient projections, and outline checkable sufficient conditions under which one is better than the other. Moreover, we analyze the query complexity for the projection-based gradient estimation and 
present a sufficient condition for query-efficient estimators. Built upon our theoretic analysis, we propose a novel 
query-efficient {Nonlinear} Gradient Projection-based {B}oundary Blackbox {A}ttack
(\textbf{\sysname}).
We conduct extensive experiments on four datasets: ImageNet, CelebA, CIFAR-10, and MNIST, and show the superiority of the proposed methods compared with the state-of-the-art baselines. 
In particular, we show that the projection-based
 boundary  blackbox attacks are able to achieve much smaller magnitude of perturbations with $100\%$ attack success rate based on efficient queries. Both linear and nonlinear projections demonstrate their advantages under different conditions. We also evaluate \sysname against the commercial online API MEGVII Face++, and demonstrate the high blackbox attack performance both quantitatively and qualitatively. The code is publicly available at \burl{https://github.com/AI-secure/NonLinear-BA}.
\end{abstract}

\section{Introduction}
Gradient estimation and vector space projection have both been extensively studied in machine learning, but largely for different purposes. Gradient estimation is used when gradient-based optimization such as back-propagation is employed but the exact gradients are not directly accessible, for example, in the case of blackbox adversarial attacks~\citep{chen2020hopskipjumpattack,cvpr2020QEBA}. Vector space projection, especially gradient projection (or sparsification), on the other hand, has been used to speedup training, for instance, by reducing the complexity of communication and/or storage when performing model update in distributed training~\citep{wangni2018gradient}. 
In this paper, we aim to bridge the gap between the two and attempt to answer the following questions:
\emph{Can we estimate gradients from a projected low-dimensional subspace?} 
\emph{How do different projections affect the gradient estimation quality?} 

Our investigation is motivated in particular by the challenging problem of blackbox adversarial attacks~\citep{bhagoji2017exploring, ilyas2018black}.
Adversarial attacks have the ability to mislead machine learning models with potentially catastrophic consequences while staying imperceptible to humen. While extensive progresses have been made in white-box attacks~\citep{carlini2016towards,evtimov2017robust,Xu_2018_CVPR} where attackers have complete knowledge about the target model, the more realistic scenario of blackbox attacks where the attacker only has query access to the target model remains challenging. One major challenge is the excessive query complexity. For example, boundary-based blackbox attacks (BA)~\citep{brendel2017decision} have shown promising attack effectiveness, but the required query number is too large to be practically feasible (e.g., many approaches require $10^5$ or more queries per attack, which could take hours or even days given the rate limit of public machine learning APIs). This inefficiency stems partially from the high dimensionality of the gradient since the Monte Carlo gradient estimation relies on sampling perturbations 
from the gradient space. 

In this work, we study the properties of a general vector space  projection $\f$, that transforms vectors from low-dimensional subspace $\bbR^n$ to the original gradient space $\bbR^m$ for gradient estimation.
We theoretically provide the lower and upper bounds of  cosine similarity between the estimated and true gradients, based on sampling distribution analysis and Taylor expansion.
These bounds imply that it is possible to estimate the gradient effectively under checkable sufficient condition.
Intuitively, the condition measures how well the estimated and true gradients of the target model align with each other.
Furthermore, we compare linear and nonlinear gradient projections in terms of the cosine similarity between the estimated and true gradients, and prove the existence of nonlinear projection that is able to achieve strictly higher cosine similarity lower bound.
We finally analyze the query complexity of gradient estimation and present a sufficient condition for query-efficient projection-based gradient estimation.
The analysis provides theoretic answers to the aforementioned questions.
\textit{Our theoretic analysis on the projection-based gradient estimation is not specific to adversarial attacks, but can shed light on a broader range of applications such as gradient sparsification and distributed training}.

Based on our analysis for query-efficient gradient estimation, we propose \textbf{NonLinear-BA}, which applies deep generative models such as AEs, VAEs, and GANs as the nonlinear projections to perform blackbox attack, and therefore evaluate the power of projection-based gradient estimation empirically.
Once trained, these generative models are used to project the sampled low-dimensional vectors back to high-dimensional gradient space and query the target model to estimate the gradient.
We experimentally evaluate \sysname with three proposed nonlinear projections on four image datasets: ImageNet~\citep{imagenet_cvpr09}, CelebA~\citep{liu2015faceattributes}, CIFAR-10~\citep{krizhevsky2009learning} and MNIST~\citep{lecun1998gradient}. We show that \sysname can achieve 100\% attack success rate more efficiently with smaller magnitude of perturbation compared with baselines.
We also evaluate the \sysname against a commercial online API MEGVII Face++~\citep{facepp_main}. Both quantitative and qualitative results are shown to demonstrate its  attack effectiveness.

\textbf{Contributions:} 
(1) We provide the first general theoretical analysis framework for the projection-based gradient estimation, analyzing the cosine similarity between estimated and true gradients under different vector space projections. 
(2) We prove and compare the lower bounds of gradient cosine similarities for linear and nonlinear projections. We also analyze the query complexity of the projection based gradient estimators.
(3) We propose a novel nonlinear gradient projection-based blackbox attack (\sysname) which exploits the power of nonlinear-projection based gradient estimation.
(4) We conduct extensive experiments on both offline ML models and commercial online APIs with high-dimensional image datasets to demonstrate the high attack performance of \sysname. The empirical results verify our theoretical findings that the projection-based gradient estimation via sampling is query efficient, and some projections outperform others under certain conditions. 

\textbf{Related Work:} 
    The vulnerability of ML to adversarial attacks has been demonstrated by recent studies~\citep{szegedy2014intriguing,goodfellow2014explaining}.
    To better understand such threat, new attacks have been consistently proposed over years, which lie in two major branches: \emph{whitebox attacks} and \emph{blackbox attacks}.
    The whitebox attacks, e.g., \citep{goodfellow2014explaining,carlini2016towards,kurakin2016adversarial,madry2018towards,athalye2018obfuscated}, assume full knowledge of the victim model for an attacker;
    while blackbox attacks only require limited access to the victim model, which is more applicable in practice.

    The \emph{blackbox attacks} can be divided into two categories: transfer-based and query-based attacks.
    The transfer-based attacks rely on adversarial transferability~\citep{papernot2016transferability,tramer2017space}, where the adversarial examples generated against one ML model can also attack another model.
    Various approaches including ensemble methods have been explored to enhance the adversarial transferability~\citep{liu2016delving}.
    One type of query-based attacks utilizes the zeroth-order information, i.e., the confidence scores, to estimate the gradient of the blackbox model.
    A series of works~\citep{chen2017zoo,bhagoji2018practical,ilyas2018black,tu2019autozoom,cheng2019improving} have been proposed to improve the efficiency of gradient estimation.
    Another type of query-based attacks assumes only the final prediction labels are accessible by the attackers.
    RayS~\citep{chen2020rays} proposes a search-based gradient-free untargeted attack to minimize the $L_{\infty}$ norm perturbations.
    \emph{Boundary-based blackbox attack (BA)}~\citep{brendel2017decision} 
    focuses on the targeted attack with $L_2$ constraints.
    Works have been conducted to improve the query efficiency for BA. For instance,
    \citet{cheng2019sign} perform gradient sign estimation,
    \citet{chen2020hopskipjumpattack} apply the Monte-Carlo sampling strategy to perform gradient estimation, and \citet{cvpr2020QEBA} improve the estimation by sampling from representative low-dimensional orthonormal subspace.
    Our work, on the other hand, aims to explore more general projection-based gradient estimators with a unified theoretical analysis framework.

\section{Problem Definition}
In this section, we will first introduce the framework of \emph{boundary-based blackbox attack}, and then focus on tackling the challenge of query-based gradient estimation.

\paragraph{Boundary-Based Blackbox Attack (BA).} 
Given an instance $x$ drawn from certain distribution $\mathcal{D}:x \sim \mathcal{D}$, where $x \in \bbR^m$, 
 a $C$-way classification model $G:\,\bbR^m \mapsto \bbR^C$ is trained to output the confidence score for each class.
The final prediction of the model is obtained by selecting the class with the highest confidence score $y = \argmax_{i\in [C]} G(x)_i$~$([C] = \{1,\,\dots,\,C\})$.
The model $G$ is referred to as `target model' throughout our discussion as it is the target of the adversarial attack. 
In this work we focus on the scenario where the adversaries do not have access to the details of model $G$ (i.e. blackbox attack) and can only query the model to obtain the final prediction label $y$ instead of the confidence scores.

The general framework of a BA is as follows:
given a \targetimage ${x}_{tgt} \in \bbR^m$ whose true label is $y_{ben} \in [C]$, 
the attacker's goal is to craft an adversarial image ${x}_{adv}$ that is predicted as a maliciously chosen label $y_{mal} \in [C]$, while the distance $D({x}_{tgt}, {x}_{adv})$ between the two images is as small as possible. Here $D$ is a $L_p$-norm based distance function which aims to
restrict the perturbation added to the \targetimage in order to make it less noticeable.
In this paper we only consider targeted attack with an intentionally chosen $y_{mal}$ since untargeted attack is a trivial extension of the targeted case (by randomly sampling a $y_{mal}$).

\begin{definition}[$(G,y_{mal})$)-Difference Function]
    Given a model $G$, and malicious target $y_{mal}$, the \emph{difference function} $S:\,\bbR^m \to \bbR$ is defined as $S(x) = G(x)_{y_{mal}} - G(x)_{y_{ben}}$, where $y_{ben}$ denotes the ground truth label.
    \label{def:difference-function}
\end{definition}
The difference function $S$ is an important indicator of whether the image is successfully perturbed from being predicted as $y_{ben}$ to $y_{mal}$.
A \emph{\boundaryimage} is an image $x$ that lies on the decision boundary
between $y_{ben}$ and $y_{mal}$
,
i.e., $S(x) = 0$.

\paragraph{Projection-Based Gradient Estimation.} There are three main steps to perform the BA: (1)~gradient estimation at $G$'s decision boundary, (2)~move the boundary-image along the estimated gradient direction, and (3)~project the image back to the decision boundary. 
Typically, the first step requires to estimate the gradient based on the sign of \emph{difference function} defined in Definition~\ref{def:difference-function} given multiple queries.
It is very computationally expensive as the high-dimensional gradient estimation requires a large number of queries~\citep{chen2020hopskipjumpattack}.
Based on recent advances in efficient communication and gradient sparsification~\citep{wangni2018gradient}, we hypothesize that there exist lower dimensional \emph{supports} for gradient vectors and we aim to project the gradient to these lower dimensional supports and perform the estimation efficiently.
In particular, we theoretically analyze the impacts of linear and nonlinear gradient projections on gradient estimation.


\section{\sysname: Nonlinear Gradient Projection-based Boundary Blackbox Attack}
\begin{figure}
    \centering
    \includegraphics[width=\linewidth]{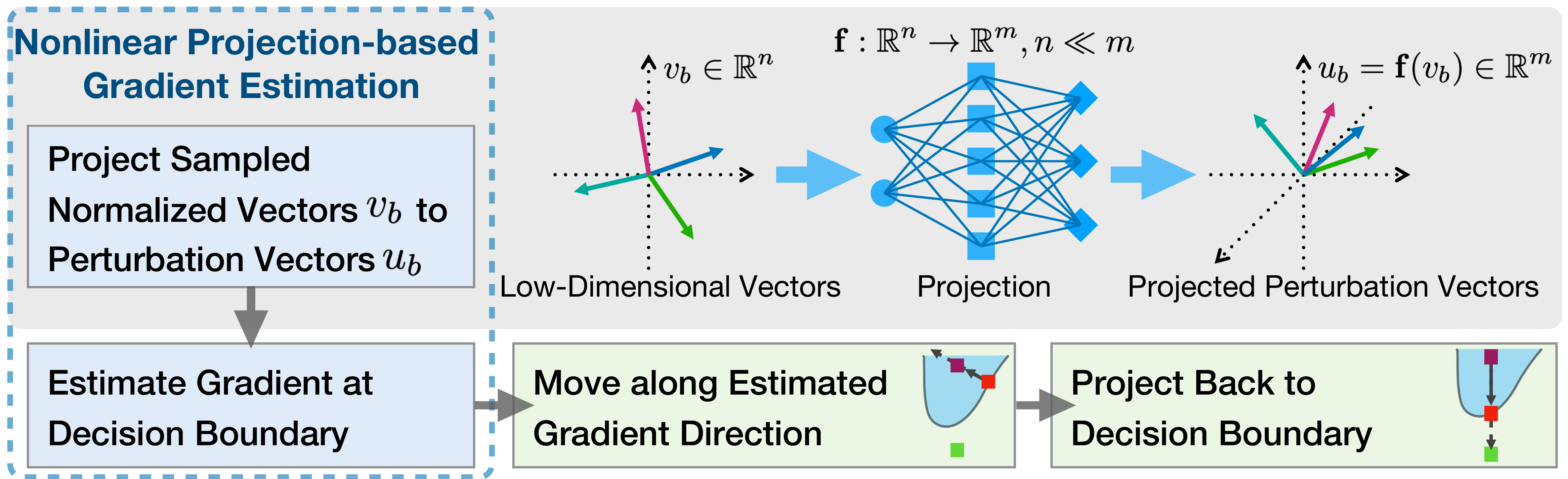}
    \caption{\small Algorithm illustration for \sysname.}
    \label{fig:overview}
    \vspace{-1em}
\end{figure}
In this section we introduce the proposed nonlinear gradient projection-based boundary blackbox attack~(\sysname) as illustrated in Figure~\ref{fig:overview}, followed by the detailed theoretical analysis and guarantees in Section~\ref{sec:theory}. 

In standard BA, the way to estimate the gradient given the query results is done by Monte Carlo sampling method~\citep{chen2020hopskipjumpattack}:
\begin{equation}
    \small
    \vspace{-0.5em}
    \widetilde{\nablaS}(\xadv{t}) = \frac1B \sum_{b=1}^B \sgn \left(S\left(\xadv{t}+\delta {u}_b\right) \right) { u}_b,
    \label{eq:MC_gradient_estimation}
\end{equation}
%
where $\xadv{t}$ is the \boundaryimage at iteration $t$ obtained by binary search with precision threshold $\theta$ following Equation~\ref{eqn:binary}~(to be shown later). The ${ u}_b$'s are $B$ perturbation vectors uniformly sampled from the unit sphere in $\mathbb{R}^m$. 
The size of random perturbation $\delta$ is chosen as a function of image size and the binary search threshold~\citep{chen2020hopskipjumpattack} to control the gradient estimation error caused by the \boundaryimage's offset from the exact decision boundary due to binary search precision.
The function $\sgn\left(S(\cdot)\right)$ denotes the sign of the difference function (Definition~\ref{def:difference-function}).
Its value is acquired by querying the victim model and comparing the output label with $y_{mal}$.
It is clear that the query cost is very high when the input dimension $m$ is large. A typical $3$-channel $224\times 224$ image gradient vector has a dimension $m$ of over $150$k. It is challenging to perform accurate estimation in such a high-dimensional space with limited queries.
To reduce the query complexity, 
\citet{cvpr2020QEBA} propose to search for a \emph{representative subspace} with orthonormal mappings $\bW=[w_1, \ldots, w_n] \in \mathbb{R}^{m\times n}, n \ll m$ and $\bW^\intercal \bW = I$. The perturbation vectors are generated by first sampling $n$-dimensional unit vectors ${ v}_b$ and project them with ${ u}_b = \bW { v}_b$.

\paragraph{Nonlinear Projection-Based Gradient Estimation.}
To search for the gradient representative subspaces more efficiently, we propose to perform the nonlinear projection-based gradient estimation.
In particular, we propose to leverage generative models given their expressive power. Here we mainly consider AE, VAE and GAN as examples.
There are two phases in \sysname: training and attacking. 
The detailed model structure and the training phase are described in Section~\ref{sec:estimator_structure}. 
Note that all these models typically have two components: an `encoder' and a `decoder' for AE and VAE, and a `generator' and a `discriminator' for the GAN. 
The `decoder' or `generator' projects a latent representation or random vector to sample space. The latent dimension is usually much lower than the sample space and this property is exactly desired. We unify the notations and denote both the `decoder' of AE and VAE and the `generator' part of GAN as `projection-based gradient estimator' in our following discussion.
The gradient estimator
is then used as the projection $\f:\bbR^n \to \bbR^m$ in the attacking phase. 
We first randomly sample unit latent vectors ${ v}_b$'s in $\bbR^n$, then the perturbation vectors generated as ${ u}_b = \f({ v}_b) \in \bbR^m$ are used in the gradient estimation,
yielding our gradient estimator as
    \vspace{-0.3em}
\begin{equation}
    \small
    \widetilde{\nablaS}(\xadv{t}) = \frac1B \sum_{b=1}^B \sgn \left(S\left(\xadv{t}+\delta \f({v}_b)\right) \right) \f({v}_b).
    \label{eq:our_gradient_estimation}
    \vspace{-0.3em}
\end{equation}
    \vspace{-1.3em}
\paragraph{Move along Estimated Gradient Direction.}
After getting the estimated gradient $\widetilde{\nabla S}$, the \boundaryimage $\xadv{t}$ is moved along that direction by:
\begin{equation}
    \label{eqn:move-grad}
    \hat{ x}_{t+1} = \xadv{t} + \xi_t \cdot \frac{\widetilde{\nabla S}}{\|\widetilde{\nabla S}\|_2},
\end{equation}
where $\xi_t$ is a step size chosen by searching with queries similar with HSJA~\citep{chen2020hopskipjumpattack}.

\paragraph{Project Back to Decision Boundary. }
In order to move closer to the \targetimage and enable the gradient estimation in the next iteration, we map the new adversarial image $\xadv{t}$ back to the decision boundary. This is achieved via binary search assisted by queries to find a suitable weight $\alpha_t$:
\begin{equation}
    \label{eqn:binary}
    \xadv{t+1} = \alpha_t \cdot { x}_{tgt} + (1-\alpha_t) \cdot \hat{ x}_{t+1}.
\end{equation}

\section{Projection-Based Gradient Estimation Analysis}
\label{sec:theory}
    To study the effectiveness of our projection-based gradient estimator in \Cref{eq:our_gradient_estimation} in terms of improving the estimation accuracy and reducing the number of queries, in this section, we theoretically analyze the expected cosine similarity between the estimated gradient $\tnablaS(\xadv{t})$ and the true gradient $\nablaS(\xadv{t})$ for the boundary-image $\xadv{t}$ at step $t$.
    
    \subsection{Generalized Gradient Estimator}
        \label{sec:generalized-gradient-estimator}
        We first formally define the gradient projection function $\f:\,\bbR^n \to \bbR^m$, which maps from the low-dimensional representative space $\bbR^n$ to the original high-dimensional space $\bbR^m$, where $n \le m$.
        Note that the projection function here could be \textit{nonlinear}, which is different from projections (linear transformations) defined in standard linear algebra\footnote{\url{https://en.wikipedia.org/wiki/Projection_(linear_algebra)}}.


        \begin{definition}[Generalized Projection-Based Gradient Estimator]
            \label{def:generalized-gradient-estimator}
            Suppose $\f(x_0)$ is a boundary-image, i.e., $S\left(\f(x_0)\right) = 0$,
            let $u_1,\,u_2,\,\dots,\,u_B$ be a subset of orthonormal basis of space $\bbR^n$ sampled uniformly~($B \le n$), we define
            \begin{equation}
                \widetilde{\nabla\f^\T \nablaS} := \dfrac{1}{B}
                \sum_{i=1}^{B} \sgn\left(S\left(\f(x_0 + \delta u_i)\right)\right) u_i.
                \label{eq:tildefs}
            \end{equation}
            Then, the \emph{generalized gradient estimator} for $\nablaS\left(\f(x_0)\right)$ is defined as
            \begin{equation}
                \tnablaS\left(\f(x_0)\right) := \nabla\f(x_0) \widetilde{\nabla\f^\T \nablaS}.
                \label{eq:tildeS}
            \end{equation}
            We abbreviate $\tnablaS\left(\f(x_0)\right)$ as $\tnablaS$ when there is no ambiguity.
        \end{definition}
        All the aforementioned gradient estimators are concretization of this generalized gradient estimator with different projections $\f$'s, including HSJA~\citep{chen2020hopskipjumpattack}, QEBA~\citep{cvpr2020QEBA} and our proposed NonLinear-BA.
        We defer the instantiations to Appendix~\ref{sec:adx-concretization}.
    
    
        \label{sec:lipschitz-smoothness-assumptions}
    
        We now impose local Lipschitz and local smoothness conditions on the projection $\f$ and the difference function $S$.
        \begin{definition}[Local $L$-Lipschitz]
            A (scalar or vector) function $f$ is called local $L$-Lipschitz around $x_0$ with radius $r$, if
            for any two inputs $x, x' \in \{x_0 + \delta:\,\|\delta\|_2 \le r \}$, 
            $$
                \frac{\|f(x) - f(x')\|_2}{\|x - x'\|_2} \le L.
            $$
        \end{definition}
        
        \begin{definition}[Local $\beta$-Smoothness]
            A (scalar or vector) function $f$ is called local $\beta$-smooth around $x_0$ with radius $r$, if (1)~$f$ is differentiable everywhere in region $\{ x_0+\delta:\, \|\delta\|_2 \le r \}$; and (2)~for any two inputs $x, x' \in \{x_0 + \delta:\,\|\delta\|_2 \le r\}$, 
                $$
                    \frac{\lambda_{\max} \left(\nabla f(x) - \nabla f(x')\right)}{\|x-x'\|_2} \le \beta,
                $$
            where $\lambda_{\max}(\mathbf{M})$ denotes the maximum eigenvalue of the matrix $\mathbf{M}$.
            Specifically, if $M$ is a vector, $\lambda_{\max}(M) = \|M\|_2$.
        \end{definition}
        The Lipschitz and smoothness definitions follow the general definitions in the literature~\citep{boyd2004convex,bubeck2015convex,hardt2016train}.
        Specifically, for a general function $f$ (e.g. $S$ or $\f$), when $\beta = 0$, the gradient $\nabla f$ is a constant in the region thus $f$ is locally linear.
        If $f$ is a neural network, there exists a local Lipschitz constant $L$~\citep{zhang2019recurjac}, and under generalized differential operator, there also exists a local smoothness constant $\beta$~\citep{nesterov2013introductory,clarke2008nonsmooth}.
        
        \paragraph{Assumptions.} Throughout the section, we assume the projection $\f$ is $L_\f$-Lipschitz and $\beta_\f$-smooth around $x_0$ with radius $\delta$, and the difference function $S$ is $L_S$-Lipschitz and $\beta_S$-smooth around $\f(x_0)$ with radius $L_\f\delta$.

    
        
        
        
    
    
        For the convenience of our analysis, we define the constant $\omega$ as such:
        \begin{definition}[Gradient Cosine Similarity Indicator $\omega$]
            \label{def:w}
            \begin{equation}
                \small
                \omega := \delta \left( 
                \dfrac{1}{2} \beta_\f L_S + 
                \dfrac{1}{2}\beta_S L_\f^2 + 
                \dfrac{1}{2} \delta \beta_\f \beta_S L_\f + 
                \dfrac{1}{8} \delta^2 \beta_\f^2 \beta_S 
                \right).
                \label{eq:def-w}
            \end{equation}
        \end{definition}
        The gradient cosine similarity indicator $\omega$ is an important quantity appearing in the cosine similarity lower bound.
        The $\delta$ in definition denotes the step size used in gradient estimation which is chosen according to HSJA~\citep{chen2020hopskipjumpattack}.
        \Linyi{Added}
    
        
        \begin{theorem}[General Bound for Gradient Estimator]
            Let $\f(x_0)$ be a boundary-image, i.e., $S\left(\f(x_0)\right) = 0$.
            The projection $\f$ and the difference function $S$ satisfy the assumptions in \Cref{sec:lipschitz-smoothness-assumptions}.
            Over the randomness of the sampling of orthogonal basis subset $u_1, u_2, \dots, u_B$ in $\bbR^n$ space,
            the expectation of cosine similarity between $\tnablaS\left(\f(x_0)\right)$~($\tnablaS$ for short) and $\nablaS\left(\f(x_0)\right)$~($\nablaS$ for short) satisfies
            %
            \begin{equation}
                \label{eq:main-bound}
                \small
                \begin{aligned}
                & \left( 2\left(1-\dfrac{\omega^2}{\|\nablafS\|_2^2} \right)^{(n-1)/{2}} - 1 \right)
                \dfrac{\|\nablafS\|_2}{L_\f \|\nablaS\|_2} 
                \sqrt{\frac B n}
                c_n\\
                & \hspace{2em} \le \, \bbE\, \cos\,\langle \tnablaS,\, \nablaS \rangle 
                \le \, 
                \dfrac{\|\nablafS\|_2}{l_\f \|\nablaS\|_2}
                \sqrt{\frac B n}
                c_n
                ,
                \end{aligned}
            \end{equation}

            where $\omega$ is defined in \Cref{def:w}, and we assume $\omega \le \|\nablafS\|_2$;
            $c_n \in (2/\pi,\,1)$ is a constant depended on $n$;
            $l_{\f} := \lambda_{\min}(\nabla\f(x_0))$.
            \label{thm:1}
        \end{theorem}
        
        \begin{proof}[Proof sketch.]
            Based on Taylor expansion, the projected length $\langle u_i,\,\nabla\f(x_0)^\T \nablaS\left(\f(x_0)\right) \rangle$ is correlated with $\sgn\left(S\left(\f(x_0 + \delta u_i)\right)\right)$, where $u_i$ is a sampled base vector.
            Concretely, when the projection smoothness is bounded, we show that when $|\langle u_i,\,\nabla\f(x_0)^\T \nablaS\left(\f(x_0)\right) \rangle|$ is larger than some threshold, it always has the same sign as $\sgn\left(S\left(\f(x_0 + \delta u_i)\right)\right)$.
            On the other hand, we study the distribution of $\langle u_i,\, \frac{\nabla\f(x_0)^\T \nablaS\left(\f(x_0)\right)}{ \|\nabla\f(x_0)^\T \nablaS\left(\f(x_0)\right)\|_2}  \rangle$, and derive the closed-form PDF for the distribution.
            \Linyi{Changed.}
            The cosine similarity between $\tnablafS$ and $\nablafS$ can be expressed as the sum of the products of these two terms over the basis:
            $\sgn\left(S\left(\f(x_0 + \delta u_i)\right)\right) \cdot \langle u_i,\, \frac{\nabla\f(x_0)^\T \nablaS\left(\f(x_0)\right)}{ \|\nabla\f(x_0)^\T \nablaS\left(\f(x_0)\right)\|_2}  \rangle$.
            Thus, we can derive the bounds for cosine similarity between $\tnablafS$ and $\nablafS$.
            From these bounds, we obtain the bounds for cosine similarity between $\tnablaS$ and $\nablaS$.
            We defer the detailed proof to Appendix~\ref{sec:adx-proof-of-cosine-similarity-bounds}.
        \end{proof}
        
        \begin{remark}
            This theorem provides the lower and upper bounds of the cosine similarity  between our generalized gradient estimator $\tnablaS$ and the true gradient $\nablaS$ for different models.
            As long as the Lipschitz and smoothness conditions in \Cref{sec:lipschitz-smoothness-assumptions} are satisfied, this bound is valid regardless of the concrete form of the projection $\f$ or how well the projection $\f$ is aligned with $S$,
            so we call it a `general bound'.
            We remark that smaller $\omega$ implies larger lower bound for the cosine similarity, which induces a tighter and improved gradient estimation. 
            Detailed discussions of these bounds are presented in \Cref{sec:better-nonlinear}.
        \end{remark}

    \subsection{Gradient Estimation Based on Different Gradient Projections}
        \label{sec:better-nonlinear}
    
        From Theorem~\ref{thm:1}, one may think that linear projection is better than nonlinear one since when the $\nabla \f$ is the same,
        linear projection implies $\beta_\f = 0$, which leads to smaller $\omega$ and higher lower bound of the gradient cosine similarity.
        However, this lower bound is applied to all models satisfying the Lipschitz and smoothness condition.
        In fact, there exists nonlinear projection $\f$ leading to higher cosine similarity lower bound. 
        (We will focus on the discussion of lower bound below, since the upper bound is irrelevant with $\beta_\f$ from Theorem~\ref{thm:1}, meaning linear and nonlinear projections would share the same upper bound.)
        
        \paragraph{Linear Projection.}
            \label{sec:linear-case}
            First, let us consider the linear projection $\f$.
            Throughout the text, we use $\lambda_{\max} (\bM)$ to denote the largest eigenvalue of matrix $\bM$, and $\lambda_{\min} (\bM)$ the smallest eigenvalue of matrix $\bM$. 
            
            \begin{corollary}[Linear projection Bound, informal]
                Under the same setting of \Cref{thm:1} with additional condition that projection $\f$ is locally linear around $x_0$ with radius $\delta$ and $L_\f := \lambda_{\max} (\nabla \f(x_0))$,
                the expectation of cosine similarity satisfies \Cref{eq:main-bound} with
                \begin{equation}
                    \omega := \dfrac{1}{2}\delta \beta_S L_\f^2.
                    \label{eq:linear-case-omega}
                \end{equation}
                We assume that $\omega \le \|\nablafS\|_2$.
                $c_n \in (2/\pi,\,1)$ is a constant depended on $n$.
                \label{cor:1}
            \end{corollary}
            
            
            \begin{remark}
                We defer the formal statement to \Cref{sec:adx-comparison-of-different-gradient-estimators}.
                This is a direct application of Theorem \ref{thm:1} with $\beta_\f = 0$ due to linearity.
                The main difference between the corollary and Theorem \ref{thm:1} is in $\omega$, where the general $\omega$ in \Cref{eq:def-w} is altered by \Cref{eq:linear-case-omega}.
                Furthermore, if $S$ is also locally linear, then $\beta_S = 0$ and hence $\omega = 0$, which closes the gap between lower bound and upper bound and implies that the gradient estimation is pretty precise~(cosine similarity between estimated and true gradient is $c_n \in (2/\pi,\,1)$).
                In addition, \citet{cvpr2020QEBA} provide a cosine similarity bound based on projection taken the form of orthogonal transformation, which can be recovered from \Cref{cor:1} by setting $L_\f = l_\f = 1$, $\beta_\f = 0$, and replacing $\|\nablaS^\T \nabla \f\|_2$ with $\|\nablaS\|_2$ in \Cref{eq:main-bound} due to the randomness of projection $\f$.
            \end{remark}

            
            
        
        \paragraph{Nonlinear Projection.}
            For nonlinear projection, we have the following theorem.
            
            \begin{theorem}[Existence of Better Nonlinear Projection, informal]
                Under the same setting of \Cref{cor:1}, there exists a \emph{nonlinear} projection $\f'$ satisfying the assumptions in \Cref{sec:lipschitz-smoothness-assumptions}, with $\f'(x_0) = \f(x_0)$ and $\nabla \f'(x_0) = \nabla \f(x_0)$, such that
                the expectation of cosine similarity between $\tnablaS\left(\f'(x_0)\right)$~($\tnablaS$ for short) and $\nablaS\left(\f'(x_0)\right)$~($\nablaS$ for short) satisfies
                \Cref{eq:main-bound} with
                \begin{equation}
                    \omega := \dfrac{1}{2}\delta \beta_S L_\f^2 - \dfrac{1}{5} \beta_\f \beta_S \delta^2 L_\f < \dfrac{1}{2} \delta \beta_S L_\f^2.
                    \label{eq:nonlinear-case-omega}
                \end{equation}
                We assume that $\omega \le \|\nablafS\|_2$.
                $c_n \in (2/\pi,\,1)$ is a constant depended on $n$.
                \label{thm:2}
            \end{theorem}
            
                
            \begin{proof}[Proof Sketch]
                We prove by construction---we construct the nonlinear projection $\f'$ explicitly from $\f(x_0)$, $\nabla\f(x_0)$ and the difference function $S$.
                Comparing with the linear projection $\f$, the $\f'$ is allowed to have curvature since $\beta_\f > 0$.
                The constructed $\f'$ exploits this curvature to cancel out the impreciseness caused by large $\nabla \f$ and thus reduces the actual $L_\f$.
                After showing that $\f'$ satisfies the assumptions in \Cref{sec:lipschitz-smoothness-assumptions}, we derive its cosine similarity bound with corresponding $\omega$.
                \Linyi{Changed}
            \end{proof}
            
            \begin{remark}
                This theorem shows that if the difference function $S$ is nonlinear~(i.e., $\beta_S > 0$), for any linear projection $\f$,
                we can define a particular nonlinear projection $\f'$ that aligns with $\f$ in both zeroth order and first order.
                Compared with  $\omega$ of $\f$~(\Cref{eq:linear-case-omega}),  $\omega$ of $\f'$~(\Cref{eq:nonlinear-case-omega}) is thus reduced.
                Then, \Cref{eq:main-bound} implies that using $\f'$, the cosine similarity between the estimated gradient and the true gradient can be improved.
            \end{remark}
            The formal statement and the full proof are deferred to Appendix~\ref{sec:adx-existence-better-nonlinear-projection}.
    
        \label{sec:theory-implications}
       Based on the above results, we aim to further analyze two research questions.
       
       \textbf{Can we estimate gradients from a projected low-dimension subspace? }\\
        The answer is yes. 
        According to \Cref{thm:1,thm:2}, the cosine similarity between true gradient and estimated gradient depends on the ratio $B / n$, rather than only the subspace dimension $n$.
        Now we assume the number of queries $B$ is equal to the subspace dimensionality $n$.
        We can observe a \emph{sufficient condition} for good cosine similarity: $\|\nabla \f^\T \nablaS\|_2 / \|\nablaS\|_2$ is large, i.e., \emph{the gradient of projection function $\nabla\f$ and the gradient of difference function $\nablaS$ align well.}
        We remark that the condition is independent with subspace dimensionality $n$, and is checkable when the gradient of the victim model
        is known.
        Specifically, when $\|\nabla \f^\T \nablaS\|_2 / \|\nablaS\|_2$ achieves its maximum $L_\f$, the cosine similarity lower bound becomes
        \begin{equation}
            \left( 2\left(1-\dfrac{\omega^2}{\|\nablafS\|_2^2} \right)^{(n-1)/2} - 1 \right) c_n,
            \label{eq:low-dimension-good-lower-bound}
        \end{equation}
        where $\omega$ could be either defined by \Cref{def:w} for general projection or defined by \Cref{thm:2} for good nonlinear projection.
        \Linyi{Changed the explanation for this research question.}
        
        We can clearly observe that smaller $\omega$ leads to better lower bound for cosine similarity, and when $\omega = 0$ the cosine similarity becomes $c_n \in (2/\pi,\,1) \approx (.637,\,1)$ which is high.
        To verify this negative correlation between the $\omega$ values and the cosine similarity measurements, we conduct empirical experiments in Appendix~\ref{sec:correlation_omega_cos}.
        
        To better analyze the lower bound, we further lower bound \Cref{eq:low-dimension-good-lower-bound} as such:
        \begin{align}
            & \left( 2\left(1-\dfrac{\omega^2}{\|\nablafS\|_2^2} \right)^{(n-1)/2} - 1 \right) c_n
            \tag{\ref{eq:low-dimension-good-lower-bound}} \\
            \ge & \left(
                1 - (n-1) \dfrac{\omega^2}{\|\nablafS\|_2^2} \right) c_n. \label{eq:low-low-bound}
        \end{align}
        Since the dimensionality $n$ of the sampling space is small, and $\omega = \Theta(\delta)$ where $\delta$ is the step size and is also small, we can observe that with the projection, the similarity lower bound is non-trivial, i.e., \emph{we can estimate gradients from a projected low-dimension space.}
        
        When step size $\delta$ in $\omega$ approaches $0$, the $\omega$ approaches $0$.
        If $S$ and $\f$ are both locally linear, with $\beta_S = \beta_\f = 0$ we again have $\omega = 0$ and the cosine similarity also becomes $c_n$, which implies that we can achieve high cosine similarity using just linear projection if the difference function $S$ is locally linear.
        \Huichen{New content for AISTATS}
        \Linyi{Move a sentence added by Huichen to preceding paragraph.}
        
        On the other hand, we inspect the relation between cosine similarity bound and the number of queries $B$.
        As shown in \Cref{eq:main-bound}, both the lower and upper bound are in $\Theta(\sqrt B)$ with respect to number of queries.
        In other words, to achieve a cosine similarity $s$, one need to perform $\Theta(s^2)$ number of queries.
        As a result, moderate cosine similarity requires a small number of queries but high cosine similarity requires much more, and it is indeed better to leverage the reduced subspace dimension $n$.
        We formalize the query complexity analysis as below and defer the proof and detail discussion to \Cref{sec:adx-proof-of-cosine-similarity-bounds}.
        \begin{corollary}[Query Complexity]
            Given the projection $\f$ and the difference function $S$, to achieve expected cosine similarity $\bbE \langle \nablaS(\f(x_0)),\,\tnablaS(\f(x_0)) \rangle = s$, the required query number $B$ is in $\Theta(s^2)$.
            \label{cor:query-complexity}
        \end{corollary}
        
        \textbf{How do different projections affect the gradient estimation quality?} \\
        The above analysis allows us to compare projection-based gradient estimators in different boundary attacks directly. 
        We instantiate the general bound in \Cref{thm:1} for HSJA and QEBA respectively, which shows that QEBA is significantly better than HSJA as it achieves the same cosine similarity with much fewer queries.
        For NonLinear-BA, \Cref{thm:2} points out the possibility and a checkable sufficient condition where NonLinear-BA could be better than corresponding linear projection including HSJA and QEBA, in terms of providing higher lower bound of cosine similarity.
        We further present another sufficient condition in Appendix~\ref{sec:adx-comparison-of-different-gradient-estimators} under which NonLinear-BA achieves higher lower bound.
        \emph{In a nutshell, the nonlinear projection which outperforms linear projection is not rare, however, the efficient search algorithm for it with theoretical guarantees is still open.}
        Thus, in NonLinear-BA we heuristically train popular neural network structures (e.g. AE and GANs) to function as nonlinear projections proxies, so as to analyze their ability of reducing query complexity and achieving precise gradient estimation.
        The detailed results and discussions can be found in Appendix~\ref{sec:adx-comparison-of-different-gradient-estimators}.
        We further discuss potential ways of improving the gradient estimation inspired from the theoretical analysis in Appendix~\ref{sec:adx-improve-gradient-estimation}.

\begin{figure*}[t!]
    \centering
    \includegraphics[width=\textwidth]{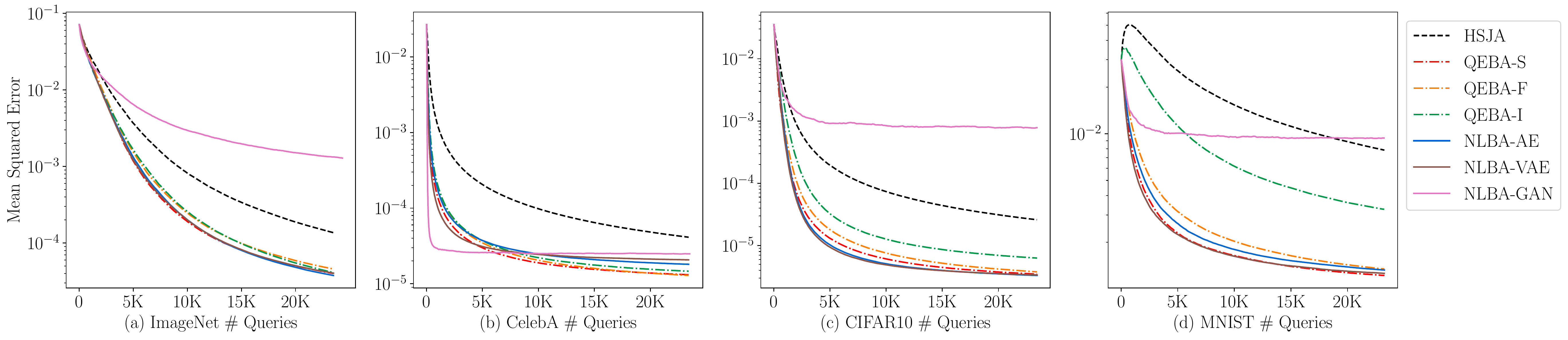}
    \caption{\small The perturbation magnitude based on different queries for attacks on diverse datasets.}
    \label{fig:mean_mse_nq}
\end{figure*}

\vspace{-1mm}
\section{Experiments}
\vspace{-1mm}
In this section, we conduct extensive experiments to evaluate the performance of different boundary blackbox attacks, and show that (1) with the nonlinear and linear projection-based gradient estimation methods, blackbox attacks can achieve better performance compared with the state-of-the-art baselines;
(2) both the nonlinear and linear projection-based gradient estimation methods demonstrate their own advantages under certain conditions.
In addition, we also show the high blackbox attack performance against commercial face recognition APIs for \sysname.

\label{sec:exp}
\subsection{Experimental Setup}
\paragraph{Target Models.}
We use both offline models on ImageNet, CelebA, CIFAR10 and MNIST datasets, and commercial online APIs, as target models following~\citet{cvpr2020QEBA}.
For offline models, on ImageNet, we use a pretrained ResNet-18 as the target model.
On CelebA, a pretrained ResNet-18 is fine-tuned to perform classification on attributes as the target model. The most balanced attribute (e.g., `Mouth\_Slightly\_Open') is chosen to enhance benign model performance.
On CIFAR10 and MNIST datasets, we scale up the input images to $224 \times 224$ with linear interpolation to demonstrate the query reduction for high-dimensional input space. Fine-tuned ResNet-18 models are used as target models.
The benign target model performance is shown in Appendix~\ref{sec:target_model_performance}.
For commercial online APIs, we use the `Compare' API from MEGVII Face++~(\citeyear{facepp-compare-api}). Given two images, the API returns a confidence score of whether they are of the same person. Following~\citet{cvpr2020QEBA}, we convert the confidence score to a discrete prediction by taking scores greater than or equal to $50\%$ as `same person', and vice versa. 
The implementation details are discussed in Appendix~\ref{sec:target_model_details}. 

\paragraph{Nonlinear Projection Models.}
The nonlinear projection models are trained on image gradient dataset. The goal is to train the projection models to project from low-dimensional random vectors to higher-dimensional gradient space so that the projected vectors mimic the distribution of the gradient of the target model. The more aligned the projected vectors are with the ground truth gradient vectors, the more effective the Monte Carlo estimation would be.
Image gradients are generated using PyTorch's~\citep{NEURIPS2019_9015} automatic differentiation functions on five reference models for each dataset.
The details including model architectures and training parameters are described in Appendix~\ref{sec:estimator_structure}. The benign accuracy for the reference models are shown in Appendix~\ref{sec:ref_model_performance}.
Note that although in the offline experiments, we use data from the same distribution to train the reference models with different architectures from the target model, this is not a necessary condition: for the attacks on commercial online APIs, we do not have any information about the training dataset or the model structure, and we use the same projection models trained on ImageNet dataset to attack the face recognition APIs. The experimental results in Section~\ref{sec:atk_api} and Appendix~\ref{sec:api_qualitative} show that the projection-based boundary blackbox attack works well despite the mismatch of training data distributions between the target model and the projection model.

\paragraph{Evaluation Metrics.}

\Huichen{modified for post-conf cr}
We evaluate \sysname and compare with the baseline methods based on two standard evaluation metrics: (1) the average magnitude of perturbation at each step, as indicated by the mean squared error (MSE) between the optimized adversarial example and \targetimage; (2) the attack success rate defined as reaching a specified MSE threshold.

\begin{figure*}[t!]
    \centering
    \includegraphics[width=\textwidth]{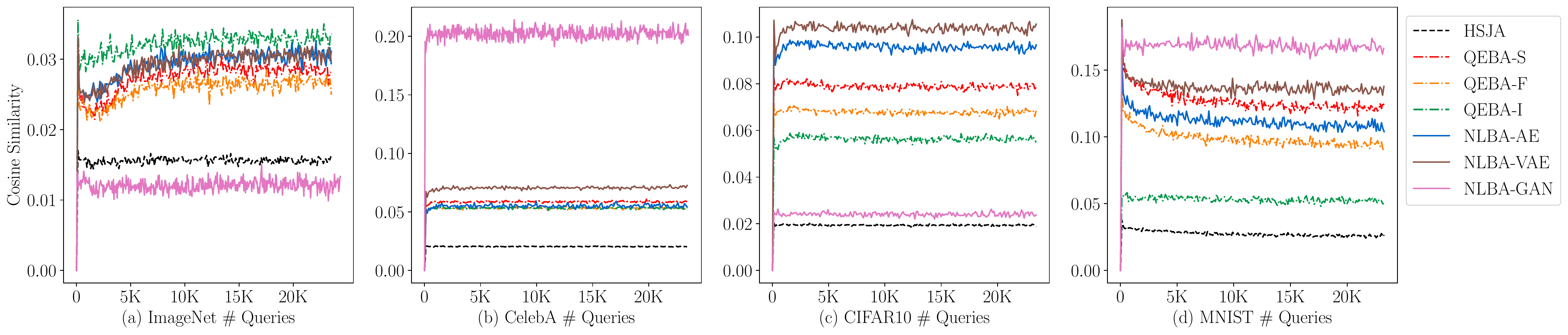}
    \caption{\small The cosine similarity between the estimated and true gradients with respect to query numbers for attacks on diverse datasets.}
    \label{fig:mean_cos_sim_nq}
    \vspace{-0.5cm}
\end{figure*}

\subsection{Blackbox Attack Performance Against Offline Models}
\Huichen{modified for post-conf cr}
Figure~\ref{fig:mean_mse_nq} shows the attack performance of different approaches in terms of the perturbation magnitude (MSE) between the generated sample and the \targetimage. 
The attack success rates are shown in Figure~\ref{fig:mean_success_nq} in appendix. The `\sysname' is denoted as `NLBA' in figures.
All the results are averaged over $50$ randomly sampled pairs of correctly classified images from the corresponding datasets.

The \sysname with three projection methods exhibit different patterns on the four datasets. 
\sysname-AE and \sysname-VAE are the most consistent across various datasets. They achieve significantly better performance compared with baseline HSJA, and outperform QEBA in many cases. 
The \sysname-GAN method, on the other hand, is less stable. For attribute classification model on CelebA, where the model's ground truth gradients have a simpler pattern, it is significant better than the other methods with very few queries (Fig~\ref{fig:mean_mse_nq}(b)). 
For example, on CelebA dataset, the \sysname-GAN method takes only a few hundred of queries to get to a distance smaller than $10^{-4}$ while other projection-based methods take more than one thousand, and the HSJA baseline takes over $10$ thousand queries to get to the same distance magnitude. 
\Huichen{modified for post-conf cr}
On the other hand, when the gradient patterns are more complex, the \sysname-GAN method fails to keep reducing the MSE after some relatively small number of queries and converges to a bad local optima.
We conjecture this is due to the instability of GAN training, and it would be interesting future work to develop in-depth understanding about the properties of nonlinear GAN-based projection.

Our theoretical analysis and conclusions are well supported by the fact that nonlinear and linear projection models have advantages over each other under various different scenarios, and that except for the unstable \sysname-GAN case, they both outperform the HSJA baseline which does not have a dimension reduction module via projection.


\begin{figure}[t]
\centering
\begin{subfigure}[t]{.3\linewidth}
  \centering
  \includegraphics[width=\linewidth]{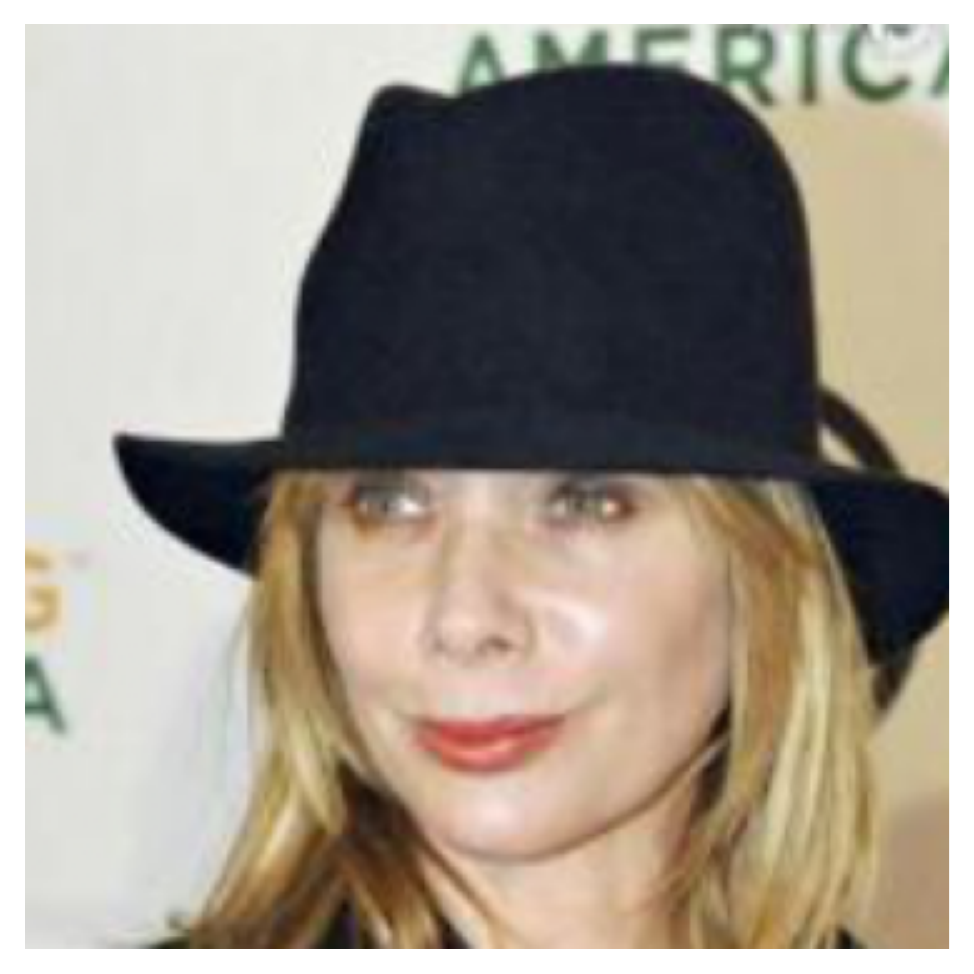}
  \caption{\Sourceimage}
  \label{fig:case_study_celeba_src_img}
\end{subfigure}
\hspace{.1\linewidth}
\begin{subfigure}[t]{.3\linewidth}
  \centering
  \includegraphics[width=\linewidth]{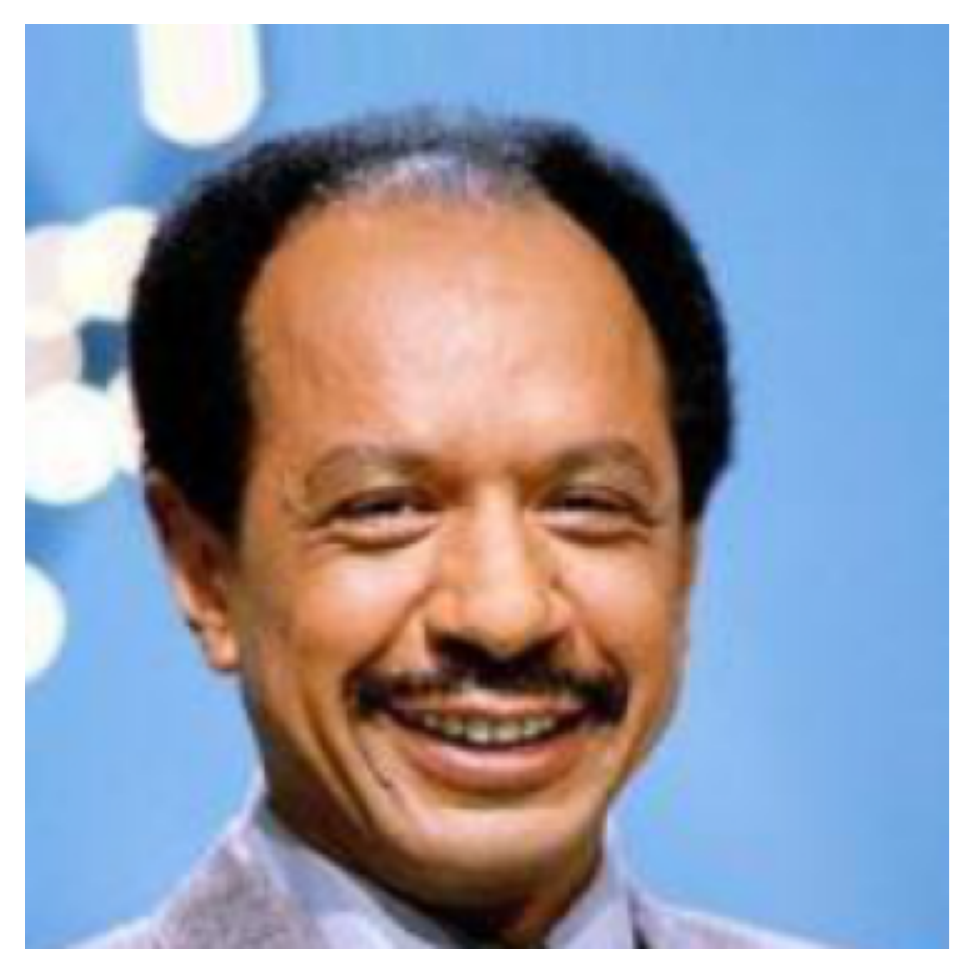}
  \caption{\Targetimage}
  \label{fig:case_study_celeba_tgt_img}
\end{subfigure}
\caption{\small The source and target images for the qualitative case study on CelebA dataset. The \textit{attribute-of-interest} is `Mouth\_Slightly\_Open', which is labeled as `False' for the \sourceimage while `True' for the \targetimage.}
\label{fig:case_study_celeba_src_tgt_imgs}
\end{figure}

\begin{figure}[t]
    \centering
    \includegraphics[width=0.9\linewidth]{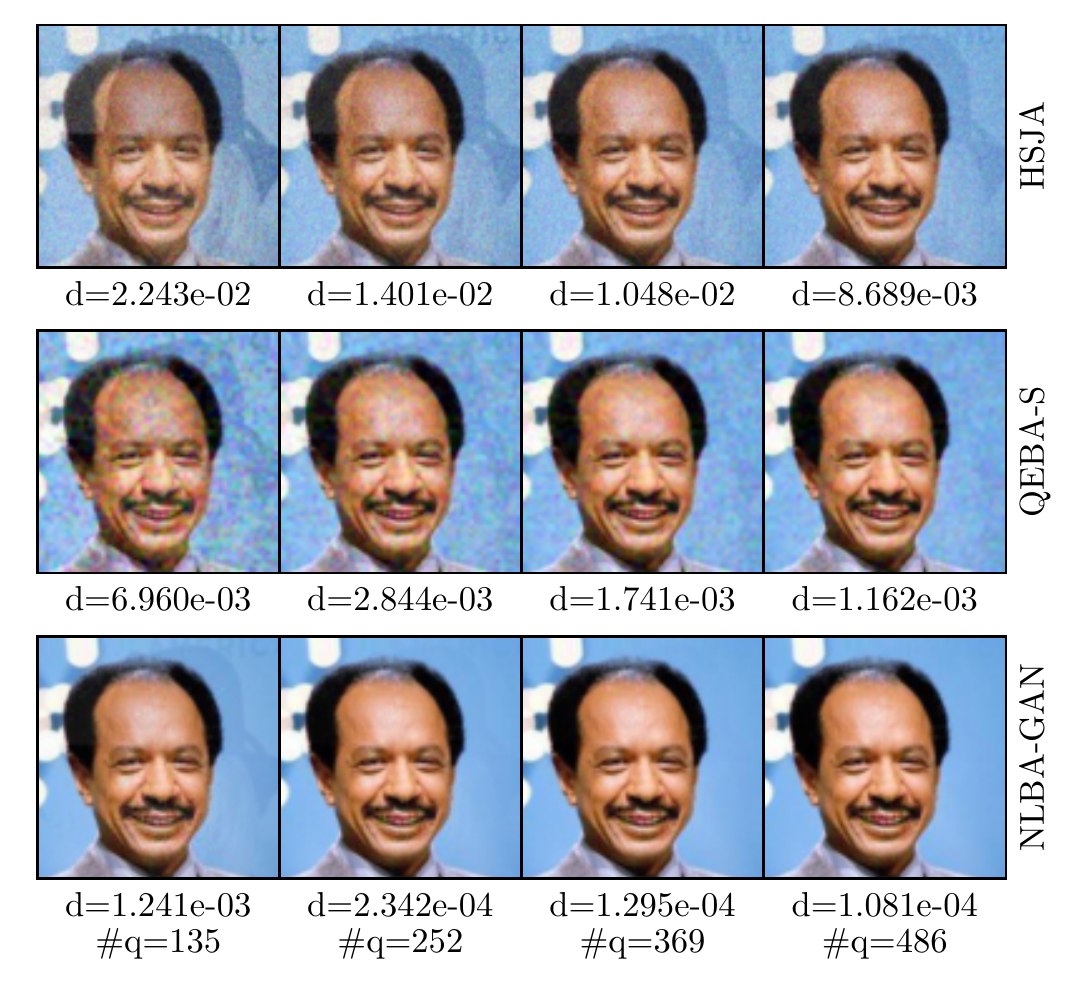}
    \caption{\small The attack process of different methods on CelebA dataset under different query numbers. $d$  denotes the perturbation magnitude of the generated adversarial example with respect to the \targetimage. $\#q$ denotes the number of queries.}
    \label{fig:casestudy_early_atk_process}
\end{figure}



\paragraph{Verification for Gradient Cosine Similarity and Attack Performance.}
\Linyi{Propose to change this paragraph as follows for saving some space and providing better explanation}
\Huichen{modified for post-conf cr}
To verify that the cosine similarity of gradients indeed reflects the blackbox attack performance, we plot the gradient cosine similarity corresponding to different queries in Figure~\ref{fig:mean_cos_sim_nq}. It is clear that the blackbox attack performance highly correlates with the cosine similarity positively: when the cosine similarity is high, the attack performance is better and can converge to a smaller MSE faster.

\Cref{thm:1,thm:2} suggest that smaller $\omega$~(\Cref{def:w}) leads to higher cosine similarity between the estimated and true gradients.
To verify it, we use an alternative method to evaluate the effects of $\omega$ approximately, and show that cosine similarity is strongly  correlated with $\omega$ as proved. Details can be found in Appendix~\ref{sec:correlation_omega_cos}.



\paragraph{Case Study: Attack Performance at an Early Stage for CelebA Dataset.}
We perform a qualitative case study to show the effectiveness of  \sysname. The \sourceimage and \targetimage are shown in Figure~\ref{fig:case_study_celeba_src_tgt_imgs}. The goal is to generate an \advimage that has a small distance from the \targetimage with an open mouth, to be mis-recognized as `Mouth\_Slightly\_Open=False' by the victim model. The attack results at early attack stages with fewer than $500$ queries are shown in Figure~\ref{fig:casestudy_early_atk_process}. We present the results for the baseline HSJA, as well as the baseline QEBA and the proposed \sysname with the best performance among their variations (e.g., QEBA-S and \sysname-GAN).
The full case study results for all the attack methods are shown in Figure~\ref{fig:celeba_3_all_attack_process_early} in Appendix~\ref{sec:celeba_3_all_attack_process_early}. It is obvious that with less than $150$ queries, the \sysname-GAN method already achieves similar or even better performance compared with QEBA-S and HSJA with about $500$ queries; with about $250$ queries, the quality of the \advimage produced by \sysname-GAN is so high that visually it is almost indistinguishable compared with the \targetimage.
More qualitative case studies of the attacks can be found in Appendix~\ref{sec:offline_qualitative}.

\subsection{Blackbox Attack Performance Against Commercial APIs}
\begin{figure}[t!]
    \centering
    \includegraphics[width=\linewidth]{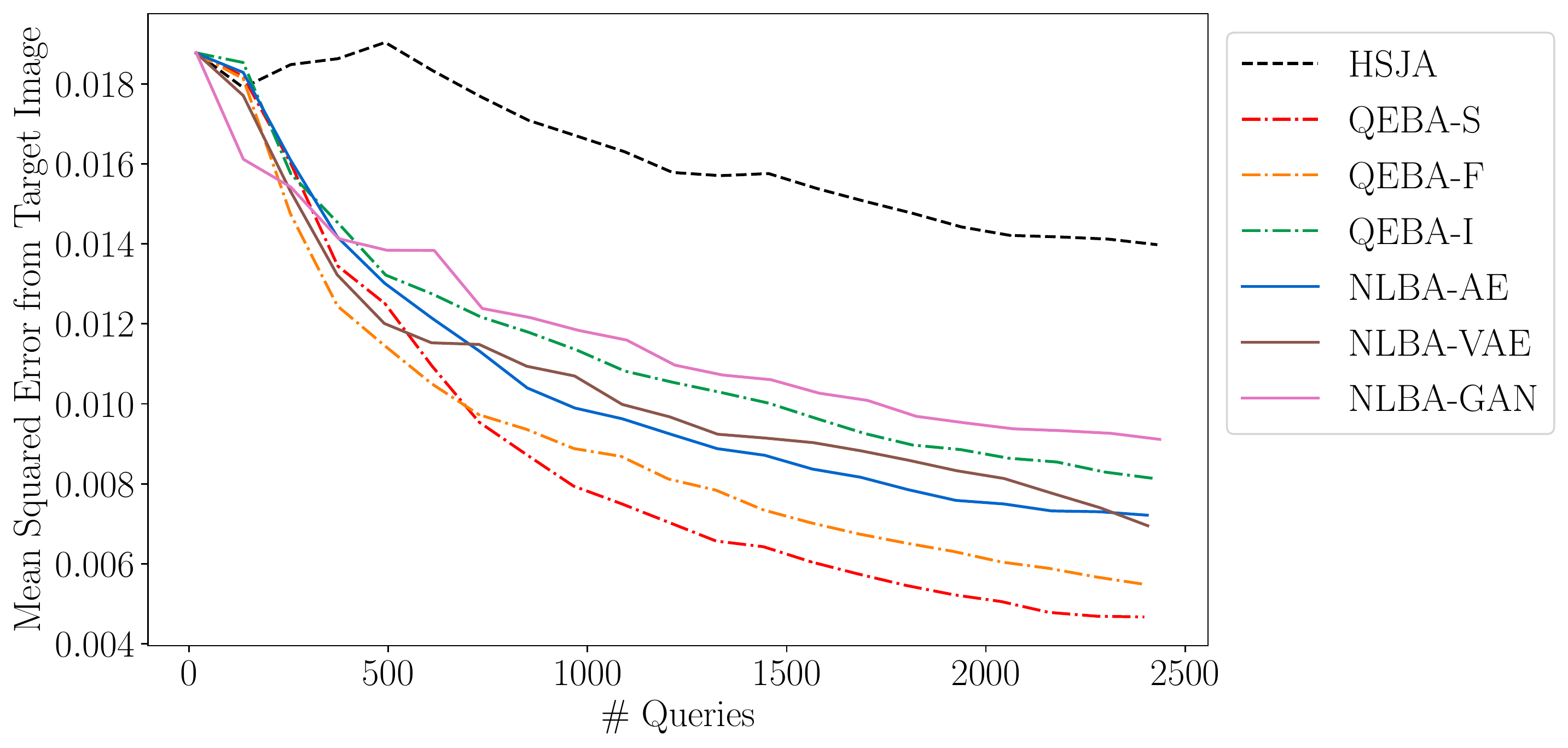}
    \vspace{-0.5cm}
    \caption{\small The perturbation magnitude based on different queries against Face++ `Compare' API.}
    \label{fig:api_facepp_means}
    \vspace{-0.5cm}
\end{figure}
\label{sec:atk_api}
To demonstrate the practicality of the proposed \sysname, we perform the blackbox attack against real-world online commercial APIs.
Figure~\ref{fig:api_facepp_means} shows the MSE between the \advimage and the \targetimage with different numbers of queries. The attack success rate is always $100\%$ during the whole process.
The results are averaged over $40$ randomly sampled CelebA face image pairs. The image pairs are the same for each of the seven methods for fair comparison. 
From Figure~\ref{fig:api_facepp_means}
\Huichen{modified post-conf}
it is clear that all the six gradient projection-based methods including both linear and non-linear projections are better than the baseline HSJA in terms of the MSE under the same number of queries, and the nonlinear projection converges faster while observes slightly higher perturbation magnitude.
The qualitative results of case studies are shown in Appendix~\ref{sec:api_qualitative}.


    
    

\section{Conclusion}
    We provide the first theoretic analysis framework for projection-based gradient estimation. We then propose \sysname, a nonlinear projection-based gradient estimation approach for query-efficient boundary blackbox attack.
    We theoretically show nontrivial cosine similarity bounds for a group of projection based gradient estimation approaches and analyze the properties of different projections.
    We evaluate the efficiency of \sysname with extensive experiments against both offline ML models and commercial online APIs.
    
\noindent\textbf{Acknowledgement}
This work is partially supported by Amazon research award.

\newpage

\newpage

{
\bibliographystyle{plainnat}
\bibliography{bibliography}
}




\newpage
\onecolumn
\appendix

\section{Instantiations of Generalized Gradient Estimator}
    \label{sec:adx-concretization}

    As discussed in \Cref{sec:generalized-gradient-estimator}, the generalized gradient estimator in \Cref{def:generalized-gradient-estimator} unifies the boundary gradient estimator in HSJA~\citep{chen2020hopskipjumpattack}, QEBA~\citep{cvpr2020QEBA}, and our NonLinear-BA.
    In this section we discuss the instantiations of them in detail.
    
    In the generalized gradient estimator, the $u_1,\,u_2,\,\dots,u_B$ are a sampled subset of orthonormal basis, whereas in practice, all these methods only uniformly sample normalized vectors for efficiency concern.
    As implied by \Cref{lemma:1}, when $n$ becomes large, $\langle u_i,\,v \rangle$'s PDF is highly concentrated at $x = 0$, implying that with high probability the sampled normalized vectors are close to orthogonal.
    Therefore, the orthonormal basis sampling can be approximated by normalized vector sampling.
    With this mindset, we express each gradient estimator using generalized gradient estimator~(\Cref{def:generalized-gradient-estimator}).
    
    \paragraph{HSJA.}
    At a boundary-image $\xadv{t}$, the HSJA gradient estimator~\citep{chen2020hopskipjumpattack} is 
    $$
        \widetilde{\nablaS(\xadv{t})} = \frac1B \sum_{b=1}^B \sgn \left(S\left(\xadv{t}+\delta {u}_b\right) \right) { u}_b.
    $$
    We define the projection $\f: \bbR^m \to \bbR^m$ as an identical mapping.
    The gradient estimator reduces to
    \begin{equation}
        \tnablaS(\f(x_0)) = \frac1B \sum_{i=1}^B \sgn\left( S\left( \f(x_0 + \delta u_i) \right) \right) \f(u_i) = \frac1B \sum_{i=1}^B \sgn\left( S\left( x_0 + \delta u_i \right) \right) u_i,
        \label{eq:hsja-estimator}
    \end{equation}
    which is exactly the HSJA gradient estimator.
    
    \paragraph{QEBA.}
    At a boundary-image $\xadv{t}$,
    the QEBA gradient estimator~\citep{cvpr2020QEBA} is 
    $$
        \widetilde{\nablaS(\xadv{t})} = \frac1B \sum_{b=1}^B \sgn \left(S\left(\xadv{t}+\bW\delta {u}_b\right) \right) \bW{ u}_b.
    $$
    The $\bW \in \bbR^{m\times n}$ is an orthogonal matrix.
    We define the projection $\f: \bbR^n \to \bbR^m$ by $\f(v) = \bW v + x_0$.
    Notice that $\f(0) = x_0$ is a boundary-image of difference function $S$.
    At the origin, the \Cref{eq:tildefs} becomes
    $$
        \widetilde{\nabla\f^\T \nablaS} = \frac1B \sum_{i=1}^B \sgn\left(S\left( \f(\delta u_i) \right) \right) u_i = \frac1B \sum_{i=1}^B \sgn\left(S\left( x_0 +  \delta \bW u_i \right) \right) u_i,
    $$
    and the gradient estimator becomes
    \begin{equation}
        \tnablaS(\f(0)) = \bW \widetilde{\nabla\f^\T \nablaS} = \frac1B \sum_{i=1}^B \sgn\left(S\left( x_0 +  \delta \bW u_i \right) \right) \bW u_i,
        \label{eq:qeba-estimator}
    \end{equation}
    which is the QEBA gradient estimator.
    
    \paragraph{\sysname.}
    In \sysname, a nonlinear projection $\f$ is already trained.
    The gradient estimation uses \Cref{eq:our_gradient_estimation}.
    To bridge the gap between \Cref{eq:our_gradient_estimation} and the generalized gradient estimator in \Cref{eq:tildefs}, we define a new projection $\mathbf{g}$ such that $\mathbf{g}(v) = x_0 + \|v\| \f\left(v / \|v\|\right)$.
    We assume that $\f$ is highly linear within the $L_2$ ball $\{r:\,\|r\| \le 1\}$.
    Therefore, $\nabla \mathbf{g}(0)$ exists, and for normalized vector $u_i$, $\mathbf{g}(u_i) - \mathbf{g}(0) \approx \nabla \mathbf{g}(0) u_i$.
    Notice that $\mathbf{g}(u_i) = x_0 + \f(u_i)$ and $\mathbf{g}(0) = x_0$, so $\f(u_i) \approx \nabla \mathbf{g}(0) u_i$.
    
    We apply generalized gradient estimator with projection $\mathbf{g}$ at the boundary-image $\mathbf{g}(0) = x_0$:
        \begin{align}
            \tnablaS(\mathbf{g}(0)) & = \nabla\mathbf{g}(0) \left( \frac1B \sum_{i=1}^B \sgn\left(S\left(\mathbf{g}(\delta u_i)\right)\right) u_i \right) = \frac1B \sum_{i=1}^B \sgn\left(S\left(x_0 + \delta \f(u_i)\right)\right) \nabla \mathbf{g}(0) u_i \label{eq:adx-A-nonlinearBA-precise} \\
            & \approx \frac1B \sum_{i=1}^B \sgn\left(S\left(x_0 + \delta \f(u_i)\right)\right) \f(u_i), \label{eq:adx-A-nonlinearBA-approx}
        \end{align}
    where the \Cref{eq:adx-A-nonlinearBA-approx} is the NonLinear-BA gradient estimator in \Cref{eq:our_gradient_estimation}.
    We implement \sysname gradient estimator by \Cref{eq:adx-A-nonlinearBA-approx} instead of the precise \Cref{eq:adx-A-nonlinearBA-precise} to avoid gradient computation and improve the efficiency.
    
    Notice that in all these methods we perform boundary attack iterations in the raw input space.
    However, for the gradient estimation, QEBA and \sysname use low dimension space while HSJA uses raw input space.
    To reflect the boundary point $x_0$ found in raw input space, in QEBA and NonLinear-BA, the projection is defined as the difference from the bounadry image $x_0$, i.e., $\f(0) = x_0$ and the gradient estimation is for $\f(0)$.
    In this way, we circumvent the possible sparsity of the boundary-images in low dimension space.
    
    In summary, all these gradient estimators are instances of generalized gradient estimator in \Cref{def:generalized-gradient-estimator}.
    Moreover, we can observe that HSJA and QEBA use linear projection, and \sysname permits nonlinear projection.

\section{Proof of Cosine Similarity Bounds}

    \label{sec:adx-proof-of-cosine-similarity-bounds}

    In this section, we prove the universal cosine similarity bounds as shown in \Cref{thm:1}.
    The proof is derived from careful analysis of the distribution of randomly sampled orthonormal basis, combining with Taylor expansion and breaking down the cosine operator.
    

    \begin{lemma}
        Let $u_1, u_2, \dots, u_B$ be randomly chosen subset of orthonormal basis of $\bbR^n$~($B \le n$).
        Let $v$ be any fixed unit vector in $\bbR^n$.
        For any $i \in [B]$,
        define $a_i := \langle u_i, \, v \rangle$.
        Then each $a_i$ follows the distribution $p_a$ with PDF
        \begin{equation}
            p_a(x) :=  \dfrac{(1-x^2)^{({n-3})/{2}}}{\cB\left( \frac{n-1}{2}, \frac{1}{2} \right)},\quad x \in [-1,\, 1],
        \end{equation}
        where $\cB$ is the Beta function.
        \label{lemma:1}
    \end{lemma}
    
    \begin{remark}
        Lemma~\ref{lemma:1} shows the distribution of projection of orthonormal base vector on arbitrary normalized vector.
        Later we will apply the lemma to any normalized vector.
    \end{remark}
    
    \begin{proof}[Proof of \Cref{lemma:1}]
    
        Since $u_i$ is the randomly chosen orthonormal base vector, the marginal distribution of each $u_i$ is the uniform distribution sampled from $(n-1)$-unit sphere.
        As a result, for any unit vector $v$, the distribution of $\langle u_i,\, v\rangle$ should be the same.
        Consider $e_1 = (1,\,0,\,0,\,\dots,\,0)^\T$,
        \begin{equation}
            a_i = \langle u_i, e_1 \rangle = {u_i}_1.
            \label{lem:eq-1}
        \end{equation}
        Now consider the distribution of ${u_i}_1$, i.e., the first component of $u_i$.
        We know that ${u_i}_1 = {x_1}/{\sqrt{x_1^2 + \cdots + x_n^2}}$ where each $x_i \sim \cN(0,1)$ independently~\citep{muller1959note,marsaglia1972choosing}.
        Therefore, let $X \sim \cN(0,1)$, and $Y \sim \chi^2(n-1)$, ${u_i}_1 = {X}/{\sqrt{X^2 +Y}}$.
        Denote $f(x)$ to the PDF of ${u_i}_1$, from calculus, we obtain
        \begin{equation}
            f(x) = \int_{0}^{\infty} \dfrac{y^{\frac{n-1}{2}-1} \exp\left(-\frac{y}{2}\right)}{2^{\frac{n-1}{2}} \Gamma\left(\frac{n-1}{2}\right)} \cdot \dfrac{1}{\sqrt{2\pi}} \exp\left( - \dfrac{x^2y}{2(1-x^2)} \right) \dfrac{\sqrt{y}}{(1-x^2)^{-3/2}} \diff y 
            = \dfrac{(1-x^2)^{\frac{n-3}{2}}}{\cB\left(\frac{n-1}{2},\, \frac{1}{2}\right)}
            \label{lem:eq-2}
        \end{equation}
        for $x \in (-1,\,1)$.
        Combining \Cref{lem:eq-1} and \Cref{lem:eq-2}, we have
        $$
            p_a(x) = \dfrac{(1-x^2)^{({n-3})/{2}}}{\cB\left( \frac{n-1}{2}, \frac{1}{2} \right)},\, x \in [-1,\, 1].
        $$
    \end{proof}
    
    
    \begin{lemma}
        Define $\omega$ as in \Cref{def:w}.
        Let $\f(x_0)$ be a boundary-image. 
        The projection $\f$ and the difference function $S$ satisfy the assumptions in \Cref{sec:lipschitz-smoothness-assumptions}.
        Let
        $$
            J := \nabla\f(x_0),\, \nablaS := \nablaS\left(\f(x_0)\right), \text{ and } v := \dfrac{J^\T \nabla S}{\| J^\T \nabla S \|_2}.
        $$
        When $0 < \delta \ll 1$, for any unit vector $u \in \bbR^n$,
        \begin{align*}
            \langle u,\, v \rangle > \dfrac{\omega}{\|J^\T \nablaS\|_2} & \,\Longrightarrow\, \sgn\left( S\left(\f(x_0+\delta u)\right) \right) = 1, \\
            \langle u,\, v \rangle < -\dfrac{\omega}{\|J^\T \nablaS\|_2} & \,\Longrightarrow\, \sgn\left( S\left(\f(x_0+\delta u)\right) \right) = -1. \\
        \end{align*}
        \label{lemma:2}
    \end{lemma}
    
    \begin{remark}
        The Lemma~\ref{lemma:2} reveals that $\langle u,\,v \rangle$ in some degree aligns with the sign of $S(\f(x_0 + \delta u))$.
        Later, we will write the cosine similarity as the sum of the product $\langle u,\,v \rangle \sgn\left(\f(x_0 + \delta u)\right)$.
        Such alignment, along with \Cref{lemma:1}, provides the bound for this sum of the product.
    \end{remark}
    
    \begin{proof}[Proof of \Cref{lemma:2}]
        We do Taylor expansion at point $x_0$ and $\f(x_0)$ for $\f$ and $S$ to the second order respectively using Lagrange remainder:
        \begin{align}
            \f(x_0 + \delta u) & = \f(x_0) + J\cdot \delta u + \dfrac{1}{2} \sum_{i=1}^n (\theta \delta u)^\T \mathbf{T}(x_0)_i (\theta \delta u) = \f(x_0) + \delta Ju + \dfrac{1}{2} \beta_\f \delta^2 \epsilon, \label{eq:tmp1} \\
            S\left( \f(x_0 + \delta u) \right) & = S\left(\f(x_0)\right) + \nabla S^\T \left( \delta J u + \dfrac{1}{2} \beta_\f \delta^2 \epsilon \right) + \dfrac{1}{2} \beta_S \left( \delta L_\f + \dfrac{1}{2} \beta_\f \delta^2 \right)^2 \theta_1 \label{eq:tmp2} \\
            & = \delta \nabla S^\T J u + \delta^2 \left( \dfrac{1}{2} \beta_\f L_S + \dfrac{1}{2} \beta_S L_\f^2 + \dfrac{1}{2} \delta \beta_\f \beta_S L_\f + \dfrac{1}{8}\delta^2 \beta_S \beta_\f^2\right) \theta_2. \label{eq:tmp3}
        \end{align}
        In above expressions, $\theta \in [0,\,1]$, $\theta_1, \theta_2 \in [-1,\, 1]$, $\epsilon \in \bbR^m$ is an error vector such that $\|\epsilon\|_2 \le 1$.
        
        In \Cref{eq:tmp1}, we use the smoothness condition of $\f$, which leads to $\| \sum_{i=1}^n v^\T \mathbf{T}(x_0)_i v \|_2 \le \beta_\f \|v\|_2^2$, where $\mathbf{T}$ is the second-order gradient tensor, i.e., $\mathbf{T}(x)_{ijk} = {\partial \f(x)_i}/\left({\partial x_j \partial x_k}\right)$. 
        In \Cref{eq:tmp2}, similarly, the smoothness condition of $S$ leads to $v^\T \mathbf{H} v \le \beta_S \|v\|_2^2$ where $\mathbf{H}$ is the Hessian matrix of $S$ and its spectral radius is bounded by $\beta_S$.
        We let $v = \delta J u + \frac{1}{2}\beta_\f \delta^2 \epsilon$ and observe that $\|v\|_2 \le \| \delta J u \|_2 + \frac{1}{2}\beta_\f \delta^2 \le \delta L_\f + \frac{1}{2}\beta_\f \delta^2$.
        From Taylor expansion we get \Cref{eq:tmp2}.
        \Cref{eq:tmp3} follows from $S(\f(x_0)) = 0$ by the boundary condition and $\nabla S^\T v \le L_S \|v \|_2$ by the Lipschitz condition.
        
        Consider the expression in the parenthesis of \Cref{eq:tmp3},
        we have 
        $$
             0
             \le
             \dfrac{1}{2} \beta_\f L_S + \dfrac{1}{2} \beta_S L_\f^2 + \dfrac{1}{2} \delta \beta_\f \beta_S L_\f + \dfrac{1}{8}\delta^2 \beta_S \beta_\f^2
             =
             \omega / \delta,
        $$
        where $\omega$ is as defined in \Cref{def:w}.
        As a result, we rewrite \Cref{eq:tmp3} as
        $$
            S(\f(x_0 + \delta u)) = \delta \nablaS^\T Ju + \delta \omega \theta_2.
        $$
        Given that $\theta_2 \in [-1,\,1]$, $S\left(\f(x_0 + \delta u)\right)$ can be bounded:
        $$
            \delta \nabla S^\T J u - \delta \omega \le S\left( \f(x_0 + \delta u) \right) \le \delta \nabla S^\T J u + \delta \omega.
        $$
        Since $\nablaS^\T J u = (J^\T \nablaS)^\T u = \| J^\T \nablaS\|_2 \langle u,\, v\rangle$, we rewrite the bound as:
        $$
            \delta \left( \| J^\T \nablaS \|_2 \langle u,\, v\rangle - \omega \right) 
            \le S\left(\f(x_0 + \delta u)\right) \le 
            \delta \left( \| J^\T \nablaS \|_2 \langle u,\, v\rangle + \omega \right).
        $$
        Thus, when $\| J^\T \nablaS \|_2 \langle u,\, v\rangle - \omega > 0$, i.e., $\langle u,\, v \rangle > {\omega}/{\|J^\T \nablaS\|_2}$, $S\left( \f(x_0 + \delta u) \right) > 0$;
        when $\| J^\T \nablaS \|_2 \langle u,\, v\rangle + \omega < 0$, i.e., $\langle u,\, v \rangle < - {\omega}/{\|J^\T \nablaS\|_2}$, $S\left( \f(x_0 + \delta u) \right) < 0$, which concludes the proof.
    \end{proof}
    
    
    \begin{lemma}
         Let $\f(x_0)$ be a boundary-image, i.e., $S\left(\f(x_0)\right) = 0$.
         The projection $\f$ and the difference function $S$ satisfy the assumptions in \Cref{sec:lipschitz-smoothness-assumptions}.
        Over the randomness of the sampling of orthogonal basis subset $u_1, u_2, \dots, u_B$ for $\bbR^n$ space,
        The expectation of cosine similarity between $\tnablafS$~(defined as \Cref{eq:tildefs}) and $\nabla\f(x_0)^\T \nablaS\left(\f(x_0)\right)$~($\nablafS$ for short)  satisfies
        \begin{equation}
            \left( 2\left(1-\dfrac{\omega^2}{\|\nablafS\|_2^2}\right)^{({n-1})/{2}} - 1 \right) \cdot \dfrac{2 \sqrt B}{\cB\left( \frac{n-1}{2},\,\frac{1}{2} \right) \cdot (n-1)} 
            \le \bbE\, \cos\,\langle \tnablafS,\, \nablafS \rangle \le 
            \dfrac{2 \sqrt B}{\cB\left( \frac{n-1}{2},\, \frac{1}{2}\right) \cdot (n-1)}.
        \end{equation}
        Here, $\omega$ is as defined in \Cref{def:w}, and we assume $\omega \le \|\nablafS\|_2$.
        \label{lemma:3}
    \end{lemma}
    
    \begin{remark}
        This theorem directly relates the intermediate gradient estimation $\tnablafS$ to the mapped true gradient $\nablafS$ by providing general cosine similarity bounds between them.
        The assumption that $\omega \le \|\nablafS\|_2$ can be easily achieved since $\delta$ is typically small and $\lim_{\delta \to 0} \omega / \delta$ is a constant. 
    \end{remark}
    
    \begin{proof}[Proof of \Cref{lemma:3}]
        According to \Cref{eq:tildefs},
        $$
            \tnablafS = \dfrac{1}{B} \sum_{i=1}^B \sgn\left( S\left( \f(x_0 + \delta u_i) \right) \right) u_i.
        $$
        Define $J := \nabla \f(x_0)$.
        Since $u_1, u_2, \dots, u_B$ is a subset of the orthonormal basis,
        \begin{align*}
            \langle \tnablafS,\, \nablafS \rangle & = \dfrac{1}{B} \sum_{i=1}^B \sgn\left(S\left( \f(x_0 + \delta u_i) \right) \right) \langle J^\T \nablaS,\, u_i \rangle \\
            & = \dfrac{\|J^\T \nablaS\|_2}{B} \sum_{i=1}^B \sgn\left(S\left( \f(x_0 + \delta u_i) \right) \right) \Big\langle \dfrac{J^\T \nablaS}{\|J^\T \nablaS\|_2},\, u_i \Big\rangle.
        \end{align*}
        Let $v := J^\T \nablaS / \|J^\T \nablaS\|_2$.
        Note that $\big\|\tnablafS\big\|_2 = \sqrt{\sum_{i=1}^B (1/B)^2} = 1/\sqrt B$, 
        we have
        \begin{equation}
            \cos\,\langle \tnablafS,\, \nablafS \rangle = \dfrac{\langle \tnablafS,\, \nablafS \rangle}{\|\tnablafS\|_2 \|\nablafS\|_2} = \dfrac{1}{\sqrt B} \sum_{i=1}^B \sgn\left(S\left( \f(x_0 + \delta u_i) \right) \right) \langle v,\, u_i \rangle.
            \label{eq-5}
        \end{equation}
        According to Lemma~\ref{lemma:1}, $\langle v,\, u_i \rangle$ follows the distribution $p_a$.
        Intuitively, we know that $\langle v,\, u_i \rangle$ in some degree decides $\sgn\left( S \left( \f(x_0 + \delta u_i) \right) \right)$.
        
        Consider each component $\left(\sgn\left(S\left( \f(x_0 + \delta u_i) \right) \right) \langle v,\, u_i \rangle\right)$.
        By Lemma~\ref{lemma:2}, in the worst case, only when $\|\langle v,\, u_i\rangle\| > \omega / \|J^\T \nablaS\|_2$, the $\sgn\left( S\left( \f(x_0+\delta u_i) \right) \right)$ is aligned with the sign of $\langle v,\,u_i \rangle$, otherwise their signs are always different.
        Since $\omega / \| J^\T \nablaS \|_2 \le 1$,
        \begin{align*}
          & \bbE_{u_i} \, \sgn\left(S\left( \f(x_0 + \delta u_i) \right) \right) \langle v,\, u_i \rangle \\
          \ge & 
          \int_{-1}^{-\omega/\|J^\T \nablaS\|_2} -xp_a(x) \diff x +
          \int_{-\omega/\|J^\T \nablaS\|_2}^{0} x p_a(x) \diff x +
          \int_{0}^{\omega/\|J^\T \nablaS\|_2} -x p_a(x) \diff x +
          \int_{\omega/\|J^\T \nablaS\|_2}^1 x p_a(x) \diff x \\
          = & \int_{0}^{\omega/\|J^\T \nablaS\|_2} -2x p_a(x) \diff x +
          \int_{\omega/\|J^\T \nablaS\|_2}^1 2x p_a(x) \diff x \\
          = & \dfrac{2}{\cB\left( \frac{n-1}{2},\, \frac{1}{2}\right) \cdot (n-1)} \left( 2\left( 1 - \dfrac{\omega^2}{\|\nabla\f^\T \nablaS\|_2^2} \right)^{({n-1})/{2}} - 1 \right).
        \end{align*}
        Here we use the fact that $p_a$ is symmetric.
        Inject it into \Cref{eq-5}:
        \begin{equation}
            \bbE \, \cos\,\langle \tnablafS,\, \nablafS \rangle \ge \dfrac{2\sqrt B}{\cB\left( \frac{n-1}{2},\, \frac{1}{2}\right) \cdot (n-1)} \left( 2\left( 1 - \dfrac{\omega^2}{\|\nabla\f^\T \nablaS\|_2^2} \right)^{({n-1})/{2}} - 1 \right).
            \label{eq-6}
        \end{equation}
        
        On the other hand, the upper bound can be obtained by forcing $\langle v,\, u_i \rangle$ and $S\left(\f(x_0 + \delta u_i)\right)$ be of the same sign everywhere, which means that
        $$
            \bbE_{u_i} \, \sgn\left(S\left( \f(x_0 + \delta u_i) \right) \right) \langle v,\, u_i \rangle
            \le 
            \int_{-1}^{0} -xp_a(x) \diff x +
            \int_{0}^{1} x p_a(x) \diff x 
            = \int_{0}^1 2x p_a(x) = 
            \dfrac{2}{\cB\left( \frac{n-1}{2},\, \frac{1}{2}\right) \cdot (n-1)}.
        $$
        Inject it into \Cref{eq-5}:
        \begin{equation}
            \bbE \, \cos\,\langle \tnablafS,\, \nablafS \rangle \le \dfrac{2\sqrt B}{\cB\left( \frac{n-1}{2},\, \frac{1}{2}\right) \cdot (n-1)}.
            \label{eq-7}
        \end{equation}
    \end{proof}
    
    
    \begin{lemma}
        For any positive integer $n \ge 2$, define
        $$
            c_n := \dfrac{2\sqrt n}{\cB\left( \frac{n-1}{2},\, \frac{1}{2} \right) \cdot (n-1)},
        $$
        where $\cB$ is the Beta function.
        We have $c_n \in \left( 2/\pi,\, 1\right)$ and $c_{n+2} < c_n$.
        \label{lemma:4}
    \end{lemma}
    
    \begin{remark}
        Using Lemma~\ref{lemma:4}, we can simplify the term
        $
            2\sqrt B / \left( \cB(\frac{n-1}{2},\,\frac 1 2) \cdot (n-1) \right)
        $ 
        in Lemma~\ref{lemma:3} to
        $ c_n \sqrt{B/n}$.
    \end{remark}
    
    \begin{proof}[Proof of \Cref{lemma:4}]
        Let $d_n := {\Gamma\left( \frac{n}{2} \right)}/{\Gamma\left( \frac{n-1}{2} \right)}$, where $\Gamma(\cdot)$ is the Gamma function.
        Notice that 
        $$
            c_n = \dfrac{2 \sqrt n}{\cB\left( \frac{n-1}{2},\, \frac{1}{2} \right) \cdot (n-1)} = \dfrac{2 \sqrt n \Gamma(\frac{n}{2})}{\Gamma(\frac{n-1}{2})\sqrt{\pi}\cdot (n-1)} = d_n \dfrac{2\sqrt n}{(n-1)\sqrt \pi}.
        $$
        
        \paragraph{(I.)} 
        For $n \ge 5$, $d_n = \dfrac{\Gamma\left( \frac{n}{2} \right)}{\Gamma\left( \frac{n-1}{2} \right)} = \dfrac{n-2}{n-3} \cdot \dfrac{\Gamma\left( \frac{n-2}{2} \right)}{\Gamma\left( \frac{n-3}{2} \right)} = \dfrac{n-2}{n-3} d_{n-2}$.
        Notice that
        $$
            \dfrac{d_n}{\sqrt{n-2}} = \dfrac{\sqrt{n-2}}{n-3} d_{n-2} = \dfrac{\sqrt{(n-2) \cdot (n-4)}}{n-3}\cdot \dfrac{d_{n-2}}{\sqrt{n-4}} \le \dfrac{d_{n-2}}{\sqrt{n-4}},
        $$
        and 
        $$
            \dfrac{d_3}{\sqrt 1} = \dfrac{\sqrt \pi}{2},\, \dfrac{d_4}{\sqrt 2} = \dfrac{2}{\sqrt \pi},
        $$
        we have $\dfrac{d_n}{\sqrt{n-2}} \le \dfrac{\sqrt{\pi}}{2}$ for $n \ge 3$.
        Therefore, 
        $$
            c_n = d_n \dfrac{2 \sqrt n}{(n-1)\sqrt \pi} \le \dfrac{\sqrt \pi}{2} \cdot \dfrac{2 \sqrt{n(n-2)}}{(n-1)\sqrt \pi} < 1
        $$
        for $n \ge 3$.
        When $n = 2$, $c_n = \dfrac{2\sqrt 2}{\pi} < 1$.
        So $c_n < 1$ holds for any $n \ge 2$.
        
        \paragraph{(II.)}
        Similarly, notice that
        $$
            \dfrac{d_n}{\sqrt{n-1}} = \dfrac{n-2}{(n-3)\sqrt{n-1}} d_{n-2} = \dfrac{n-2}{\sqrt{(n-3)(n-1)}} \cdot \dfrac{d_{n-2}}{\sqrt{n-3}} \ge \dfrac{d_{n-2}}{\sqrt{n-3}},
        $$
        and
        $$
            \dfrac{d_3}{\sqrt{2}} = \dfrac{1}{4}\sqrt{2\pi},\, \dfrac{d_2}{\sqrt{1}} = \dfrac{1}{\sqrt \pi},
        $$
        we have $\dfrac{d_n}{\sqrt{n-1}} \ge \dfrac{1}{\sqrt \pi}$ for $n \ge 2$.
        Therefore,
        $$
            c_n = d_n \dfrac{2 \sqrt n}{(n-1)\sqrt \pi} \ge \sqrt{\dfrac{n-1}{\pi}} \cdot \dfrac{2 \sqrt n}{(n-1)\sqrt \pi} = \dfrac{2}{\pi}\sqrt{\dfrac{n}{n-1}} > \dfrac{2}{\pi}.
        $$
        
        \paragraph{(III.)}
        Since $d_{n+2} = d_n \cdot n/(n-1)$
        and
        $
            c_n = d_n \cdot \left(2 \sqrt n\right) / \left((n-1) \sqrt \pi\right),
        $
        we have
        $$
            \dfrac{c_{n+2}}{c_n} = \dfrac{d_{n+2}}{d_n} \cdot \dfrac{\sqrt{n+2}}{n+1} \cdot \dfrac{n-1}{\sqrt n} = \dfrac{n}{n-1} \cdot \dfrac{\sqrt{n+2}}{n+1} \cdot \dfrac{n-1}{\sqrt n} = \dfrac{\sqrt{n(n+2)}}{n+1} < 1.
        $$
        
        In summary, for any positive integer $n \ge 2$, we have shown ${2}/{\pi} < c_n < 1$ and $c_{n+2} < c_n$.
    \end{proof}
    
    Now we are ready to prove the main theorem which provides the general cosine similarity bounds for our gradient estimator.

    \begin{customthm}{1}[restated]
        Let $\f(x_0)$ be a boundary-image, i.e., $S\left(\f(x_0)\right) = 0$.
        The projection $\f$ and the difference function $S$ satisfy the assumptions in \Cref{sec:lipschitz-smoothness-assumptions}.
        Over the randomness of the sampling of orthogonal basis subset $u_1, u_2, \dots, u_B$ for $\bbR^n$ space,
        the expectation of cosine similarity between $\tnablaS\left(\f(x_0)\right)$~($\tnablaS$ for short) and $\nablaS\left(\f(x_0)\right)$~($\nablaS$ for short) satisfies
        \begin{equation}
            \left( 2\left(1-\dfrac{\omega^2}{\|\nablafS\|_2^2} \right)^{({n-1})/{2}} - 1 \right)
            \dfrac{\|\nablafS\|_2}{L_\f \|\nablaS\|_2} 
            \sqrt{\frac B n}
            c_n
            \le \, \bbE\, \cos\,\langle \tnablaS,\, \nablaS \rangle 
            \le \, 
            \dfrac{\|\nablafS\|_2}{l_\f \|\nablaS\|_2}
            \sqrt{\frac B n}
            c_n
            ,
        \end{equation}
        where $\omega$ is as defined in \Cref{def:w}, and we assume $\omega \le \|\nablafS\|_2$;
        $c_n \in (2/\pi,\,1)$ is a constant depended on $n$;
        $L_\f$ is as defined in assumptions in \Cref{sec:lipschitz-smoothness-assumptions};
        and $l_{\f} := \lambda_{\min}(\nabla\f(x_0))$.
    \end{customthm}
    
    \begin{proof}[Proof of \Cref{thm:1}]
        According to \Cref{eq:tildeS}, we know $\tnablaS = \nabla\f \tnablafS$, where $\nabla\f$ is the shorthand of $\nabla\f(x_0)$.
        Thus,
        $$
            \langle \tnablaS,\, \nablaS \rangle = \tnablaS^\T \nablaS = \tnablafS^\T \nabla \f^\T \nablaS = \langle \tnablafS,\, \nablafS \rangle = \cos \,\langle \tnablafS,\, \nablafS \rangle \cdot \big\|\tnablafS\big\|_2 \|\nablafS\|_2.
        $$
        Therefore, 
        \begin{equation}
            \cos\,\langle \tnablaS,\, \nablaS \rangle = \cos \,\langle \tnablafS,\, \nablafS \rangle \dfrac{\big\|\tnablafS\big\|_2 \|\nablafS\|_2}{\big\|\tnablaS\big\|_2 \|\nablaS\|_2}.
            \label{eq:thm1-1}
        \end{equation}
        
        According to the estimation formula of $\tnablafS$~(\Cref{eq:tildefs}), $\big\|\tnablafS\big\|_2 = \sqrt B$.
        Furthermore, 
        $\big\|\tnablaS\big\| \le \lambda_{\max}(\nabla\f) \cdot \big\|\tnablafS\big\|_2 \le L_\f \sqrt B$,
        $\big\|\tnablaS\big\| \ge \lambda_{\min}(\nabla\f) \cdot \big\|\tnablafS\big\|_2 = l_\f \sqrt B$, which means that
        $$
            \dfrac{1}{L_\f} \le \dfrac{\big\|\tnablafS\big\|_2}{\big\|\tnablaS\big\|_2} \le \dfrac{1}{l_\f}.
        $$
        According to \Cref{eq:thm1-1}, we have
        \begin{equation}
            \cos \,\langle \tnablafS,\, \nablafS \rangle \dfrac{\|\nablafS\|_2}{L_\f \|\nablaS\|_2}
            \le
            \cos\,\langle \tnablaS,\, \nablaS \rangle
            \le 
            \cos \,\langle \tnablafS,\, \nablafS \rangle \dfrac{\|\nablafS\|_2}{l_\f \|\nablaS\|_2}.
            \label{eq:thm1-2}
        \end{equation}
        Inject the bound for $\bbE\,\cos \,\langle \tnablafS,\, \nablafS \rangle$ in Lemma~\ref{lemma:3} and the simplification from Lemma~\ref{lemma:4} to \Cref{eq:thm1-2} yields the desired bound.
    \end{proof}
    
    We discuss the implications of the bound in \Cref{sec:theory-implications} and \Cref{sec:adx-implications}.
    
    \begin{customcor}{1}[restated]
        Let $\f(x_0)$ be a boundary-image, i.e., $S\left(\f(x_0)\right) = 0$.
        The projection $\f$ is locally linear around $x_0$ with radius $\delta$. 
        $L_\f := \lambda_{\max} (\nabla \f(x_0)), l_\f := \lambda_{\min} (\nabla \f(x_0))$.
        The difference function $S$ satisfies the assumptions in \Cref{sec:lipschitz-smoothness-assumptions}.
        Over the randomness of the sampling of orthogonal basis subset $u_1, u_2, \dots, u_B$ for $\bbR^n$ space,
        the expectation of cosine similarity between $\tnablaS\left(\f(x_0)\right)$~($\tnablaS$ for short) and $\nablaS\left(\f(x_0)\right)$~($\nablaS$ for short) satisfies \Cref{eq:main-bound}
        with
        \begin{equation}
            \omega := \dfrac{1}{2}\delta \beta_S L_\f^2.
        \end{equation}
        We assume $\omega \le \|\nablafS\|_2$.
        The $c_n \in (2/\pi,\,1)$ is a constant depended on $n$.
    \end{customcor}
    
    \begin{remark}
        This is a direct application of \Cref{thm:1}. Since $\f$ is locally linear, we have $\beta_\f = 0$, and the corollary follows.
        We discuss its implication in \Cref{sec:linear-case}.
    \end{remark}
    
    \begin{customcor}{2}[restated]
        Given the projection $\f$ and the difference function $S$, to achieve expected cosine similarity $\bbE \langle \nablaS(\f(x_0)),\,\tnablaS(\f(x_0)) \rangle = s$, the required query number $B$ is in $\Theta(s^2)$.
    \end{customcor}
    
    \begin{proof}[Proof of \Cref{cor:query-complexity}]
        From \Cref{thm:1}, we can observe that 
        $$
            \Theta(\sqrt B) \le \bbE\,\cos\langle \tnablaS,\,\nablaS \rangle \le \Theta(\sqrt B).
        $$
        Therefore, when $\bbE\,\cos\langle \tnablaS,\,\nablaS \rangle = s$, the number of queries $B$ is in $\Theta(s^2)$.
    \end{proof}
    
    \begin{remark}
        The above corollary shows the relation between the expected cosine similarity and  the query number when the projection $\f$ is fixed.
        Note that the cosine similarity is bounded, i.e., the cosine similarity between two totally aligned vectors is $1$.
        The $\Theta(s^2)$ order implies that to achieve moderate cosine similarity, a small number of queries is needed, while high cosine similarity needs much more queries.
        Therefore, to achieve high cosine similarity, it is better to fix the number of queries and reduce the dimension of subspace, $n$, which is related with cosine similarity with order $\Theta(1 / \sqrt n)$.
        The reduction on subspace dimension is the shared technique between QEBA and \sysname.
    \end{remark}

\section{Proof of Existence of Better Nonlinear Projection}

    \label{sec:adx-existence-better-nonlinear-projection}

    \begin{customthm}{2}[restated]
        Let $\f(x_0)$ be a boundary-image, i.e., $S\left(\f(x_0)\right) = 0$.
        The projection $\f$ is locally linear around $x_0$ with radius $\delta$. 
        $L_\f := \lambda_{\max} (\nabla \f(x_0))$, $l_\f := \lambda_{\min} (\nabla \f(x_0))$.
        The difference function $S$ satisfies the assumptions in \Cref{sec:lipschitz-smoothness-assumptions}.
        
        There exists a nonlinear projection $\f'$ satisfying the assumptions in \Cref{sec:lipschitz-smoothness-assumptions}, with $\f'(x_0) = \f(x_0)$ and $\nabla \f'(x_0) = \nabla \f(x_0)$, such that
        over the randomness of the sampling of orthogonal basis subset $u_1, u_2, \dots, u_B$ for $\bbR^n$ space,
        the expectation of cosine similarity between $\tnablaS\left(\f'(x_0)\right)$~($\tnablaS$ for short) and $\nablaS\left(\f'(x_0)\right)$~($\nablaS$ for short) satisfies
        \Cref{eq:main-bound} with
        \begin{equation}
            \omega := \dfrac{1}{2}\delta \beta_S L_\f^2 - \dfrac{1}{5} \beta_\f \beta_S \delta^2 L_\f < \dfrac{1}{2} \delta \beta_S L_\f^2.
        \end{equation}
        We assume $\omega \le \|\nablafS\|_2$.
        The $c_n \in (2/\pi,\,1)$ is a constant depended on $n$.
    \end{customthm}
    
    \begin{proof}[Proof of \Cref{thm:2}]
        For convenience, in the proof, we define $J := \nabla \f(x_0)$.
            According to the proof of \Cref{thm:1}~(especially the usage of \Cref{lemma:2}), we only need to show that for arbitrary $S$, there exists a projection $\f'$ such that $\nabla \f'(x_0) = \nabla\f(x_0) = J$, $\f'(x_0) = \f(x_0)$ and $\f'$ satisfies the smoothness and Lipschitz assumptions, so that for arbitrary vector $u$ with $\|u\|_2 = 1$,
            \begin{equation}
                \begin{aligned}
                    \Big\langle u,\, \dfrac{J^\T \nabla S}{\|J^\T \nabla S\|_2} \Big\rangle > \dfrac{\omega}{\|J^\T \nabla S\|_2} \, & \Longrightarrow\, \sgn(S(\f(x_0+\delta u))) = 1, \\
                    \Big\langle u,\, \dfrac{J^\T \nabla S}{\|J^\T \nabla S\|_2} \Big\rangle < \dfrac{\omega}{\|J^\T \nabla S\|_2} \, & \Longrightarrow\, \sgn(S(\f(x_0+\delta u))) = -1.
                \end{aligned}
                \label{eq:non-linear-omega-cond}
            \end{equation}
            We prove this by construction:
            we define $\f': \bbR^n \to \bbR^m$ such that
            for arbitrary $u \in \bbR^n$, 
            \begin{equation}
                \f'(x_0 + u) = \f(x_0) + J\cdot u - \dfrac{1}{2} \alpha \|u\|_2 Ju,
                \label{eq:nonlinear-f-def-2}
            \end{equation}
            where $\alpha \in [0,\,0.8\beta_\f/L_\f]$ is an adjustable parameter~(it is later fixed to $0.8\beta_\f / L_\f$, but for the generality of the proof, we deem it as an adjustable parameter for now).
                
            \begin{fact}
                The $\f'$ defined as in \Cref{eq:nonlinear-f-def-2}: (1)~has gradient $J$ at point $x_0$, (2)~is $L_\f$-Lipschitz, and (3)~is $\beta_\f$-smooth around $x_0$ with radius $\delta$.
                \label{fact:non-linear-f-def-2-valid}
            \end{fact}
                
                \begin{proof}[Proof of \Cref{fact:non-linear-f-def-2-valid}]
                    \hfill
                    \paragraph{Gradient at $x_0$.}
                        Since
                        $$
                            \lim_{u \to 0} \dfrac{\Big\|\dfrac{1}{2}\alpha\|u\|_2 Ju\Big\|_2}{\|u\|_2} = \dfrac{1}{2}\alpha \lim_{u\to 0} \|Ju\|_2 \le \dfrac{1}{2}\alpha L_\f \|u\|_2 = 0,
                        $$
                        we have $\f'(x_0 + u) = \f'(x_0) + J \cdot u + o(u)$ so $\nabla\f' := \nabla\f'(x_0) = J$.
                        
                    \paragraph{Lipschitz.}
                    
                        Firstly, let us derive the gradient of $\f'$ at an arbitrary point.
                        Because
                        $$
                        \begin{aligned}
                            \dfrac{\partial \f'(x_0+u)_i}{\partial u_j} 
                            & = J_{ij} - \dfrac{1}{2}\alpha \dfrac{\partial \left(\|u\|_2 Ju\right)_i}{\partial u_j} = J_{ij} - \dfrac{1}{2}\alpha \left( \dfrac{u_j}{\|u\|_2} \sum_{k=1}^n J_{ik}u_k + \|u\|_2 J_{ij} \right)\\
                            & = \left( 1-\dfrac{1}{2}\alpha\|u\|_2 \right) J_{ij} - \dfrac{\alpha}{2\|u\|_2} (Juu^\T)_{ij},
                        \end{aligned}
                        $$
                        we have
                        \begin{equation}
                            \nabla \f'(x_0 + u) = 
                            \left(1 - \dfrac{1}{2}\alpha\|u\|_2 \right) J
                            - 
                            \dfrac{\alpha}{2\|u\|_2} Juu^\T
                            .
                            \label{eq:non-linear-case-2-gradient}
                        \end{equation}
                        We bound its maximum eigenvalue:
                        $$
                            \lambda_{\max} \left(\nabla \f'(x_0 + u) \right) \le \left(1 - \dfrac{1}{2}\alpha\|u\|_2 \right) \lambda_{\max} (J) + \dfrac{\alpha}{2\|u\|_2} \lambda_{\max} (J) \|u\|_2^2 = \lambda_{\max}(J) = L_\f.
                        $$
                        Therefore, $\f'$ is $L_\f$-Lipschitz.
                    
                    \paragraph{Smoothness.}
                        The smoothness part is more involved.
                        
                        To show $\f'$ is $\beta_\f$-smooth, 
                        we need to consider arbitrary $u_1, u_2 \in \bbR^n$, and prove that 
                        $$
                            \dfrac{\lambda_{\max} \left(\nabla\f'(x_0 + u_1) - \nabla\f'(x_0 + u_2)\right)}{\|u_1 - u_2\|_2} \le \beta_\f
                        $$ 
                        always holds.
                        From \Cref{eq:non-linear-case-2-gradient},
                        $$
                            \nabla \f' (x_0 + u_1) - \nabla \f' (x_0 + u_2) = 
                            \dfrac{\alpha}{2} (\|u_2\|_2 - \|u_1\|_2) J 
                            -
                            \dfrac{\alpha}{2}J
                            \left(\dfrac{u_1 u_1^\T}{\|u_1\|_2}  - \dfrac{u_2 u_2^\T}{\|u_2\|_2}  \right).
                        $$
                        Thus,
                        $$
                            \dfrac{\lambda_{\max} \left(\nabla\f'(x_0 + u_1) - \nabla\f'(x_0 + u_2)\right)}{\|u_1 - u_2\|_2}
                            \le
                            \dfrac{\lambda_{\max} \left(\dfrac{\alpha}{2} (\|u_2\|_2 - \|u_1\|_2) J \right)}{\|u_1 - u_2\|_2}
                            +
                            \dfrac{\alpha L_\f}{2}
                            \cdot
                            \underbrace{
                            \dfrac{\lambda_{\max}\left(\dfrac{u_1 u_1^\T}{\|u_1\|_2}  - \dfrac{u_2 u_2^\T}{\|u_2\|_2}  \right)}{\|u_1 - u_2\|_2}
                            }_{\mathrm{(*)}}
                            .
                        $$
                        Consider the first term: from $\big|\|u_2\|_2 - \|u_1\|_2\big| \le \|u_1 - u_2\|_2$,
                        $$
                            \dfrac{\lambda_{\max} \left(\dfrac{\alpha}{2} (\|u_2\|_2 - \|u_1\|_2) J \right)}{\|u_1 - u_2\|_2} \le \dfrac{1}{2}\alpha L_\f.
                        $$
                        \begin{fact}
                            For arbitrary $u, v \in \bbR^n$,
                            $$
                                \lambda_{\max}\left(\dfrac{u u^\T}{\|u\|_2}  - \dfrac{v v^\T}{\|v\|_2}  \right) \le 1.5\|u - v\|_2.
                            $$
                            \label{fact:non-linear-case-2-eigen-bound}
                        \end{fact}
                        From \Cref{fact:non-linear-case-2-eigen-bound}, the second term $\mathrm{(*)}$ is bounded by $1.5$.
                        By summing them up,
                        we have
                        $$
                            \dfrac{\lambda_{\max} \left(\nabla\f'(x_0 + u_1) - \nabla\f'(x_0 + u_2)\right)}{\|u_1 - u_2\|_2} \le 1.25 \alpha L_\f \le \beta_\f / L_\f \cdot L_\f = \beta_\f,
                        $$
                        i.e., $\f'$ is $\beta$-smooth.
                        
                        \begin{proof}[Proof of \Cref{fact:non-linear-case-2-eigen-bound}]
                            \begin{equation}
                                \begin{aligned}
                                    \lambda_{\max} \left(\dfrac{u u^\T}{\|u\|_2}  - \dfrac{v v^\T}{\|v\|_2}  \right) & =  
                                    \max_{\|w\|_2 = 1} 
                                    w^\T \left(\dfrac{u u^\T}{\|u\|_2}  - \dfrac{v v^\T}{\|v\|_2}  \right) w =
                                    \max_{\|w\|_2 = 1}
                                    \dfrac{\|u^\T w\|_2^2}{\|u\|_2} - \dfrac{\|v^\T w\|_2^2}{\|v\|_2} \\
                                    & =
                                    \max_{\|w\|_2 = 1}
                                    \|u\| \cos^2 \langle u,\,w \rangle - \|v\| \cos^2 \langle v,\,w \rangle.
                                \end{aligned}
                                \label{eq:fact-3-3-1}
                            \end{equation}
                            
                            From geometry, we know that the $\cos\langle u,\,w \rangle$ of a unit vector $w$ lying outside the place $P_{uv}$ equals to $\|w_{uv}\|_2 \cos\langle w_{uv},\,u\rangle$, where $w_{uv}$ is its projection onto plane $P_{uv}$, having length $\|w_{uv}\|_2 \le 1$.
                            Therefore, we only need to consider all vectors with length smaller or equal to $1$ lying on the plane $P_{uv}$~(i.e., the projection of any unit vector $w$ onto the plane $P_{uv}$), i.e.,
                            $$
                                \text{\Cref{eq:fact-3-3-1}} = \max_{\substack{\|w\|_2 \le 1\\ w \in P_{uv}}} \|w\|^2 \left( \|u\| \cos^2 \langle u,\,w \rangle - \|v\| \cos^2 \langle v,\,w \rangle \right) = 
                                \max_{\substack{\|w\|_2 = 1\\ w \in P_{uv}}} \left( \|u\| \cos^2 \langle u,\,w \rangle - \|v\| \cos^2 \langle v,\,w \rangle \right).
                            $$
                            Let $\theta$ be the angle between $u$ and $v$, $\beta$ be the angle between $u$ and $w$, then the angle between $v$ and $w$ is $\beta - \theta$.
                            Written as the optimization over $\beta$, we have
                            $$
                            \begin{aligned}
                                \text{\Cref{eq:fact-3-3-1}} & = \max_{\beta} \|u\|\cos^2 \beta - \|v\| \cos^2 (\beta - \theta) \\
                                & = \max_{\beta} \dfrac{1}{2} \left( \|u\| - \|v\| \right) + \dfrac{1}{2} \left( \|u\|\cos 2\beta - \|v\|\cos 2(\beta - \theta) \right) \\
                                & = \dfrac{1}{2} \left( \|u\| - \|v\| \right) + \dfrac{1}{2} \left(\max_{\beta} \|u\| \cos \beta - \|v\| \cos (\beta - 2\theta)\right).  
                            \end{aligned}
                            $$
                            From geometry, we know for any $\beta$, $\|u\|\cos \beta - \|v\| \cos(\beta-2\theta) \le 2\|u - v\|$.
                            Furthermore, $\|u\| - \|v\| \le \|u - v\|$.
                            Thus, $\text{\Cref{eq:fact-3-3-1}} \le 1.5\|u - v\|$.
                        \end{proof}
                        
                        Given \Cref{fact:non-linear-case-2-eigen-bound}, as shown before, $\f'$ is $\beta$-smooth.
                    
                    To this point, we have proven the three arguments in \Cref{fact:non-linear-f-def-2-valid} respectively.
                \end{proof}
                
                Now we inject $\f$ into the Taylor expansion expression for $S(\f'(x_0 + \delta u))$, where $u$ is a unit vector, i.e., $\|u\|_2 = 1$.
                Similar as \Cref{eq:tmp1,eq:tmp2,eq:tmp3}:
                \begin{equation}
                    \begin{aligned}
                        & S(\f'(x_0 + \delta u)) \\
                        = & S\left( \f(x_0) + \delta Ju - \dfrac{1}{2}\alpha \delta^2 Ju \right) \\
                        = & S(\f(x_0)) + \delta \nablaS^\T J u - \dfrac{1}{2}\alpha \delta^2  \nablaS^\T Ju + \dfrac{1}{2}\theta^2 \left( \delta J u -\dfrac{1}{2} \alpha \delta^2  Ju \right)^\T \mathbf{H} \left( \delta J u -\dfrac{1}{2}\alpha \delta^2  Ju \right),
                    \end{aligned}
                    \label{eq:non-linear-caseB-1}
                \end{equation}
                where $\theta \in [-1,\,1]$ is depended on $S$, and $\mathbf{H}$ is the Hessian matrix of $S$ at point $x_0$.
                Because $\f(x_0)$ is the boundary-image, we have $S(\f(x_0)) = 0$.
                We can also bound the last term from the smoothness assumption on $S$:
                $$
                    \Big| \dfrac{1}{2}\theta^2 \left( \delta J u -\dfrac{1}{2} \alpha \delta^2  Ju \right)^\T \mathbf{H} \left( \delta J u -\dfrac{1}{2}\alpha \delta^2  Ju \right) \Big| 
                    \le \dfrac{1}{2} \beta_S \delta^2 \Big\|Ju - \dfrac{1}{2}\alpha \delta  Ju \Big\|_2^2 
                    \le \dfrac{1}{2} \beta_S \delta^2 \left( 1 - \dfrac{1}{2}\alpha \delta  \right)^2 L_\f^2.
                $$
                
                Define $v := {J^\T \nablaS(\f(x_0))}/{\|J^\T \nablaS(\f(x_0))\|_2}$.
                From \Cref{eq:non-linear-caseB-1}, we get
                $$
                \begin{aligned}
                    S(\f'(x_0 + \delta u)) \ge \delta \left( 1 - \dfrac{1}{2}\alpha \delta \right) \langle u,\,v \rangle \|v\|_2 - \dfrac{1}{2} \beta_S \delta^2 \left( 1 - \dfrac{1}{2}\alpha \delta \right)^2 L_\f^2, \\
                    S(\f'(x_0 + \delta u)) \le \delta \left( 1 - \dfrac{1}{2}\alpha \delta \right) \langle u,\,v \rangle \|v\|_2 + \dfrac{1}{2} \beta_S \delta^2 \left( 1 - \dfrac{1}{2}\alpha \delta \right)^2 L_\f^2.
                \end{aligned}
                $$
                Therefore,
                $$
                    |\langle u,\,v \rangle| \|v\|_2 \ge \dfrac{1}{2}\beta_S \delta \left( 1 - \dfrac{1}{2}\alpha\delta\right) L_\f^2
                    \,\Longrightarrow\, \sgn(S(\f(x_0 + \delta u))) = \sgn(\langle u,\,v \rangle).
                $$
                Note that $\alpha \in [0,\,0.8\beta_\f/L_\f]$, and larger $\alpha$ induces smaller RHS.
                We let $\alpha = 0.8\beta_\f / L_\f$,
                and get
                $$
                    |\langle u,\,v \rangle| \|v\|_2 \ge \dfrac{1}{2}\delta \beta_S L_\f^2 - \dfrac{1}{5} \beta_\f \beta_S \delta^2 L_\f
                    \,\Longrightarrow\, \sgn(S(\f(x_0 + \delta u))) = \sgn(\langle u,\,v \rangle).
                $$
                In other words,
                $$
                    \omega := \dfrac{1}{2}\delta \beta_S L_\f^2 - \dfrac{1}{5} \beta_\f \beta_S \delta^2 L_\f
                $$
                satisfies the condition \Cref{eq:non-linear-omega-cond}.
                Following the same proof as in \Cref{thm:1} using $\omega$, we get the desired cosine similarity bound for the projection $\f'$.
        \end{proof}

\section{Implications of Gradient Estimation Analysis}

    \label{sec:adx-implications}
    
    In this section, we provide further discussions on the gradient estimation analysis omitted in \Cref{sec:theory-implications} and the supporting theorems.

    \subsection{Comparison of Different Gradient Estimators}    
    \label{sec:adx-comparison-of-different-gradient-estimators}
    
        We instantiate the cosine similarity bounds for gradient estimators in HSJA~\citep{chen2020hopskipjumpattack} and QEBA~\citep{cvpr2020QEBA}.
        Then, we compare these bounds along with the bound for \sysname.
        The definitions of these estimators are presented in Appendix~\ref{sec:adx-concretization}.
        
        \paragraph{HSJA.}
        In HSJA, the projection is an identical function.
        Therefore, $\|\nablafS\| = \|\nablafS\|$, and $L_\f = 1$, $\beta_\f = 0$.
        We apply \Cref{thm:1} and yield the following cosine similarity bound.
        \begin{corollary}[Bound for HSJA Gradient Estimator]
            Let $x_0$ be a boundary-image, i.e., $S\left(x_0\right) = 0$.
            The difference function $S$ satisfies the assumptions in \Cref{sec:lipschitz-smoothness-assumptions}.
            Using HSJA gradient estimator as in \Cref{eq:hsja-estimator},
            over the randomness of the sampling of orthogonal basis subset $u_1, u_2, \dots, u_B$ for $\bbR^m$ space,
            the expectation of cosine similarity between $\tnablaS\left(x_0\right)$~($\tnablaS$ for short) and $\nablaS\left(x_0\right)$~($\nablaS$ for short) satisfies
            $$
                \left( 2\left(1-\dfrac{\omega^2}{\|\nablaS\|_2^2} \right)^{\frac{m-1}{2}} - 1 \right)
                \sqrt{\frac B m}
                c_m
                \le \, \bbE\, \cos\,\langle \tnablaS,\, \nablaS \rangle 
                \le \, 
                \sqrt{\frac B m}
                c_m
                ,
            $$
            where
            $\omega = \frac{1}{2}\delta \beta_S$,
            and the $c_m \in (2/\pi,\,1)$ is a constant depended on $m$.
            \label{cor:hsja}
        \end{corollary}
        
        \begin{remark}
            In the corollary, we can see that without subspace projection, all terms are directly related to the dimensionality of the input space, $m$.
        \end{remark}
        
        \paragraph{QEBA.}
        In QEBA, the projection is a random orthogonal transformation denoted by the matrix $\bW$.
        Similarly, we yield the following bound.
        \begin{corollary}[Bound for QEBA Gradient Estimator]
            Let $x_0$ be a boundary-image, i.e., $S\left(x_0\right) = 0$.
            The difference function $S$ satisfies the assumptions in \Cref{sec:lipschitz-smoothness-assumptions}.
            Using QEBA gradient estimator as in \Cref{eq:qeba-estimator},
            over the randomness of the sampling of orthogonal basis subset $u_1, u_2, \dots, u_B$ for $\bbR^n$ space,
            the expectation of cosine similarity between $\tnablaS\left(x_0\right)$~($\tnablaS$ for short) and $\nablaS\left(x_0\right)$~($\nablaS$ for short) satisfies
            $$
                \left( 2\left(1-\dfrac{\omega^2}{\|\bW^\T \nablaS\|_2^2} \right)^{\frac{n-1}{2}} - 1 \right)
                \dfrac{\|\bW^\T \nablaS\|_2}{\|\nablaS\|_2} 
                \sqrt{\frac B n}
                c_n
                \le \, \bbE\, \cos\,\langle \tnablaS,\, \nablaS \rangle 
                \le \, 
                \dfrac{\|\bW^\T \nablaS\|_2}{\|\nablaS\|_2}
                \sqrt{\frac B n}
                c_n
                ,
            $$
            where
            $\omega = \frac{1}{2}\delta \beta_S$,
            and the $c_n \in (2/\pi,\,1)$ is a constant depended on $m$.
            \label{cor:qeba}
        \end{corollary}
        
        \citet{cvpr2020QEBA} present a similar but slightly tighter  cosine similarity bound which replaces $\|\bW^\T \nabla S\|_2$ by $\|\nabla S\|_2$ leveraging the fact that the projection $\bW$ is random.
        
        \paragraph{Comparison between HSJA and QEBA.}
            In QEBA,
            when $\bW$ contains a base vector which aligns well with $\nablaS$, i.e., there exists $i \in [n]$ such that $|\cos \langle \bW_{:,i},\,\nablaS \rangle|$ is close to $1$, then $\|\bW^\T \nablaS\|_2 \approx \|\nablaS\|_2$.
            Heuristics are used in QEBA to increase the alignment between basis and the vector $\nablaS$.
            When the alignment is good,
            the bound in \Cref{cor:qeba} differs from that in \Cref{cor:hsja} only in that $m$ is replaced by $n$.
            Given that $n$ is the dimension of subspace which is usually much smaller than $m$, we know
            $$
                \left(1-\dfrac{\omega^2}{\|\bW^\T \nablaS\|_2^2} \right)^{\frac{n-1}{2}} \gg \left(1-\dfrac{\omega^2}{\|\nablaS\|_2^2} \right)^{\frac{m-1}{2}}
                \text{ and }
                \sqrt{\frac B n} \gg \sqrt{\frac B m}.
            $$
            As a result, when $B$ is the same, both the lower bound and upper bound in QEBA outperform those of HSJA significantly; and to achieve the same cosine similarity, QEBA requires much fewer queries than HSJA.
        
        \paragraph{\sysname.}
            Our proposed NonLinear-BA enables the use of nonlinear projection $\f$.
            As shown by \Cref{thm:1}, due to the nonlinearity, the cosine similarity lower bound of nonlinear projection is worse than the linear counterpart~(QEBA) due to the additional terms in $\omega$.
            However, \Cref{thm:2}, when compared with linear projection bound in \Cref{sec:better-nonlinear}, implies the existence of better nonlinear projection.
            The existence is proved by a specific construction of a `good' nonlinear projection which provides higher cosine similarity.
            Here, we present another `good' nonlinear projection, to show that such nonlinear projection is not rare or specific.
            
            \begin{theorem}[Existence of Better Nonlinear Projection, Part II]
                Let $\f(x_0)$ be a boundary-image, i.e., $S\left(\f(x_0)\right) = 0$.
                The projection $\f$ is locally linear around $x_0$ with radius $\delta$. 
                $L_\f := \lambda_{\max} (\nabla \f(x_0))$, $l_\f := \lambda_{\min} (\nabla \f(x_0))$.
                The difference function $S$ satisfies the assumptions in \Cref{sec:lipschitz-smoothness-assumptions}.
                
                There exists a nonlinear projection $\f'$ satisfying the assumptions in \Cref{sec:lipschitz-smoothness-assumptions}, with $\f'(x_0) = \f(x_0)$ and $\nabla \f'(x_0) = \nabla \f(x_0)$, such that
                over the randomness of the sampling of orthogonal basis subset $u_1, u_2, \dots, u_B$ for $\bbR^n$ space,
                the expectation of cosine similarity between $\tnablaS\left(\f'(x_0)\right)$~($\tnablaS$ for short) and $\nablaS\left(\f'(x_0)\right)$~($\nablaS$ for short) satisfies
                \Cref{eq:main-bound} with
                \begin{equation}
                    \omega < \dfrac{1}{2} \delta \beta_S L_\f^2.
                \end{equation}
                We assume $\omega \le \|\nablafS\|_2$, and $\delta < {L_S}/({\beta_S L_\f})$.
                The $c_n \in (2/\pi,\,1)$ is a constant depended on $n$.
                \label{thm:3}
            \end{theorem}
            
            \begin{proof}[Proof of \Cref{thm:3}]
                Let $J:= \nabla \f(x_0)$, and $v := {J^\T \nablaS(\f(x_0))}/{\|J^\T \nablaS(\f(x_0))\|_2}$.
                For arbitrary $u \in \bbR^n$, we define $\f'(x_0+u)$ as such:
                \begin{equation}
                    \f'(x_0 + u) = 
                        \f(x_0) + J\cdot u + \dfrac{1}{2} \sgn(\langle u,\,v \rangle) \langle u,\,v \rangle^2 k\nabla S,
                    \label{eq:nonlinear-f-def-1}
                \end{equation}
                where $k \in [0,\,\beta_\f / L_S]$ is an adjustable parameter.
            
                \begin{fact}
                    The $\f'$ defined as \Cref{eq:nonlinear-f-def-1} has gradient $J$ at point $x_0$ and is $\beta_\f$-smooth.
                    \label{fact:non-linear-f-def-1-valid}
                \end{fact}
                
                \begin{proof}[Proof of \Cref{fact:non-linear-f-def-1-valid}]
                    Since 
                    $$
                        \lim_{u\to 0} \dfrac{\Big\|\dfrac{1}{2} \langle u,\,v \rangle^2 k \nablaS\Big\|_2}{\|u\|_2} \le \lim_{u\to 0} \dfrac{1}{2} |\langle u,\, v \rangle| k \|\nablaS\|_2 \le \lim_{u\to 0} \dfrac{1}{2} \dfrac{\beta_\f}{L_S} L_S \|u\|_2 = 0,  
                    $$
                    we have $\f'(x_0 + u) = \f(x_0) + J\cdot u + o(u)$ so $\nabla \f'(x_0) := \nabla \f(x_0) = J$.
                    
                    We compute $\nabla \f'$ for arbitrary point, since
                    $$
                        \dfrac{\partial \f'(x_0 + u)_i}{\partial u_j} = J_{ij} + \sgn(\langle u,\,v \rangle) \langle u,\,v \rangle v_j k \nablaS_i,
                    $$
                    we know $\nabla \f'(x_0 + u) = J + \sgn(\langle u,\,v \rangle) k \langle u,\,v \rangle \nablaS v^\T$.
                    Consider arbitrary $u_1, u_2$:
                    \begin{itemize}[leftmargin=*]
                        \item 
                            If $\langle u_1,\,v \rangle \cdot \langle u_2,\,v \rangle \ge 0$,
                            $\nabla \f'(x_0 + u_1) - \nabla \f'(x_0 + u_2) = \sgn(\langle u_1,\,v \rangle)k\langle u_1-u_2,\,v \rangle \nablaS v^\T$.
                            Therefore,
                            $$
                                \dfrac{\lambda_{\max} (\nabla \f'(x_0 + u_1) - \nabla \f'(x_0 + u_2))}{\|u_1 - u_2\|_2} \le \dfrac{|\langle u_1 - u_2,\,v \rangle|}{\|u_1 - u_2\|_2} k\lambda_{\max} (\nablaS v^\T) \le kL_s \le \beta_\f.
                            $$
                            
                        \item 
                            If $\langle u_1,\,v\rangle \cdot \langle u_2,\,v\rangle < 0$, without loss of generality, let $\langle u_1,\,v \rangle > 0$ and $\langle u_2,\,v \rangle <0$.
                            Therefore,
                            $$
                                \nabla \f'(x_0 + u_1) - \nabla \f' (x_0 + u_2) = k\langle u_1 + u_2,\,v \rangle \nablaS v^\T.
                            $$
                            Since $\langle u_1,\, v \rangle > 0$ and $\langle u_2,\, v \rangle < 0$, $|\langle u_1 + u_2,\, v \rangle| \le |\langle u_1 - u_2,\, v\rangle|$.
                            Thus,
                            $$
                                \dfrac{\lambda_{\max} (\nabla \f'(x_0 + u_1) - \nabla \f'(x_0 + u_2))}{\|u_1 - u_2\|_2} \le \dfrac{|\langle u_1 + u_2,\, v\rangle|}{\|u_1 - u_2\|_2} k \lambda_{\max} (\nablaS v^\T) \le \dfrac{|\langle u_1 - u_2,\, v\rangle|}{\|u_1 - u_2\|_2} k \lambda_{\max} (\nablaS v^\T) \le \beta_\f.
                            $$
                    \end{itemize}
                    According to the smoothness definition, $\f'$ is $\beta_\f$-smooth.
                \end{proof}
                
                Now let us inject $\f'$ into the Taylor expansion expression for $S\left( \f'(x_0 + \delta u) \right)$ in a similar way as \Cref{eq:tmp1,eq:tmp2,eq:tmp3}, where $u$ is a unit vector, i.e., $\|u\|_2 = 1$:
                \begin{equation}
                    \begin{aligned}
                        & S(\f'(x_0 + \delta u)) \\
                        = & S\left( \f(x_0) + \delta Ju + \dfrac{1}{2}\sgn\left(\langle u,\,v\rangle\right) \langle u,\,v\rangle^2 \delta^2 k \nabla S \right) \\
                        = & S(\f(x_0)) + \delta \nablaS^\T J u + \dfrac{1}{2}\sgn(\langle u,\,v\rangle)\langle u,\,v \rangle^2\delta^2k\|\nablaS\|^2 +  \\
                        & \hspace{3em} \dfrac{1}{2}\theta^2 \left( \delta Ju +\dfrac{1}{2}\sgn(\langle u,\,v \rangle)\langle u,\,v \rangle^2 \delta^2 k \nablaS \right)^\T \mathbf{H} \left( \delta Ju + \dfrac{1}{2}\sgn(\langle u,\,v \rangle)\langle u,\,v \rangle^2 \delta^2 k \nablaS \right),
                    \end{aligned}
                    \label{eq:non-linear-caseA-1}
                \end{equation}
                where $\theta \in [-1,\,1]$ is depended on $S$, and $\mathbf{H}$ is the Hessian matrix of $S$ at point $x_0$.
                Because $x_0$ is the boundary point, we have $S(\f(x_0)) = 0$.
                
                We can bound the last term as such:
                $$
                \begin{aligned}
                    & \Big| \dfrac{1}{2}\theta^2 \left( \delta Ju +\dfrac{1}{2}\sgn(\langle u,\,v \rangle)\langle u,\,v \rangle^2 \delta^2 k \nablaS \right)^\T \mathbf{H} \left( \delta Ju + \dfrac{1}{2}\sgn(\langle u,\,v \rangle)\langle u,\,v \rangle^2 \delta^2 k \nablaS \right) \Big| \\
                    \le & \dfrac{1}{2}\beta_S \left(\delta L_\f + \dfrac{1}{2} \langle u,\,v \rangle^2 \delta^2 k L_S \right)^2 = \dfrac{1}{2}\beta_S \delta^2 \left( L_\f + \dfrac{1}{2} \langle u,\,v \rangle^2 \delta k L_S \right)^2.
                \end{aligned}
                $$
                
                When $\langle u,\,v \rangle > 0$,
                from \Cref{eq:non-linear-caseA-1}, we get
                $$
                \begin{aligned}
                    S(\f'(x_0 + \delta u)) & \ge \delta \nablaS^\T J u + \dfrac{1}{2} \langle u,\,v \rangle^2 \delta^2 k L_S^2 - \dfrac{1}{2}\beta_S \delta^2 \left( L_\f + \dfrac{1}{2} \langle u,\,v \rangle^2 \delta k L_S \right)^2 \\
                    & = \delta \langle u,\,v \rangle \|v\|_2 + \dfrac{1}{2} \langle u,\,v \rangle^2 \delta^2 k L_S^2 - \dfrac{1}{2}\beta_S \delta^2 \left( L_\f + \dfrac{1}{2} \langle u,\,v \rangle^2 \delta k L_S \right)^2,
                \end{aligned}
                $$
                and similarly, when $\langle u,\,v \rangle < 0$, we get
                $$
                \begin{aligned}
                    S(\f'(x_0 + \delta u)) & \le \delta \langle u,\,v \rangle \|v\|_2 - \dfrac{1}{2} \langle u,\,v \rangle^2 \delta^2 k L_S^2 + \dfrac{1}{2}\beta_S \delta^2 \left( L_\f + \dfrac{1}{2} \langle u,\,v \rangle^2 \delta k L_S \right)^2.
                \end{aligned}
                $$
                
                Therefore,
                \begin{equation}
                    |\langle u,\,v \rangle| \|v\|_2
                    \ge 
                    - \dfrac{1}{2} \langle u,\,v \rangle^2 \delta k L_S^2 + \dfrac{1}{2}\beta_S \delta \left( L_\f + \dfrac{1}{2} \langle u,\,v \rangle^2 \delta k L_S \right)^2
                    \,\Longrightarrow\,
                    \sgn(S(\f(x_0 + \delta u))) = \sgn(\langle u,\,v \rangle).
                    \label{eq:non-linear-caseA-2}
                \end{equation}
                Denote $h\left(k;\,\langle u,\,v \rangle\right)$ to the RHS:
                $$
                    h\left(k;\,\langle u,\,v \rangle\right) := - \dfrac{1}{2} \langle u,\,v \rangle^2 \delta k L_S^2 + \dfrac{1}{2}\beta_S \delta \left( L_\f + \dfrac{1}{2} \langle u,\,v \rangle^2 \delta k L_S \right)^2.
                $$
                When $k = 0$,
                $$
                    h(k;\,\langle u,\,v \rangle) = \dfrac{1}{2} \beta_S \delta L_\f^2,
                    \quad
                    \dfrac{\partial h(k;\,\langle u,\,v \rangle)}{\partial k}\Big|_{k=0} = -\dfrac{1}{2} \langle u,\,v \rangle^2 \delta L_S^2 + \dfrac{1}{2} \langle u,\,v \rangle^2 \delta^2 L_S L_\f \beta_S = \dfrac{1}{2}\langle u,\,v \rangle^2 \delta L_S (\delta L_\f \beta_S - L_S).
                $$
                Therefore, when $|\langle u,\,v \rangle| \ge \epsilon' > 0$,
                $$
                    \dfrac{\partial h(k;\,\langle u,\,v \rangle)}{\partial k}\Big|_{k=0} \le \dfrac{1}{2} \epsilon'^2 \delta L_S (\delta L_\f \beta_S - L_S) < 0,
                $$
                and thus there exists small $\epsilon > 0, \eta > 0$, when $k = \epsilon$ and $|\langle u,\,v \rangle| \ge \epsilon'$, 
                $h(k;\,\langle u,\,v\rangle) < \dfrac{1}{2}\beta_S \delta L_\f^2 - \eta$.
                
                As a result, from \Cref{eq:non-linear-caseA-2}, we know that 
                when $|\langle u,\,v \rangle| \ge \epsilon'$, if $|\langle u,\,v \rangle| \|v\|_2 \ge \dfrac{1}{2}\beta_S \delta L_\f^2 - \eta$, 
                $\sgn(S(\f(x_0 + \delta u))) = \sgn(\langle u,\,v \rangle)$.
                In other words, let
                $$
                    \omega' := \dfrac{1}{2}\beta_S \delta L_\f^2 - \eta,
                $$
                then this $\omega'$ satisfies the condition in \Cref{eq:non-linear-omega-cond}.
                
                Following the same proof as in \Cref{thm:1} using $\omega'$, we get the desired lower bound.
            \end{proof}
            
            \Cref{thm:2,thm:3} present two constructions of nonlinear projection $\f'$ which is better than the corresponding linear projection, and they also provide a checkable condition to examine whether the given nonlinear projection is `good' in terms of outperforming corresponding linear projection.
            Since the two constructed projections are quite different from each other, we conjecture that such nonlinear projection is not rare or specific.
            Even though there is no theoretically guaranteed approach for searching such `good' nonlinear projection, in experiments, we show that AE, VAE, or GAN are possible choices that usually work well in practice.

    \subsection{Improve The Gradient Estimation}
    \label{sec:adx-improve-gradient-estimation}

        In \Cref{thm:1,thm:2}, we relate the cosine similarity bound to variables characterizing the projection $\f$ such as $\nabla \f$, $L_\f$, $\beta_\f$.
        By examining the change tendency of the bound with respect to these variables, we learn ways for improving the gradient estimation in terms of improving its cosine similarity with the true gradient.
        
        \begin{itemize}[leftmargin=*]
            \item 
            Increase the alignment between $\nablaS$ and $\nabla \f$:\\
            The term ${\|\nablafS\|_2}/{\|\nablaS\|_2}$ reveals that, we should increase the alignment between $\nablaS$ and $\nabla \f$ to improve the cosine similarity.
            When $L_S$ and $L_\f$ are fixed, if they are more aligned, $\|\nablaS^\T \nabla \f\|_2^2$ is larger so that the lower bound becomes larger.
            It implies that the mapping $\f$ should reflect the main components of $\nablaS$ as much as possible.
            Similar conclusion is shown for QEBA in \Cref{sec:adx-comparison-of-different-gradient-estimators}.
            
            \item
            Reduce the subspace dimension $n$ and increase number of queries $B$:\\
            When $\nablaS$ and $\nabla \f$ can be aligned, it is better to keep the subspace dimension of $\f$, $n$, be small.
            The reason is analyzed in \Cref{sec:adx-comparison-of-different-gradient-estimators} when comparing HSJA and QEBA.
            At the same time, increasing number of queries $B$ is also helpful, according to the query complexity analysis in \Cref{sec:theory-implications}.
            
            
            \item
            If we can find good nonlinear projection, decrease the smoothness; 
            otherwise, increase the smoothness and decrease step size $\delta$:\\
            If the a good nonlinear projection can be found, we consider the bound in \Cref{thm:2}, which shows the outcome of a good nonlinear projection.
            Learned from its $\omega$ in \Cref{eq:nonlinear-case-omega}, increasing $\beta_\f$, i.e., decreasing the smoothness, could reduce $\omega$ and hence improve cosine similarity bound.
            If the good nonlienar projection cannot be found, we consider the bound in \Cref{thm:1}, which bounds the general projections.
            To reduce $\omega$ in this case which is defined by \Cref{def:w}, we need to reduce $\beta_\f$, i.e., increase the smoothness, and reduce the step size $\delta$.
            We remark that the choice of step size $\delta$ needs to consider many other factors as \citet {chen2020hopskipjumpattack} outlined.

        \end{itemize}

\section{Target Models} 
In this section, we introduce the target models used in the experiments including the implementation details and the model performance.
\subsection{Implementation Details}
\label{sec:target_model_details}
\paragraph{Offline Models.}
Following~\citet{cvpr2020QEBA}, we use models based on a pretrained ResNet-18 model as the target models. For models that are finetuned, cross entropy error is employed as the loss function and is implemented as `\texttt{torch.nn.CrossEntropyLoss}' in PyTorch.

For ImageNet, no finetuning is performed as the pretrained target model is trained exactly on ImageNet. The model is loaded with PyTorch command `\texttt{torchvision.models.resnet18(pretrained=True)}' following the documentation~\citep{pytorch_pretrained_models}.

For CelebA, the target model is finetuned to do binary classification on image attributes. Among the 40 binary attributes associated with each image, we sort the attributes according to how balance the numbers of positive and negative samples are. The more balanced the dataset is, it is better for the classification model training. The top-5 balanced attributes are `Attractive', `Mouth\_Slightly\_Open', `Smiling', `Wearing\_Lipstick', `High\_Cheekbones'. Though the `Attractive' attribute is the most balanced one, it is more objective than subjective, thus we instead use the second attribute `Mouth\_Slightly\_Open'. 

For MNIST and Cifar10 datasets, we first do linear interpolation and get $224\times 224$ images, then the target model is finetuned to do 10-way classification. One reason for doing interpolation is that our proposed method reduces query complexity when the original data dimension is high so it is more illustrative after upsampling. The linear interpolation step also makes image sizes consistent among all the tasks and experiments.

We report the benign target model performance for the four datasets in Table~\ref{tab:target_model_performance}.

\paragraph{Commercial Online API.}
Among all the APIs provided by the Face++ platform~\citep{facepp_main}, we use the `Compare' API~\citep{facepp-compare-api} which takes two images as input and returns a confidence score of whether they are the same person if there are faces in the two images. This is also consistent with the same experiment in QEBA~\citep{cvpr2020QEBA}.
In implementation during the attack process, the two image arrays with floating number values are first converted to integers and stored as jpg images on disk. Then they are encoded as base64 binary data and sent as POST request to the request URL~\citep{facepp-compare-url}. We set the similarity threshold as $50\%$ in the experiments following QEBA~\citep{cvpr2020QEBA}: when the confidence score is equal to or larger than $50\%$, we consider the two faces to belong to the `same person', vice versa.

For source-target images that are from two different persons, the goal of the attack is to get an \advimage that looks like the \targetimage (has low mean squared error distance between the \advimage and \targetimage), but is predicted as `same person' with the \sourceimage.
We randomly sample source-target image pairs from the CelebA dataset that are predicted as different persons by the `Compare' API. Then we apply the \sysname pipeline with various nonlinear projection models for comparison. 

\subsection{Model Performance of Target Models}
\label{sec:target_model_performance}
The benign accuracies of the target model ResNet-18 on the datasets are shown in Table~\ref{tab:target_model_performance}.
\begin{table}[t]
    \centering
    \caption{The benign model accuracies of the target model (ResNet-18).}
    \begin{tabular}{c|c|c|c}
    \hline
        Dataset & CelebA & CIFAR10 & MNIST \\
        \hline
        Benign Accuracy & 0.9417 & 0.8796 & 0.9938\\
        \hline
    \end{tabular}
    \label{tab:target_model_performance}
\end{table}

\section{Nonlinear Projection Based Gradient Estimator}
In this section, we introduce the details of nonlinear projection models including the model structure, training procedure. We also introduce how the projection models are used in the \sysname process including the gradient estimation and attack implementation details.

\subsection{Generative Model Structure}
\label{sec:estimator_structure}
\paragraph{AE and VAE.}
We borrow the idea from U-Net~\citep{ronneberger2015u} which has the structure of an information contraction path and an expanding path, with a small latent representation in the middle.

Define 2D convolution layer Conv2d(in\_channels, out\_channels, kernel\_size, padding\_size).

Define the DoubleConv(in\_channels, out\_channels) layer as composed of 6 layers: a 2D convolution layer Conv2d(in\_channels, out\_channels) with kernel size $3$ and padding size $1$; a 2D batch normalization layer BatchNorm2d(out\_channels); a ReLU layer; another 2D convolution layer Conv2d(out\_channels, out\_channels) with kernel size $3$ and padding size $1$; a 2D batch normalization layer BatchNorm2d(out\_channels); and a ReLU layer.

Define the Down(in\_channels, out\_channels) layer with two components: a max-pooling layer MaxPool2d with kernel size $2$; a DoubleConv(in\_channels, out\_channels) as defined above.

Likewise, the Up(in\_channels, out\_channels) is defined with two components: a up-scaling layer and a DoubleConv(in\_channels, out\_channels) as defined above.

The AE and VAE models have similar structures except for the fact that the encoder part of VAE has two output layers to produce the mean and standard deviation vectors, and the AE only has one. The detailed network structures are shown in Table~\ref{tab:ae_vae_structure}. The n\_channels is the number of image channels determined by the image dataset. For the grey-scale images in MNIST, there is only $1$ channel; for the other three colored datasets (ImageNet, CelebA and CIFAR10), there are RGB channels so n\_channels is $3$. The latent dimension of the two models is $48\times 14\times 14 = 9408$.

\begin{table}[t]
    \centering
    \caption{The detailed network structure for AE and VAE models.}
    \label{tab:ae_vae_structure}
    \begin{tabular}{c|c||c|c}
    \hline
        Layer Name & AE & Layer Name & VAE \\
        \hline
        InConv & DoubleConv(n\_channels, 24) & InConv & DoubleConv(n\_channels, 24) \\
        Down1 & Down(24, 24) & Down1 & Down(24, 24)  \\
        Down2 & Down(24, 48) & Down2 & Down(24, 48)  \\
        Down3 & Down(48, 48) & Down3 & Down(48, 48)  \\
        Down4 & Down(48, 48) & DownMu & Down(48, 48) \\
        - & - & DownStd & Down(48, 48) \\
        \hline
        Up1 & Up(48, 48) & Up1 & Up(48, 48) \\
        Up2 & Up(48, 48) & Up2 & Up(48, 48) \\
        Up3 & Up(48, 24) & Up3 & Up(48, 24) \\
        Up4 & Up(24, 24) & Up4 & Up(24, 24) \\
        OutConv & Conv2d(24, n\_channels, 1, 0) & OutConv & Conv2d(24, n\_channels, 1, 0)\\
    \hline
    \end{tabular}
\end{table}

\paragraph{GAN.}
Define ConvBlock(in\_channels, out\_channels, n\_kernel, n\_stride, n\_pad, transpose, leaky) with three layers: a 2D convolution layer; a batch normalization layer; and a nonlinear ReLU layer.

For ImageNet and CelebA, the detailed model network structures for the generator and discriminator are listed in Table~\ref{tab:gan_generator_structure} and Table~\ref{tab:gan_discriminator_structure}.
\begin{table}[t]
    \centering
    \caption{The detailed model structure for generator in GAN.}
    \label{tab:gan_generator_structure}
    \begin{tabular}{c}
    \hline
        Generator \\
        \hline
        ConvBlock(z\_latent, 128, 4, 1, 0, transpose=True, leaky=True) \\
        ConvBlock(128, 64, 3, 2, 1, transpose=True, leaky=False) \\
        ConvBlock(64, 64, 4, 2, 1, transpose=True, leaky=False) \\
        ConvBlock(64, 32, 4, 2, 1, transpose=True, leaky=False) \\
        ConvBlock(32, 32, 4, 2, 1, transpose=True, leaky=False) \\
        ConvBlock(32, 16, 4, 2, 1, transpose=True, leaky=False) \\
        nn.ConvTranspose2d(16, n\_channels, 4, 2, 1, bias=False) \\
        nn.Tanh() \\
        \hline
    \end{tabular}
\end{table}
\begin{table}[t]
    \centering
    \caption{The detailed model structure for discriminator in GAN.}
    \label{tab:gan_discriminator_structure}
    \begin{tabular}{c}
    \hline
        Discriminator\\
        \hline
        nn.Conv2d(n\_channels, 16, 4, 2, 1, bias=False) \\
        nn.LeakyReLU(0.2, inplace=True) \\
        ConvBlock(16, 32, 4, 2, 1, transpose=False, leaky=True) \\
        ConvBlock(32, 32, 4, 2, 1, transpose=False, leaky=True) \\
        ConvBlock(32, 64, 4, 2, 1, transpose=False, leaky=True) \\
        ConvBlock(64, 64, 4, 2, 1, transpose=False, leaky=True) \\
        ConvBlock(64, 128, 3, 2, 1, transpose=False, leaky=True) \\
        nn.Conv2d(128, 1, 4, 1, 0, transpose=False, leaky=True) \\
        \hline
    \end{tabular}
\end{table}

For CIFAR10 and MNIST, we use DCGAN~\citep{radford2015unsupervised} structure with pretrained weights from \burl{https://github.com/csinva/gan-vae-pretrained-pytorch/} and add a linear interpolation layer to resize the generated images to size $224\times 224$.

\subsection{Estimator Training Procedure}
\label{sec:estimator_training}
The attacker first trains a set of reference models that are generally assumed to have different structures compared to the blackbox target model. Nonetheless, attacker-trained reference models can generate accessible gradients and provide valuable information on the distribution of the target model gradients.

In our case, there are five reference models with different backbones compared with the target model, while the implementation and training details are similar to the target model in Section~\ref{sec:target_model_details}. The benign test accuracy results for CelebA, Cifar10 and MNIST datasets are shown in Table~\ref{tab:celeba_benign_acc}, Table~\ref{tab:cifar10_benign_acc} and Table~\ref{tab:mnist_benign_acc} respectively.
After the reference models are trained, their gradients with respect to the training data points are generated with PyTorch automatic differentiation function with command `loss.backward()'. The loss is the cross entropy between the prediction scores and the ground truth labels. 

For ImageNet and CelebA, since the number of images is large, the gradient dataset generated by reference models is also too large to be handled in our GPU memory especially when we evaluate the baseline method QEBA-I~\citep{cvpr2020QEBA} since it requires approximate PCA. Thus, we randomly sample $500,000$ gradient images ($100,000$ per reference model) for each of ImageNet and CelebA and fix them throughout the experiments for fair comparison.
For CIFAR10 and MNIST, there are fewer images and the machine can handle them properly, so we use the whole gradient dataset generated with $250,000$ gradient images for CIFAR10 ($50,000$ per reference model) and $300,000$ ($60,000$ per reference model) gradient images for MNIST.

The generative models are trained on the gradient images of the corresponding dataset generated as above. 

\subsection{Reference Model Performance}
\label{sec:ref_model_performance}
Intuitively, with well-trained reference models that perform comparatively with the target models, the attacker can get gradient images that are in a more similar distribution with the target model's gradients for training, thus increasing the chance of an attack with higher quality.
The reference model performances in terms of prediction accuracy for CelebA, Cifar10 and MNIST datasets are shown in Table~\ref{tab:celeba_benign_acc}, Table~\ref{tab:cifar10_benign_acc}, and Table~\ref{tab:mnist_benign_acc}. The model performances are comparable to those of the target models.
\begin{table}[!ht]
    \centering
    \caption{The benign model accuracies of the reference models for CelebA dataset (attribute: `mouth\_slightly\_open').}
    \begin{tabular}{c|c|c|c|c|c}
    \hline
        CelebA & DenseNet-121 & ResNet-50 & VGG16 & GoogleNet & WideResNet \\
        \hline
        Benign Accuracy & 0.9415 & 0.9410 & 0.9417 & 0.9315 & 0.9416 \\
        \hline
    \end{tabular}
    \label{tab:celeba_benign_acc}
\end{table}

\begin{table}[!ht]
    \centering
    \caption{The benign model accuracies of the reference models for Cifar10 dataset (linearly interpolated to size $3\times 224\times 224$).}
    \begin{tabular}{c|c|c|c|c|c}
    \hline
        Cifar10 & DenseNet-121 & ResNet-50 & VGG16 & GoogleNet & WideResNet \\
        \hline
        Benign Accuracy & 0.9079 & 0.8722 & 0.9230 & 0.9114 & 0.8568 \\
        \hline
    \end{tabular}
    \label{tab:cifar10_benign_acc}
\end{table}

\begin{table}[!ht]
    \centering
    \caption{The benign model accuracies of the reference models for MNIST dataset (linearly interpolated to size $224\times 224$).}
    \begin{tabular}{c|c|c|c|c|c}
    \hline
        MNIST & DenseNet-121 & ResNet-50 & VGG16 & GoogleNet & WideResNet \\
        \hline
        Benign Accuracy & 0.9919 & 0.9916 & 0.9948 & 0.9943 & 0.9938\\
        \hline
    \end{tabular}
    \label{tab:mnist_benign_acc}
\end{table}

\subsection{Nonlinear Projection Based Gradient Estimation}
We provide the pseudo code for the gradient estimation process with the nonlinear projection functions in Algorithm~\ref{alg:nonlinear_grad_est}.
\label{sec:alg_pseudocode}
\begin{algorithm}
\caption{Nonlinear Projection Based Gradient Estimation}
\label{alg:nonlinear_grad_est}
\begin{algorithmic}[1]
\renewcommand{\algorithmicrequire}{\textbf{Input:}}
 \renewcommand{\algorithmicensure}{\textbf{Output:}}
 \REQUIRE a data point on the decision boundary ${\bf x} \in \mathbb{R}^m$, nonlinear projection function $\f$, number of random sampling $B$, access to query the decision of target model $\phi(\cdot) = \sgn(S(\cdot))$.
 \ENSURE the approximated gradient $\widetilde{\nablaS}(\xadv{t})$.
 \STATE sample $B$ random Gaussian vectors of the lower dimension: $v_b \in \mathbb{R}^{n}$.
 \STATE use nonlinear projection function to project the random vectors to the gradient space: $u_b = \f(v_b)\in \mathbb{R}^m$.
 \STATE get query points by adding perturbation vectors to the original point on the decision boundary $\xadv{t} + \delta \f{v_b}$.
 \STATE Monte Carlo approximation for the gradient: $$\widetilde{\nablaS}(\xadv{t}) = \frac1B \sum_{b=1}^B \phi\left(\xadv{t}+\delta \f({v}_b)\right) \f({v}_b) = \frac1B \sum_{b=1}^B \sgn \left(S\left(\xadv{t}+\delta \f({v}_b)\right) \right) \f({v}_b).$$
 \RETURN $\widetilde{\nablaS}(\xadv{t})$.
\end{algorithmic}
\end{algorithm}

\subsection{Attack Implementation}
\label{sec:attack_process_implementation}
The goal is to generate an attack image that looks similar as the \targetimage but is predicted as the label of the \sourceimage. 
We fix the random seed to $0$ so that the samples are consistent across different runs and various methods to ensure reproducibility and to facilitate fair comparison. 
\paragraph{Offline Models.}
During the attack, we randomly sample source-target pairs of images from each of the corresponding datasets. We query the offline models with the sampled images to make sure both \sourceimage and \targetimage are predicted as their ground truth labels and the labels are different so that the attack is nontrivial.
For the same dataset, the results of different attack methods are reported as the average of the same $50$ randomly sampled pairs.
\paragraph{Online API.}
For the online API attacks, the source-target pairs are sampled from the face image dataset CelebA.

\section{Quantitative Results}

\subsection{Attack Success Rate for Offline Models}
\begin{figure*}[tpb]
    \centering
    \includegraphics[width=\textwidth]{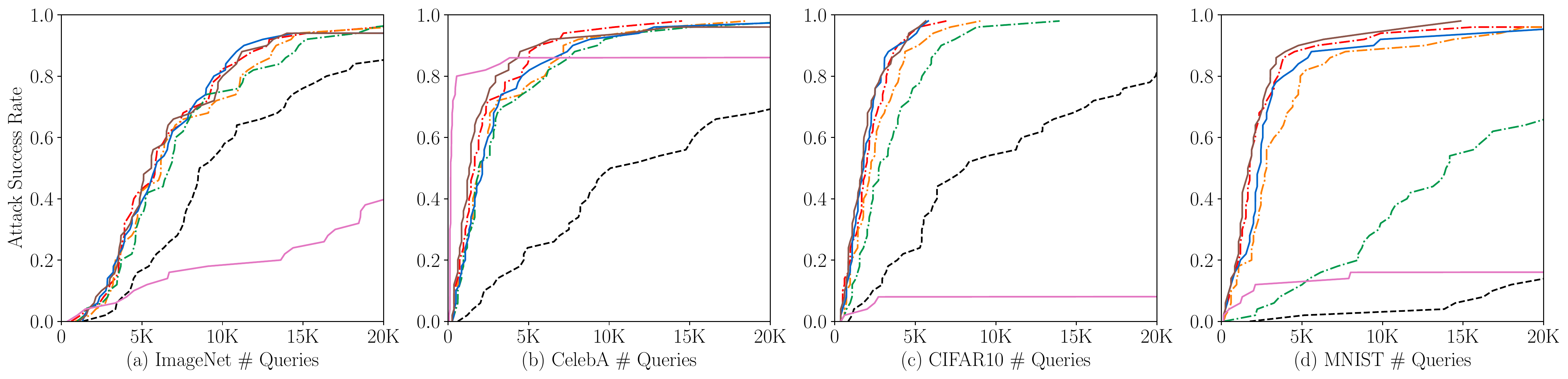}
    \caption{\small The attack success rate vs query number for four different datasets.}
    \label{fig:mean_success_nq}
\end{figure*}

The `successful attack' is defined as the \advimage reaching some predefined mean squared error (MSE) distance threshold. Note that because of the varying complexity of tasks and images among different datasets, we set different MSE distance thresholds for different datasets. For example, ImageNet images are the most complicated and the task is most difficult. Thus, we set larger (looser) threshold for it. Specifically, the thresholds are shown in Table~\ref{tab:attack_success_threshold}. The attack success rates on the four datasets are shown in Table~\ref{fig:mean_success_nq}.

\begin{table}[t]
    \centering
    \caption{The mean squared error (MSE) distance thresholds used for four datasets that determine whether the attack is successful.}
    \label{tab:attack_success_threshold}
    \begin{tabular}{c|c|c|c|c}
    \hline
        Dataset & ImageNet & CelebA & MNSIT & CIFAR10  \\
        \hline
        MSE Threshold & $1^{-3}$ & $1^{-4}$ & $5^{-3}$ & $1^{-4}$\\
        \hline
    \end{tabular}
\end{table}



\subsection{Proxy for the \texorpdfstring{$\omega$}{omega} Value}
\label{sec:omega_value}
\begin{figure*}[tpb]
    \centering
    \includegraphics[width=\textwidth]{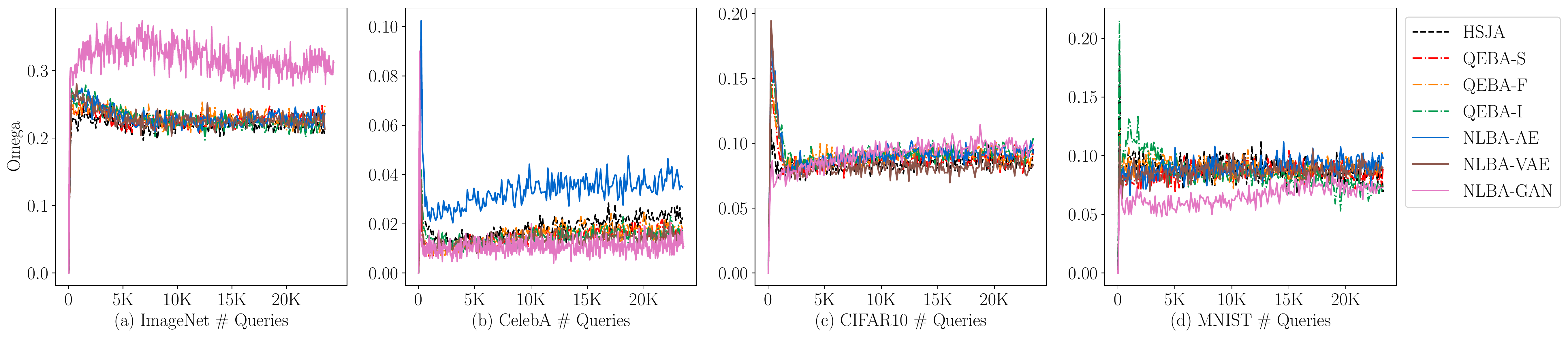}
    \caption{\small The $\omega$ value at different queries for attacks on diverse datasets.}
    \label{fig:mean_omega_nq}
\end{figure*}

According to the analysis in \Cref{sec:generalized-gradient-estimator}, smaller $\omega$ leads to better gradient estimation.
\Linyi{add a justification for using proxy}
The exact computation of $\omega$ requires computing the tight Lipschitz and smoothness constant for both the projection $\f$ and the difference function $S$, which is challenging.
Therefore, we provide a proxy of the $\omega$ variable during the training.
When estimating the gradient at each \boundaryimage $\xadv{t}$ point with \Cref{eq:our_gradient_estimation}, there are some perturbations that contribute negatively in the Monte-Carlo estimation. More formally, a perturbation vector $\f(v_b)$ has a negative contribution to the gradient estimation if 
\begin{equation}
    \sgn\left(S\left(\xadv{t}+\delta \f({v}_b)\right) \right) \neq \sgn\left(\cos \big\langle(\widetilde{\nablaS}(\xadv{t}),\, \f(v_b) \big\rangle \right).
    \label{eq:40}
\end{equation}
In other words, the sign of target model prediction disagrees with the sign of the cosine similarity between the estimated gradient and the perturbation direction.
We deem the ratio of samples that satisfy \Cref{eq:40} as the proxy of $\omega$.
The results are shown in Figure~\ref{fig:mean_omega_nq}.


\subsection{Correlation between \texorpdfstring{$\omega$}{omega} and Cosine Similarity}
\label{sec:correlation_omega_cos}
\begin{figure*}
    \centering
    \includegraphics[width=\textwidth]{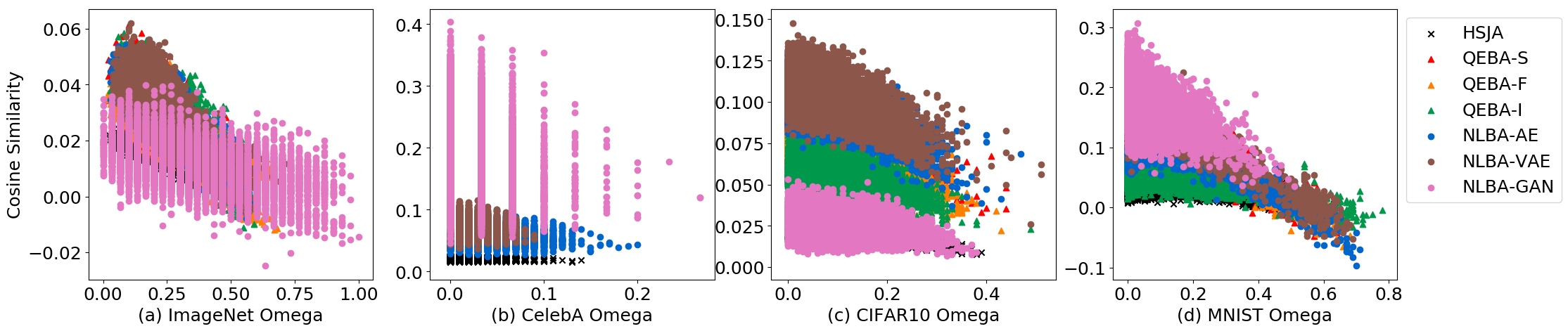}
    \caption{The cosine similarity values at different $\omega$ values for attacks on diverse datasets.}
    \label{fig:correlation_omega_cos}
\end{figure*}

\begin{figure*}[htpb]
\centering
\begin{subfigure}[t]{0.22\linewidth}
    \centering
    \includegraphics[width=\textwidth]{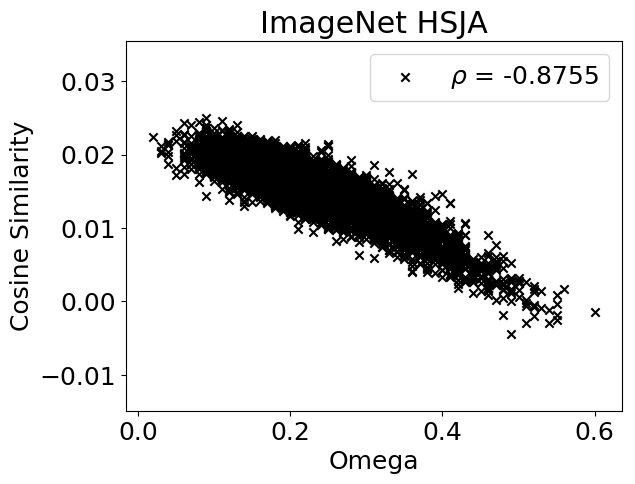}
\end{subfigure}
\begin{subfigure}[t]{0.22\linewidth}
    \centering
    \includegraphics[width=\textwidth]{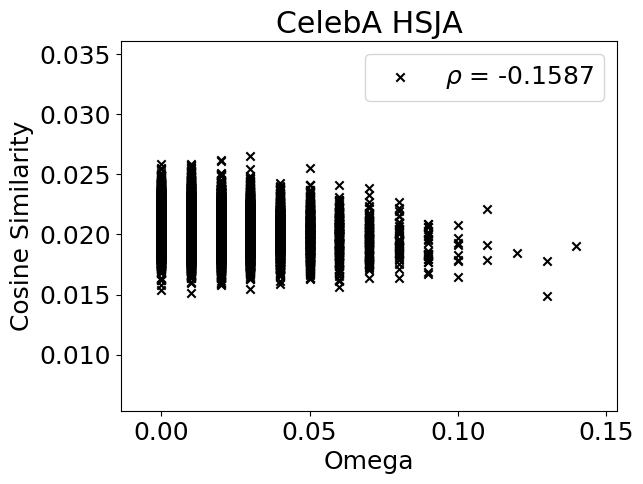}
\end{subfigure}
\begin{subfigure}[t]{0.22\linewidth}
    \centering
    \includegraphics[width=\textwidth]{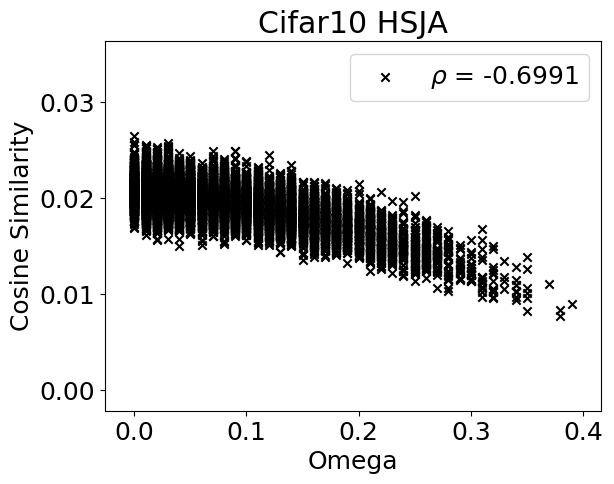}
\end{subfigure}
\begin{subfigure}[t]{0.22\linewidth}
    \centering
    \includegraphics[width=\textwidth]{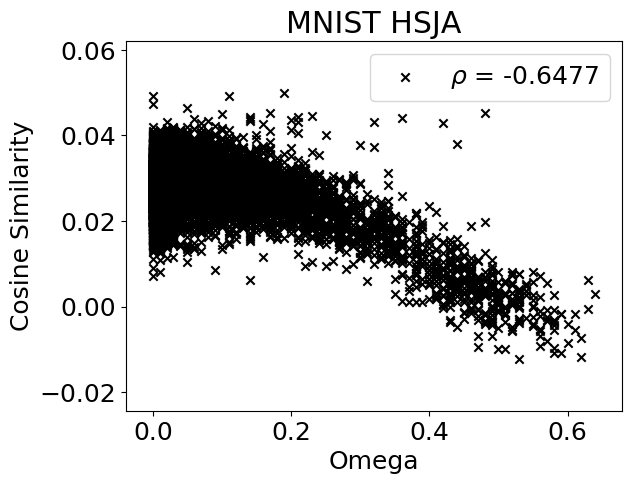}
\end{subfigure}
\hspace{1mm}

\begin{subfigure}[t]{0.22\linewidth}
    \centering
    \includegraphics[width=\textwidth]{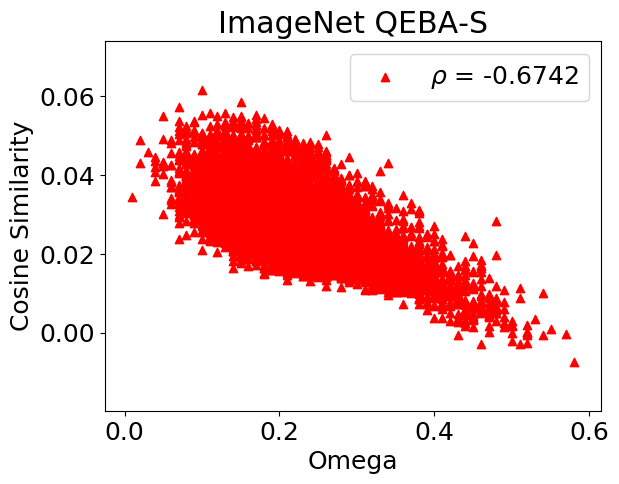}
\end{subfigure}
\begin{subfigure}[t]{0.22\linewidth}
    \centering
    \includegraphics[width=\textwidth]{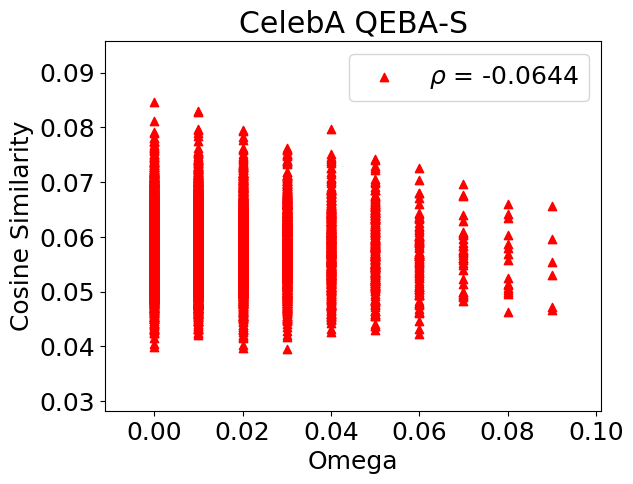}
\end{subfigure}
\begin{subfigure}[t]{0.22\linewidth}
    \centering
    \includegraphics[width=\textwidth]{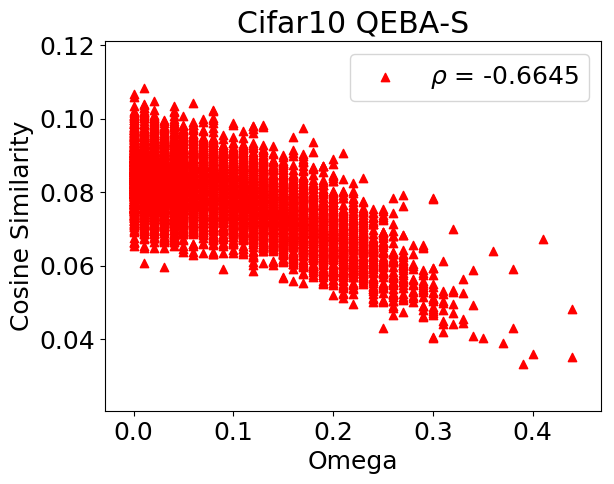}
\end{subfigure}
\begin{subfigure}[t]{0.22\linewidth}
    \centering
    \includegraphics[width=\textwidth]{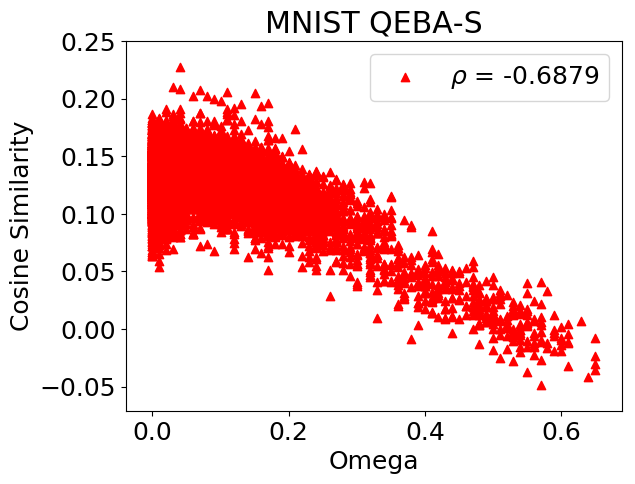}
\end{subfigure}
\hspace{1mm}

\begin{subfigure}[t]{0.22\linewidth}
    \centering
    \includegraphics[width=\textwidth]{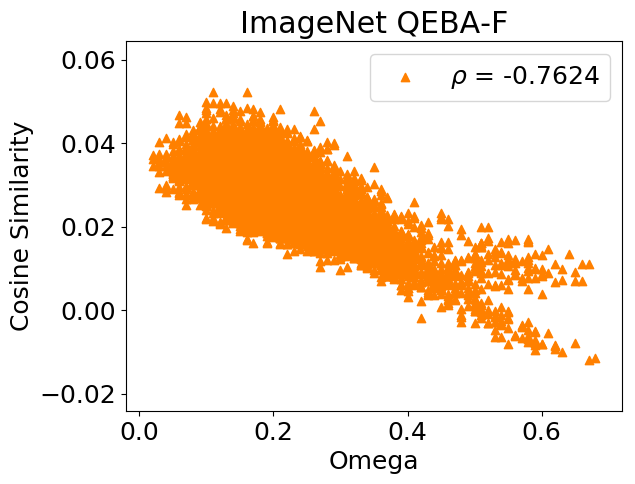}
\end{subfigure}
\begin{subfigure}[t]{0.22\linewidth}
    \centering
    \includegraphics[width=\textwidth]{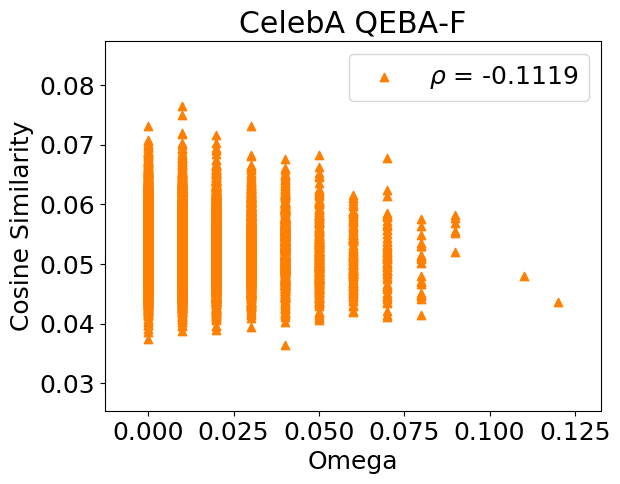}
\end{subfigure}
\begin{subfigure}[t]{0.22\linewidth}
    \centering
    \includegraphics[width=\textwidth]{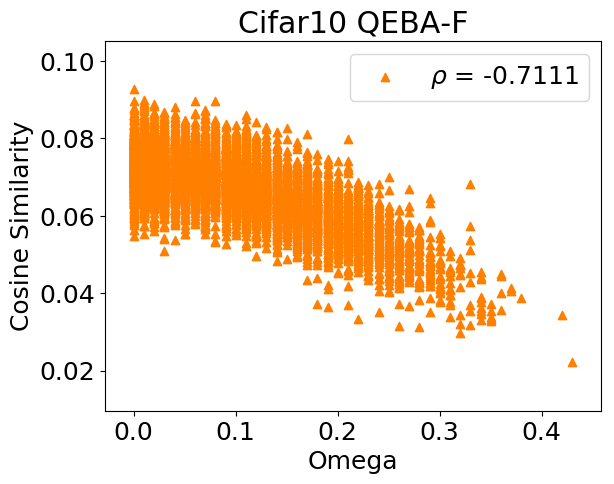}
\end{subfigure}
\begin{subfigure}[t]{0.22\linewidth}
    \centering
    \includegraphics[width=\textwidth]{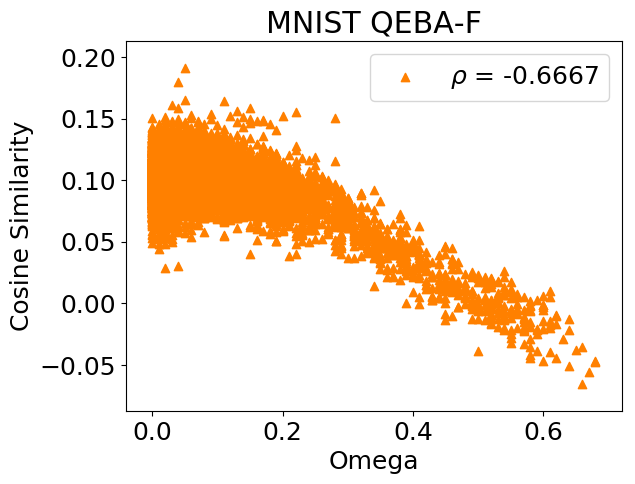}
\end{subfigure}
\hspace{1mm}

\begin{subfigure}[t]{0.22\linewidth}
    \centering
    \includegraphics[width=\textwidth]{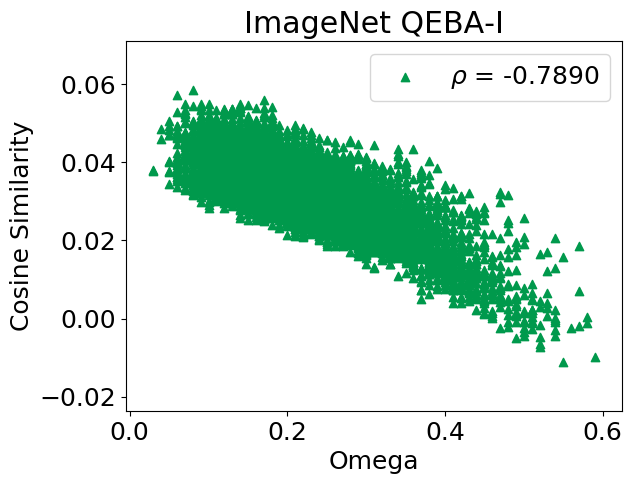}
\end{subfigure}
\begin{subfigure}[t]{0.22\linewidth}
    \centering
    \includegraphics[width=\textwidth]{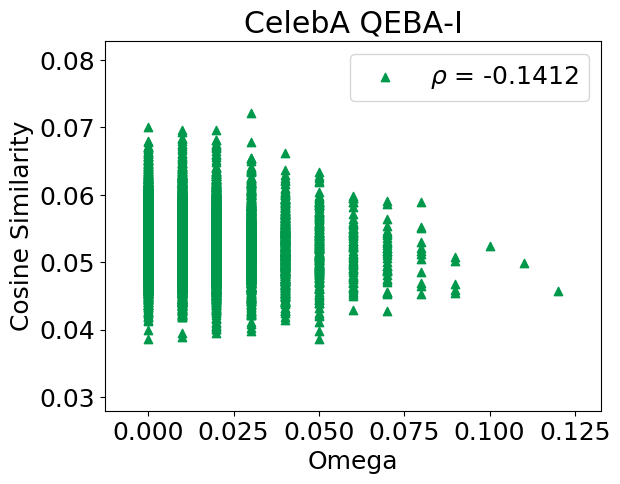}
\end{subfigure}
\begin{subfigure}[t]{0.22\linewidth}
    \centering
    \includegraphics[width=\textwidth]{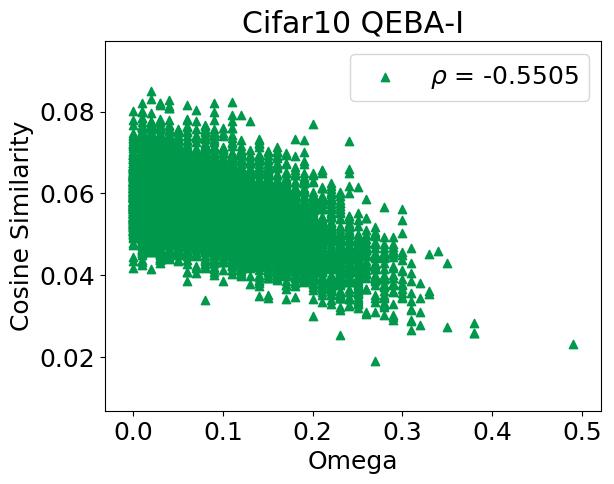}
\end{subfigure}
\begin{subfigure}[t]{0.22\linewidth}
    \centering
    \includegraphics[width=\textwidth]{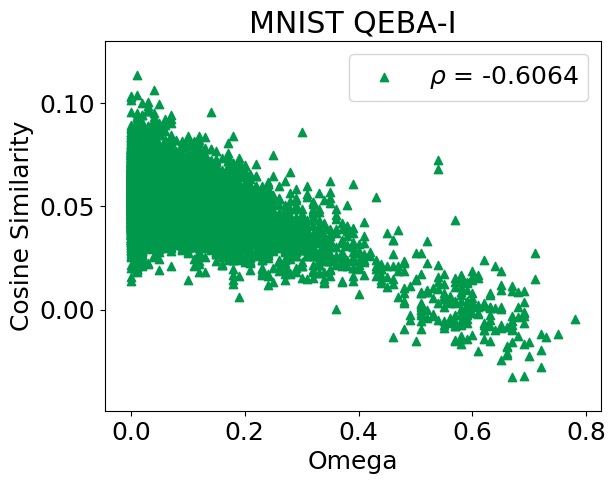}
\end{subfigure}
\hspace{1mm}

\begin{subfigure}[t]{0.22\linewidth}
    \centering
    \includegraphics[width=\textwidth]{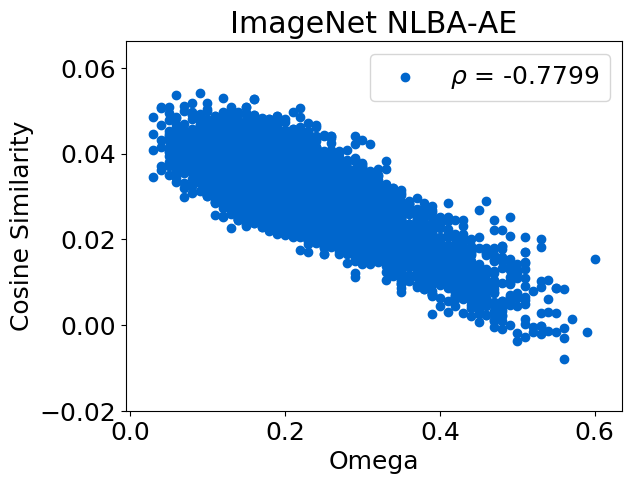}
\end{subfigure}
\begin{subfigure}[t]{0.22\linewidth}
    \centering
    \includegraphics[width=\textwidth]{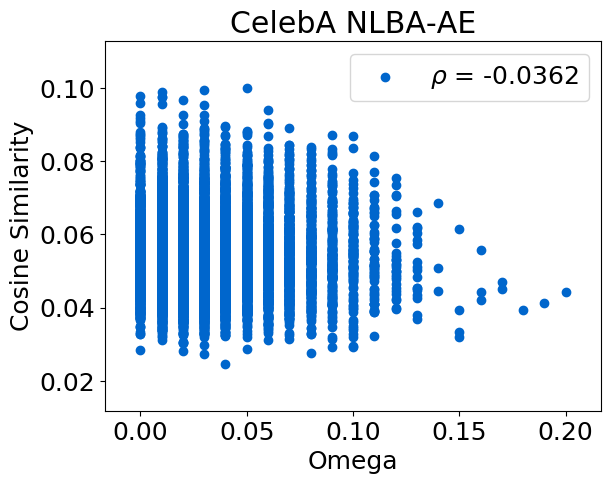}
\end{subfigure}
\begin{subfigure}[t]{0.22\linewidth}
    \centering
    \includegraphics[width=\textwidth]{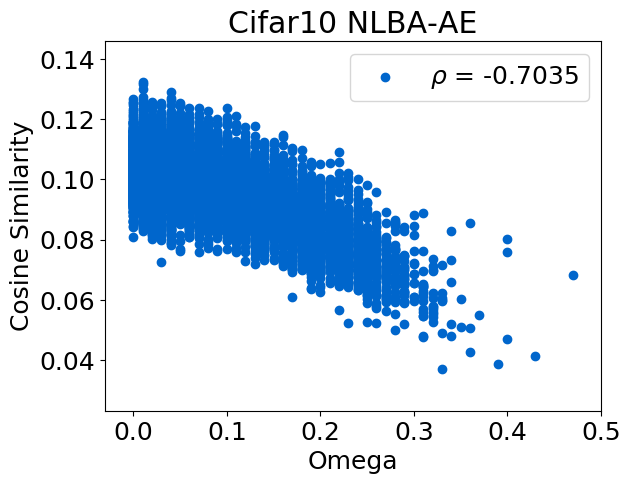}
\end{subfigure}
\begin{subfigure}[t]{0.22\linewidth}
    \centering
    \includegraphics[width=\textwidth]{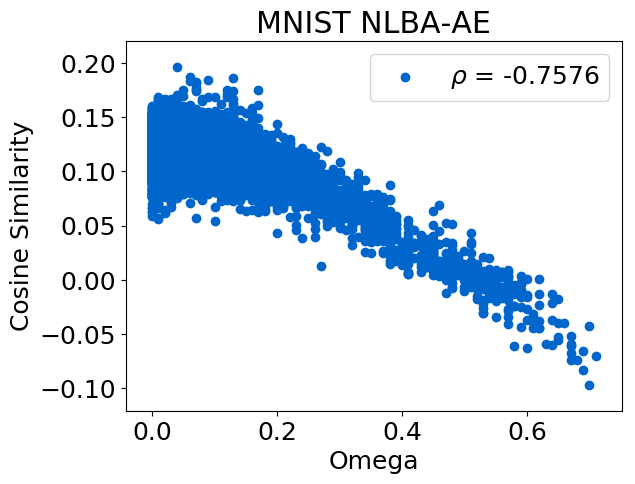}
\end{subfigure}
\hspace{1mm}

\begin{subfigure}[t]{0.22\linewidth}
    \centering
    \includegraphics[width=\textwidth]{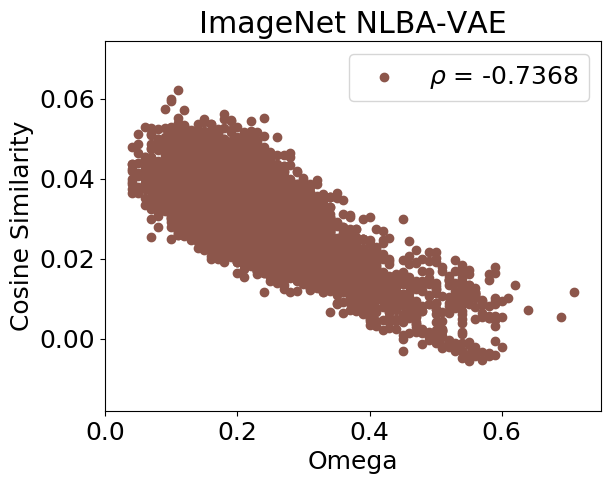}
\end{subfigure}
\begin{subfigure}[t]{0.22\linewidth}
    \centering
    \includegraphics[width=\textwidth]{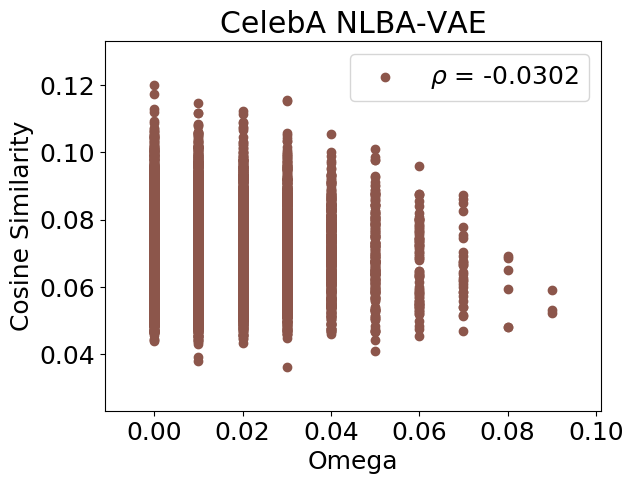}
\end{subfigure}
\begin{subfigure}[t]{0.22\linewidth}
    \centering
    \includegraphics[width=\textwidth]{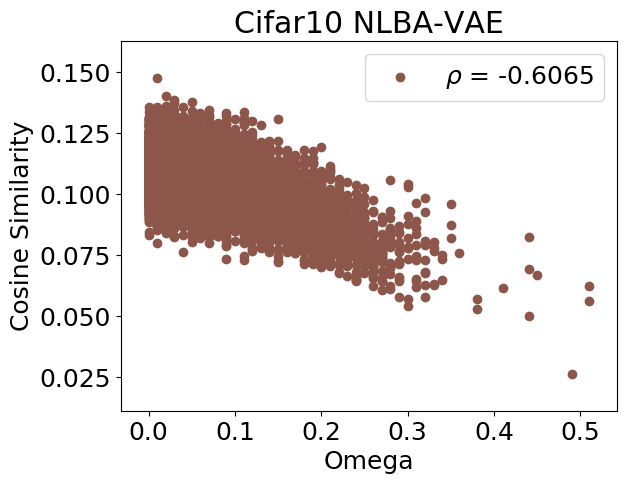}
\end{subfigure}
\begin{subfigure}[t]{0.22\linewidth}
    \centering
    \includegraphics[width=\textwidth]{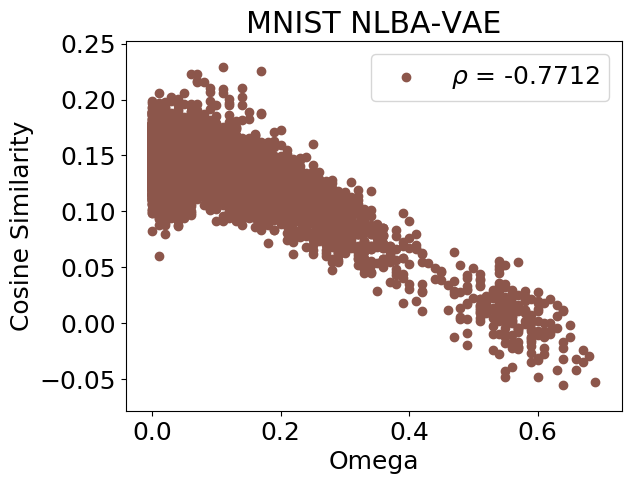}
\end{subfigure}
\hspace{1mm}

\begin{subfigure}[t]{0.22\linewidth}
    \centering
    \includegraphics[width=\textwidth]{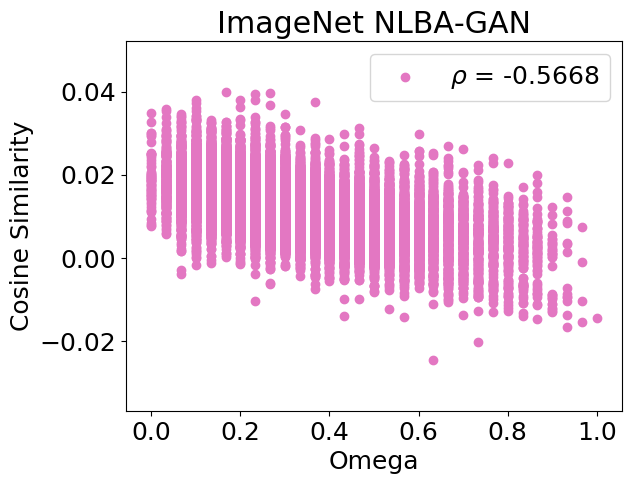}
\end{subfigure}
\begin{subfigure}[t]{0.22\linewidth}
    \centering
    \includegraphics[width=\textwidth]{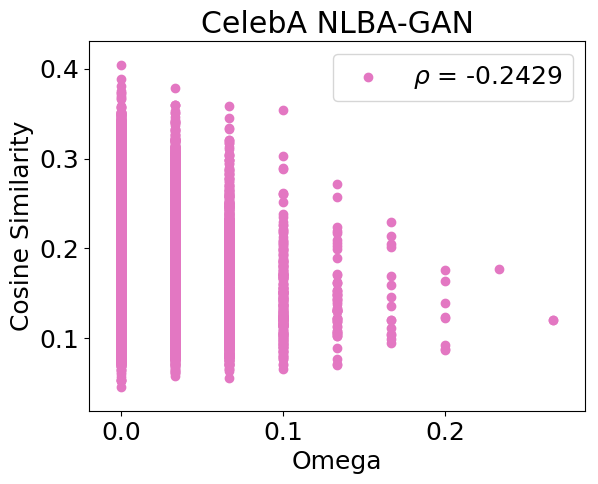}
\end{subfigure}
\begin{subfigure}[t]{0.22\linewidth}
    \centering
    \includegraphics[width=\textwidth]{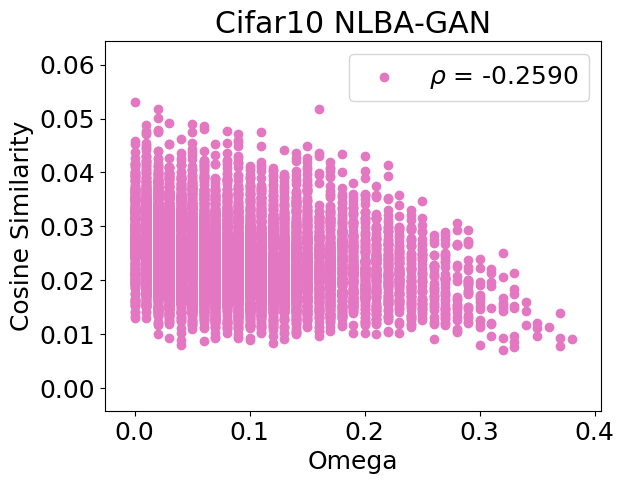}
\end{subfigure}
\begin{subfigure}[t]{0.22\linewidth}
    \centering
    \includegraphics[width=\textwidth]{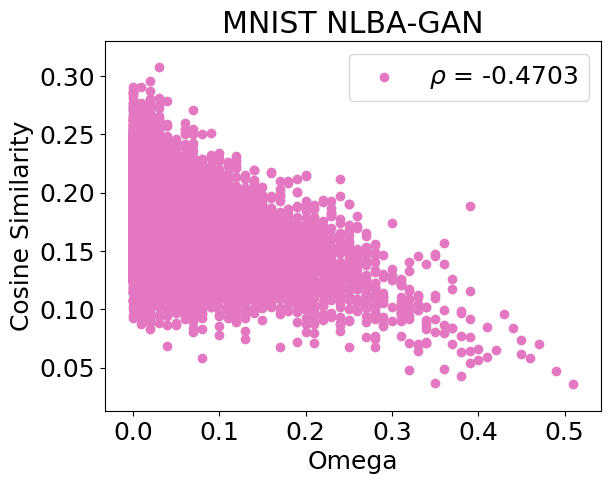}
\end{subfigure}
\hspace{1mm}

\caption{The $\omega$ vs cosine similarity values for the $4$ datasets and $7$ projection methods.}
\label{fig:omega_cos_4datasets_7methods}
\end{figure*}

To verify the correlation between variable $\omega$ and the cosine similarity measure as proposed by Equation~\ref{eq:low-dimension-good-lower-bound} in Section~\ref{sec:theory-implications}, we calculate the two variables during the attack process on different datasets with various projection models and plot them as $x$ and $y$ axis in Figure~\ref{fig:correlation_omega_cos}. 
The cosine similarity values exhibit a descending trend with the increase of the $\omega$ values. To further confirm this, we calculate Pearson's correlation score and the results are shown in Figure~\ref{fig:omega_cos_4datasets_7methods}. On ImageNet, CIFAR10, and MNIST datasets, the Pearson's correlation scores are negative with large absolute values, showing the $\omega$ and cosine similarity values have a strong negative correlation. On CelebA dataset, the negative correlation between the two variables is less statistically significant.

\section{Qualitative Results}
In this section, we present the qualitative results for attacking both offline models and online APIs.
\subsection{CelebA Case Study}
\label{sec:celeba_3_all_attack_process_early}
\begin{figure}
    \centering
    \includegraphics[width=0.9\linewidth]{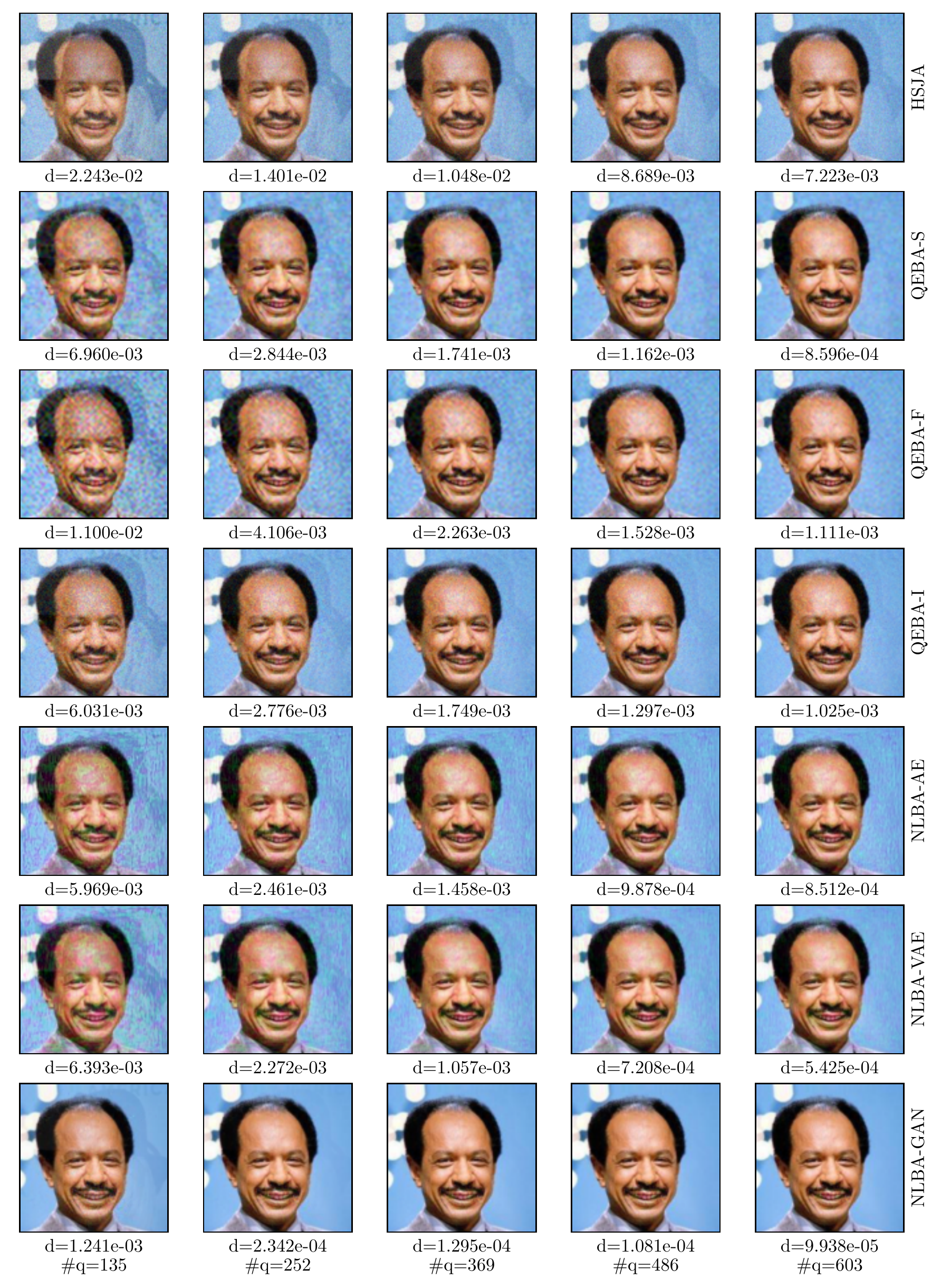}
    \caption{The attack performance of all the \sysname methods and the baseline methods on one pair of image of the CelebA dataset. The \sourceimage and \targetimage of this case study are shown in Figure~\ref{fig:case_study_celeba_src_tgt_imgs}. The $d$ in the figure denotes the perturbation magnitude (mean squared error) of the adversarial example with respect to the target-image. The $\#q$ values are the number of queries used at the point for each column.}
    \label{fig:celeba_3_all_attack_process_early}
\end{figure}

The whole figure for the case study on CelebA dataset of the attack performance at the early stage of the attack process is shown in Figure~\ref{fig:celeba_3_all_attack_process_early}.

\subsection{Offline Models}
\label{sec:offline_qualitative}
\begin{figure}[t]
    \centering
    \includegraphics[width=0.9\textwidth]{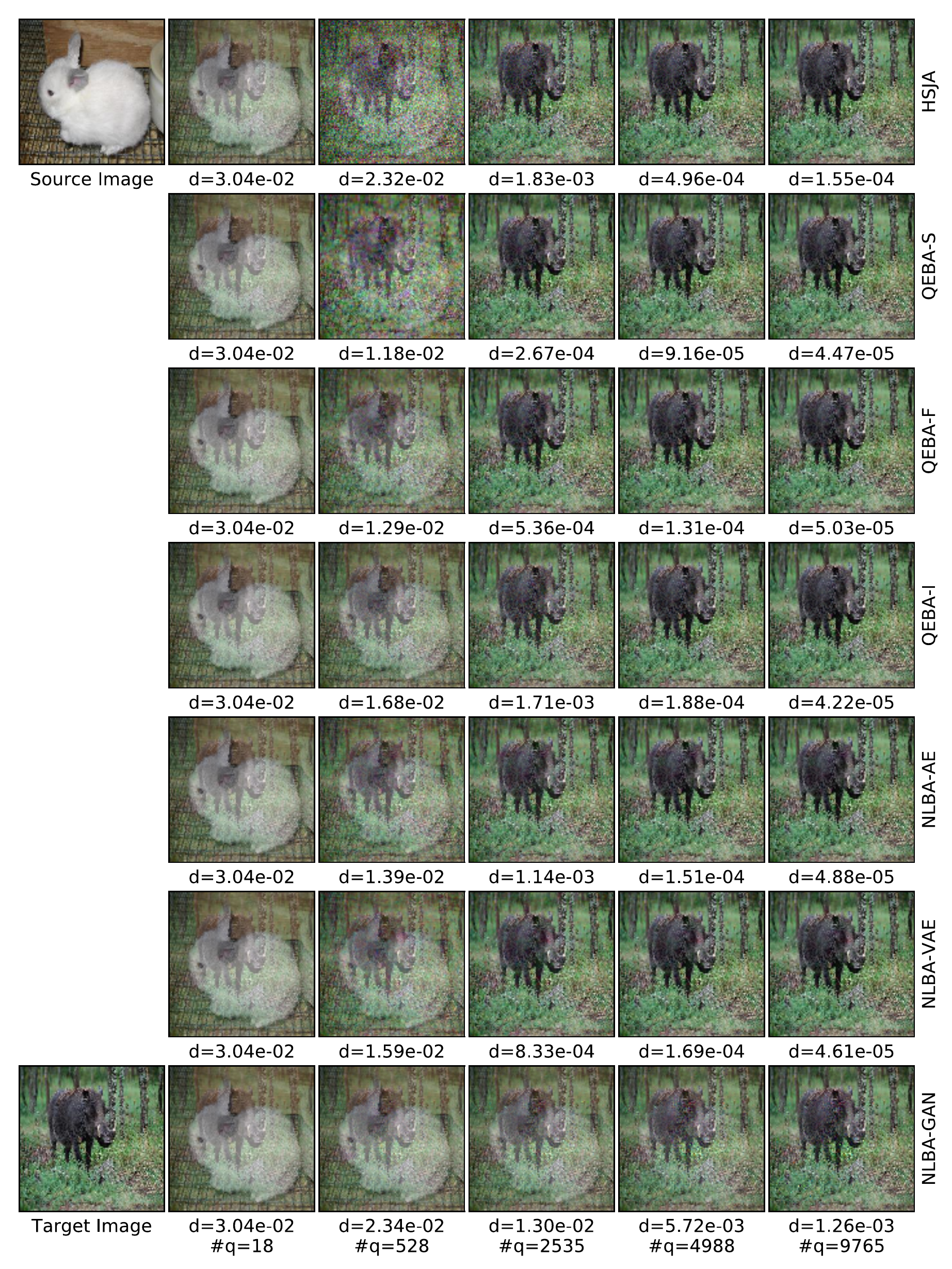}
    \caption{The qualitative case study of attacking ResNet-18 model on ImageNet dataset.}
    \label{fig:imagenet_attack_process}
\end{figure}

\begin{figure}[t]
    \centering
    \includegraphics[width=0.9\textwidth]{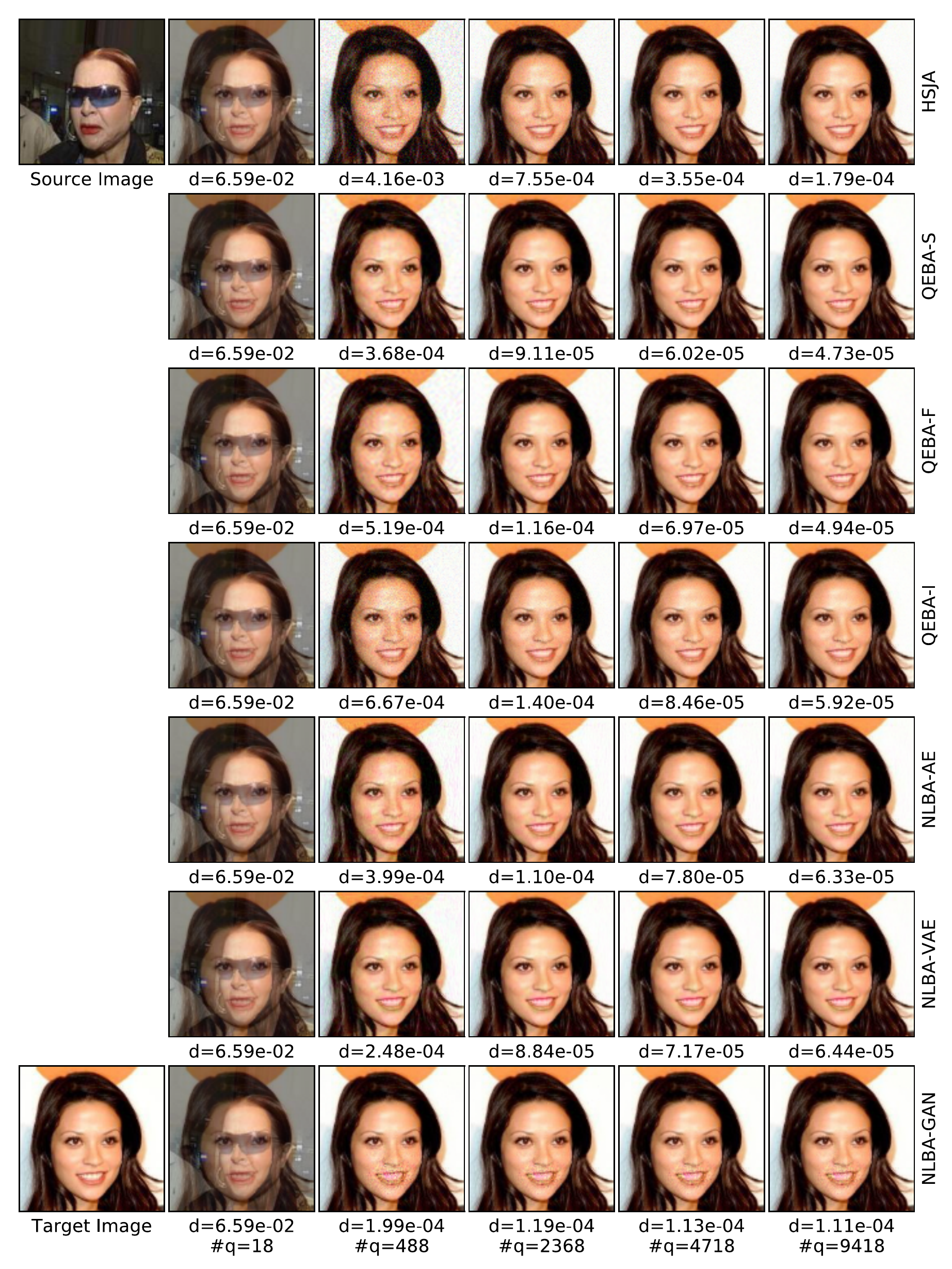}
    \caption{The qualitative case study of attacking ResNet-18 model on CelebA dataset.}
    \label{fig:celeba_attack_process}
\end{figure}

\begin{figure}[t]
    \centering
    \includegraphics[width=0.9\textwidth]{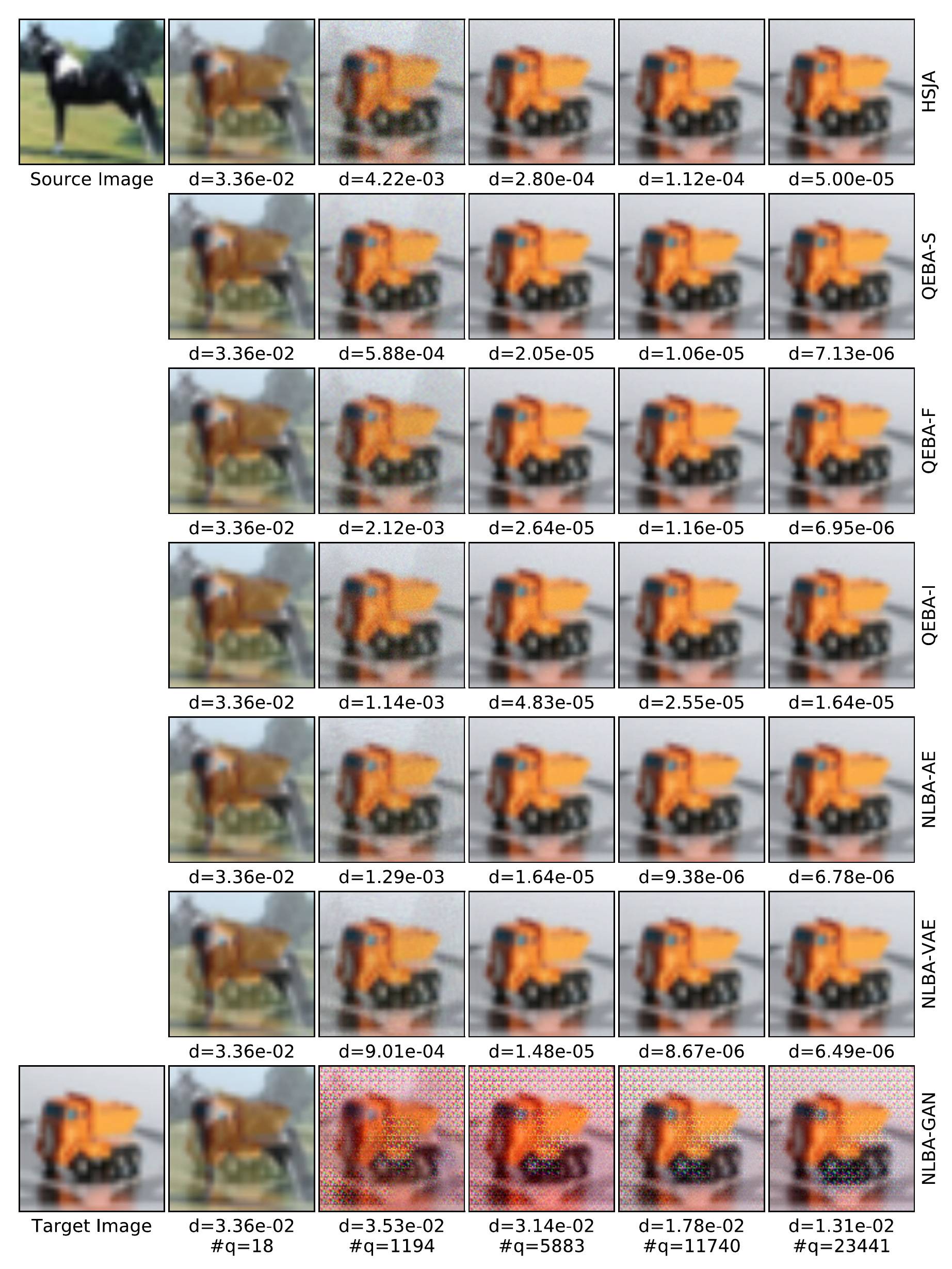}
    \caption{The qualitative case study of attacking ResNet-18 model on CIFAR10 dataset.}
    \label{fig:cifar10_attack_process}
\end{figure}

\begin{figure}[t]
    \centering
    \includegraphics[width=0.9\textwidth]{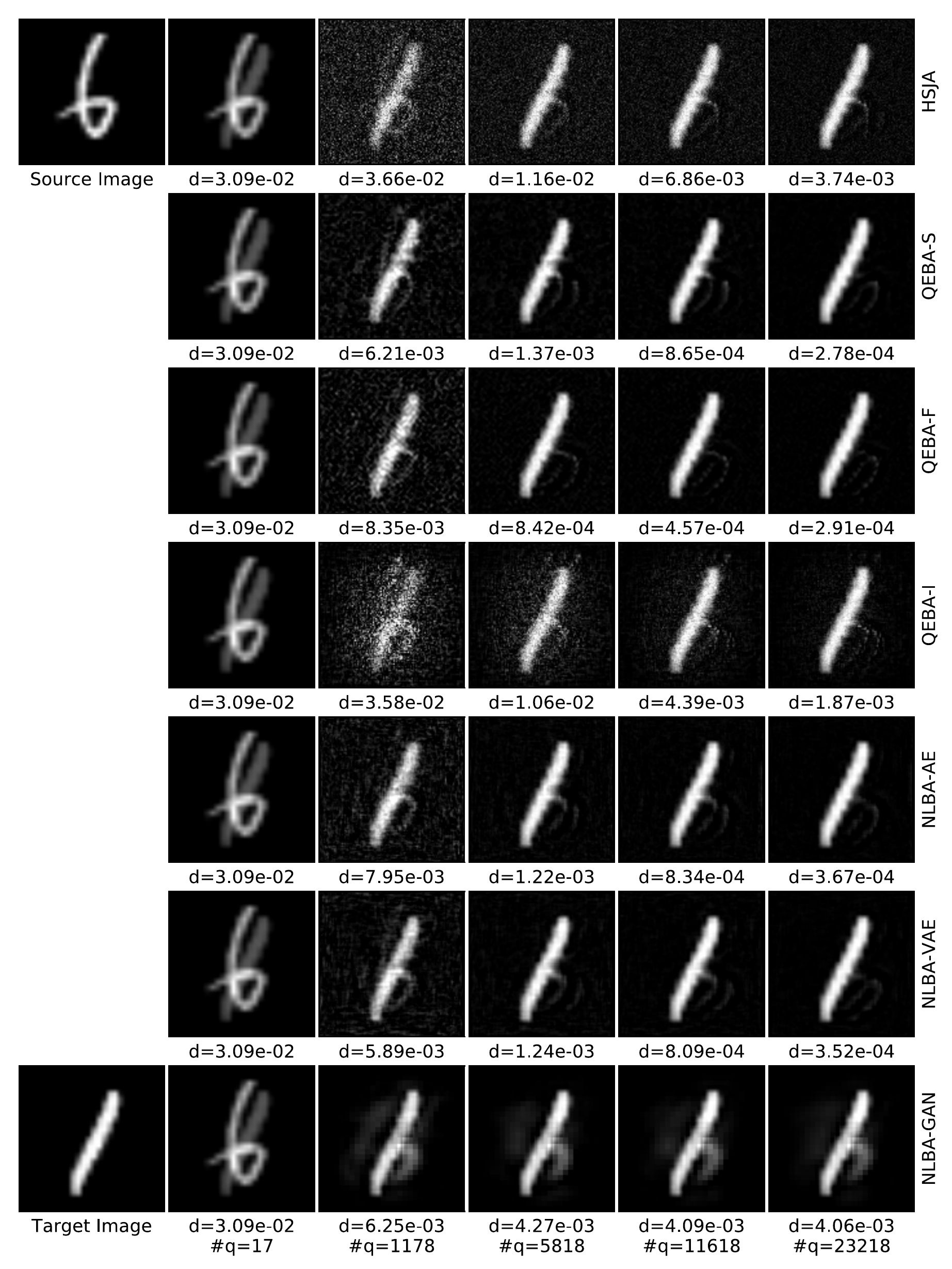}
    \caption{The qualitative case study of attacking ResNet-18 model on MNIST dataset.}
    \label{fig:mnist_attack_process}
\end{figure}

The goal of the attack is to generate an \advimage that looks like the \targetimage but has the same label with \sourceimage. 
We report qualitative results that show how the \advimage changes during the attack process in Figure~\ref{fig:imagenet_attack_process}, Figure~\ref{fig:celeba_attack_process}, Figure~\ref{fig:cifar10_attack_process} and Figure~\ref{fig:mnist_attack_process} for the four datasets respectively.
In the figures, the left-most column has two images: the \sourceimage and the \targetimage. They are randomly sampled from the corresponding dataset. We make sure images in the sampled pairs have different ground truth labels (otherwise the attack is trivial).
The other five columns each represents the \advimage at certain number of queries as indicated by $\#q$ at the bottom line. In other words, all images in these five columns can successfully attack the target model. Each row represents one method as shown on the right. The $d$ value under each image shows the MSE between the \advimage and the \targetimage. The smaller $d$ can get, the better the attack is.

\subsection{Commercial Online API Attack}
\begin{figure}[t]
    \centering
    \includegraphics[width=0.9\textwidth]{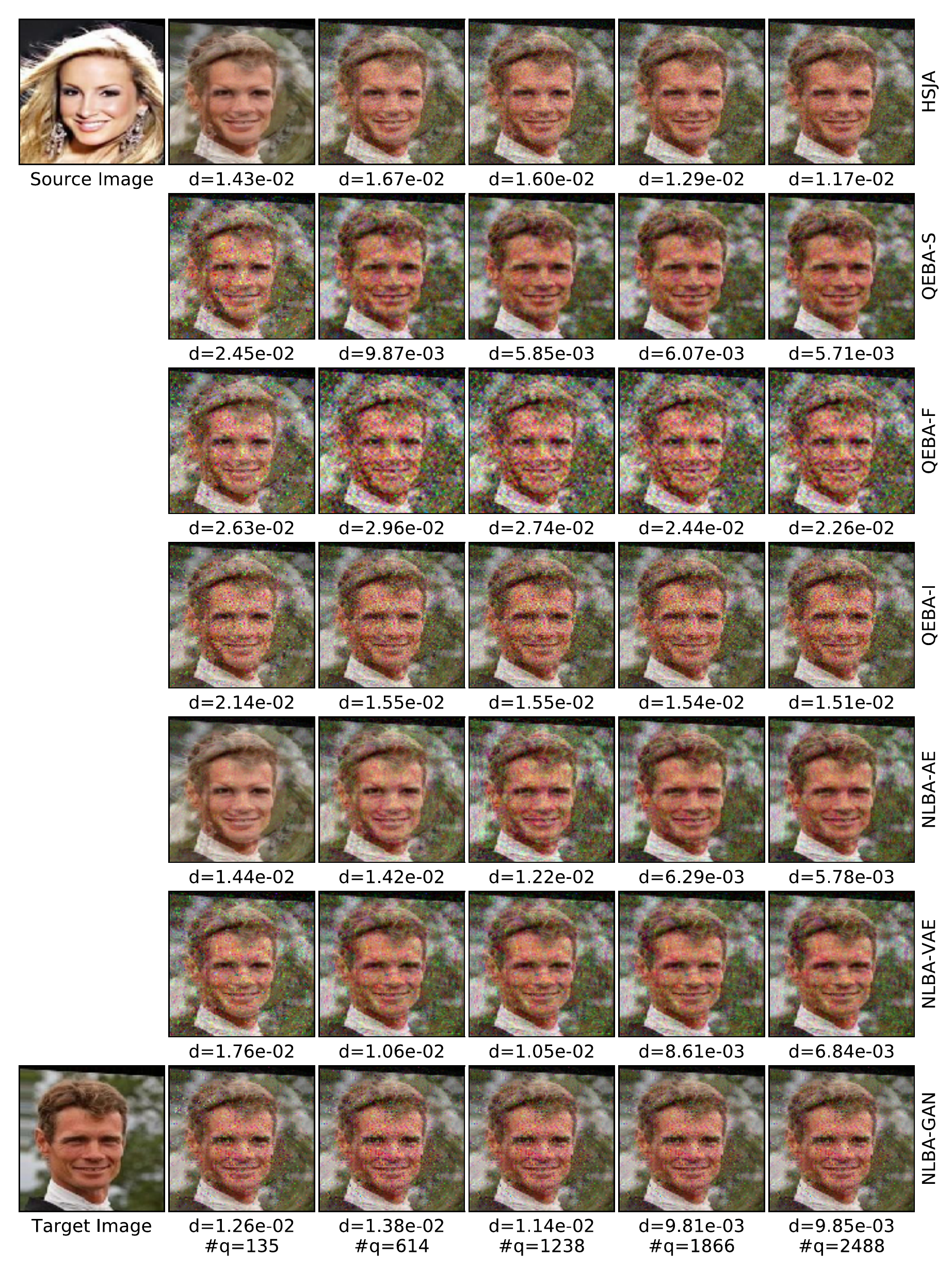}
    \caption{A case study of Face++ online API attack process. The source-target image pair is randomly sampled from CelebA dataset (ID: 163922 and 080037).}
    \label{fig:facepp_attack_process}
\end{figure}

As discussed in Section~\ref{sec:exp}, the goal is to generate an \advimage that looks like the \targetimage but is predicted as `same person' with the \sourceimage. 
In this case, we want to get images that looks like the man but is actually identified as the woman.
The qualitative results of attacking the online API Face++ `compare' is shown in Figure~\ref{fig:facepp_attack_process}.
In the figure, the \sourceimage and \targetimage are shown on the left-most column.
\label{sec:api_qualitative}

\end{document}